\documentclass{article}

\PassOptionsToPackage{numbers, sort}{natbib}

\usepackage[preprint]{neurips_2024}

\usepackage[utf8]{inputenc} %
\usepackage[T1]{fontenc}    %
\usepackage{url}            %
\usepackage{booktabs}       %
\usepackage{amsfonts}       %
\usepackage{nicefrac}       %
\usepackage{microtype}      %
\usepackage{times}
\usepackage{epsfig}
\usepackage{amsmath}
\usepackage{makecell} 
\usepackage{tabularx}
\usepackage{diagbox}
\usepackage{enumitem} 
\usepackage[table]{xcolor}
\usepackage{threeparttable}

\usepackage{xspace}
\usepackage{graphicx}
\usepackage{subcaption}
\usepackage{multirow} 
\usepackage{colortbl}
\usepackage{pifont}
\usepackage{wrapfig}
\usepackage{tcolorbox}
\usepackage{float}
\usepackage{marvosym}
\definecolor{citecolor}{HTML}{0071BC}

\definecolor{oursgray}{gray}{0.93}

\usepackage[
    pagebackref,           %
    breaklinks=true,       %
    colorlinks=true,       %
    linkcolor=red,         %
    citecolor=citecolor,   %
    urlcolor=black,        %
    bookmarks=false        %
]{hyperref}

\newcommand{\ie}{{\emph{i.e.}}}
\newcommand{\eg}{{\emph{e.g.}}}

\newcommand{\rsp}{\ \ --\ \ \ ~}

\newcommand{\todo}[1]{{\color{red}#1}}

\definecolor{myblue}{HTML}{d6f0ff}

\newcommand{\mytoprule}{
    \toprule
    \noalign{\vspace{-0.2mm}}
}

\newcommand{\mymidrule}{
    \noalign{\vspace{-0.8mm}}
    \midrule
    \noalign{\vspace{-1mm}}
}

\newcommand{\mybottomrule}{
    \noalign{\vspace{-0.6mm}}
    \bottomrule
}

\title{
InternVL3.5: Advancing Open-Source Multimodal Models in Versatility, Reasoning, and Efficiency
}

\author{
\vspace{-15px}\\
\scalebox{0.84}{
\textbf{
Weiyun Wang$^{*}$,
Zhangwei Gao$^{*}$,
Lixin Gu$^{*}$,
Hengjun Pu$^{*}$,
Long Cui$^{*}$,
Xingguang Wei$^{*}$,
Zhaoyang Liu$^{*}$,
}
}\\
\scalebox{0.84}{
\textbf{
Linglin Jing$^{*}$,
Shenglong Ye$^{*}$,
Jie Shao$^{*}$,
Zhaokai Wang$^{*}$,
Zhe Chen$^{*}$,
Hongjie Zhang,
Ganlin Yang,
Haomin Wang,
}
}\\
\scalebox{0.84}{
\textbf{
Qi Wei,
Jinhui Yin,
Wenhao Li,
Erfei Cui,
Guanzhou Chen,
Zichen Ding,
Changyao Tian,
Zhenyu Wu,
Jingjing Xie,
}
}\\
\scalebox{0.84}{
\textbf{
Zehao Li,
Bowen Yang,
Yuchen Duan,
Xuehui Wang,
Zhi Hou,
Haoran Hao,
Tianyi Zhang,
Songze Li,
Xiangyu Zhao,
}
}\\
\scalebox{0.84}{
\textbf{
Haodong Duan,
Nianchen Deng,
Bin Fu,
Yinan He,
Yi Wang,
Conghui He,
Botian Shi,
Junjun He,
Yingtong Xiong,
}
}\\
\scalebox{0.84}{
\textbf{
Han Lv,
Lijun Wu,
Wenqi Shao,
Kaipeng Zhang,
Huipeng Deng,
Biqing Qi,
Jiaye Ge,
Qipeng Guo,
Wenwei Zhang,
}
}\\
\scalebox{0.84}{
\textbf{
Songyang Zhang,
Maosong Cao,
Junyao Lin,
Kexian Tang,
Jianfei Gao,
Haian Huang,
Yuzhe Gu,
Chengqi Lyu,
}
}\\
\scalebox{0.84}{
\textbf{
Huanze Tang,
Rui Wang,
Haijun Lv,
Wanli Ouyang,
Limin Wang,
Min Dou,
Xizhou Zhu,
Tong Lu,
Dahua Lin,
}
}\\
\scalebox{0.84}{
\textbf{
Jifeng Dai,
Weijie Su,
Bowen Zhou\textsuperscript{\Letter},
Kai Chen\textsuperscript{\Letter},
Yu Qiao\textsuperscript{\Letter},
Wenhai Wang\textsuperscript{\Letter}$^\dagger$,
Gen Luo\textsuperscript{\Letter}$^\dagger$}
}
\vspace{4px}\\
\scalebox{0.9}{InternVL Team, Shanghai AI Laboratory }
\vspace{4px}\\
\small Code: \url{https://github.com/OpenGVLab/InternVL} \\
\small Model: \url{https://huggingface.co/OpenGVLab/InternVL3_5-241B-A28B} \\
\vspace{-15px}\\
}

\newcommand\blfootnote[1]{%
\begingroup
\renewcommand\thefootnote{}\footnote{#1}%
\addtocounter{footnote}{-1}%
\endgroup
}

\definecolor{baselinecolor}{gray}{.9}
\definecolor{reduce-color}{RGB}{67,178,68}

\begin{document}

\maketitle
\blfootnote{
* equal contribution; \Letter\  corresponding authors; $\dagger$ technical leaders. 
}

\begin{abstract}
We introduce \textit{InternVL3.5}, a new family of open-source multimodal models that significantly advances versatility, reasoning capability, and inference efficiency along the InternVL series. A key innovation is the \textit{Cascade Reinforcement Learning} (Cascade RL) framework, which enhances reasoning through a two-stage process: offline RL for stable convergence and online RL for refined alignment. This coarse-to-fine training strategy leads to substantial improvements on downstream reasoning tasks, \textit{e.g.,} MMMU and MathVista. To optimize efficiency, we propose a \textit{Visual Resolution Router} (ViR) that dynamically adjusts the resolution of visual tokens without compromising performance. Coupled with ViR, our \textit{Decoupled Vision-Language Deployment} (DvD) strategy separates the vision encoder and language model across different GPUs, effectively balancing computational load. These contributions collectively enable InternVL3.5 to achieve up to a +16.0\% gain in overall reasoning performance and a 4.05$\times$  inference speedup compared to its predecessor, \ie, InternVL3. In addition, InternVL3.5 supports novel capabilities such as GUI interaction and embodied agency. Notably, our largest model,  \ie,  InternVL3.5-241B-A28B, attains state-of-the-art results among open-source MLLMs across general multimodal, reasoning, text, and agentic tasks—narrowing the performance gap with leading commercial models like GPT-5. All models and code are publicly released.

\end{abstract}

\begin{figure}[!htbp] 
  \centering 
  \includegraphics[width=\linewidth]{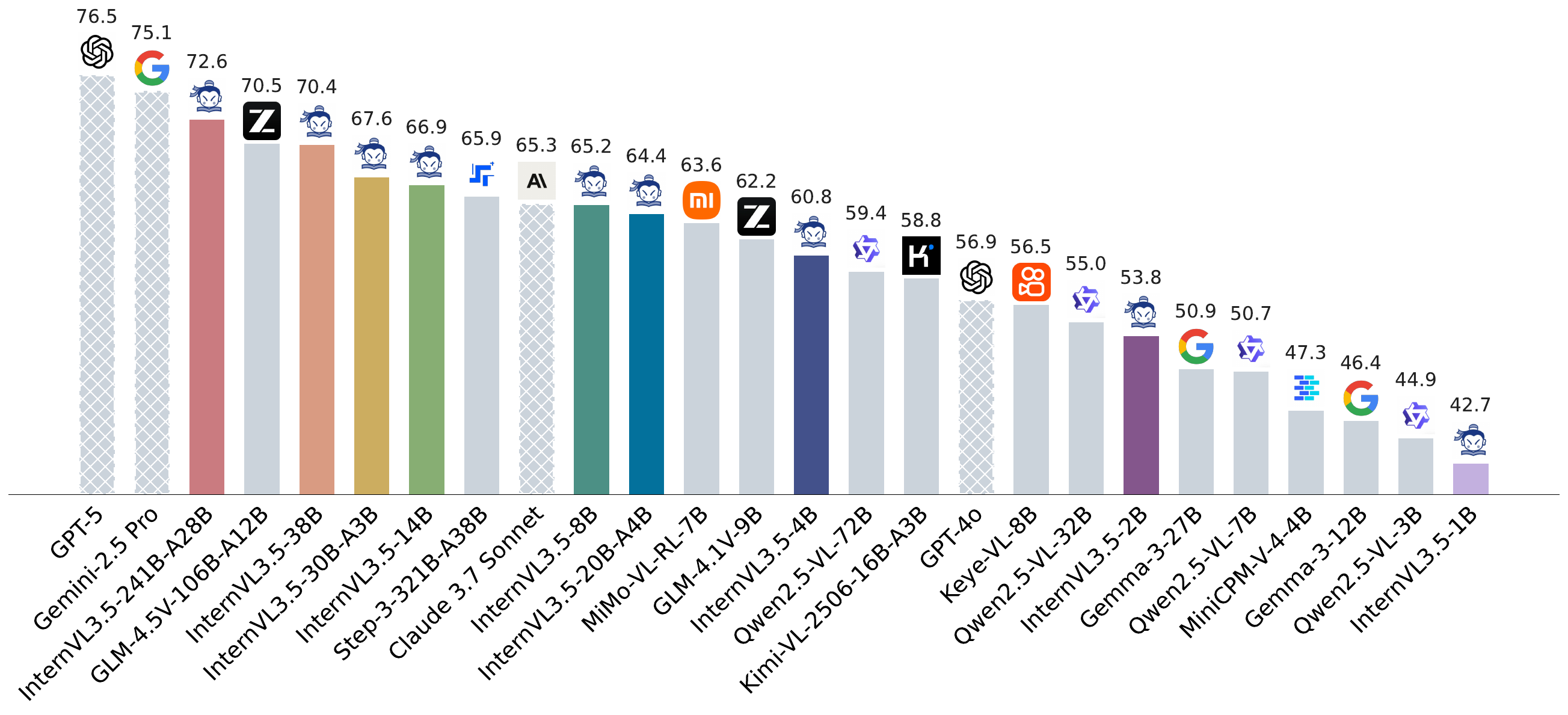}
  \caption{
  \textbf{Comparison between InternVL3.5 and leading MLLMs in general capabilities.} Hatched bars represent closed-source commercial models.
  We report average scores on a set of multimodal general, reasoning, text, and agentic benchmarks:
  MMBench v1.1 (en)~\cite{liu2023mmbench}, MMStar~\cite{chen2024mmstar}, BLINK~\cite{fu2024blink}, HallusionBench~\cite{guan2023hallusionbench}, AI2D~\cite{kembhavi2016ai2d}, OCRBench~\cite{liu2023ocrbench}, MMVet~\cite{yu2023mmvet}, MME-RealWorld (en)~\cite{zhang2024mme}, MVBench~\cite{li2024mvbench}, VideoMME~\cite{fu2024video}, MMMU~\cite{yue2023mmmu}, MathVista~\cite{lu2023mathvista}, MathVision~\cite{wang2024mathvision}, MathVerse~\cite{zhang2024mathverse}, DynaMath~\cite{zou2024dynamath}, WeMath~\cite{qiao2024wemath}, LogicVista~\cite{xiao2024logicvista}, MATH500~\cite{DBLP:conf/nips/HendrycksBKABTS21}, AIME24~\cite{aime2024}, AIME25~\cite{aime2025}, GPQA~\cite{rein2024gpqa}, MMLU-Pro~\cite{wang2024mmlu}, GAOKAO~\cite{Zhang2023gaokao}, IFEval~\cite{zhou2023instruction},
SGP-Bench~\cite{qiu2024can}, VSI-Bench~\cite{yang2024think}, ERQA~\cite{erqa}, SpaCE-10~\cite{gong2025space10}, and OmniSpatial~\cite{jia2025omnispatial}.
  }
  \label{fig:teaser}
\end{figure}

\section{Introduction}

 The recent trend of Multimodal Large Language Models (MLLMs)~\cite{qwen2.5,team2025keye_vl,hong2025glm_v_thinking} has gone beyond simple multimodal understanding and gradually focused on more general, complex, and realistic tasks such as text-related tasks~\cite{aime2024,chen2021evaluating,hendrycks2020measuring,Zhang2023gaokao,huang2023ceval,rein2024gpqa}, reasoning tasks~\cite{yang2024think, wang2024mathvision,lu2023mathvista,yue2023mmmu,he2024olympiadbench} and agentic tasks~\cite{qin2025ui, hu2025os_agents, xie2024osworld, zhang2025guimid, yang2025zerogui,  sun2025scienceboard,jia2025omnispatial,yang2024think,gong2025space10}. 
In these aspects, commercial models have created huge gaps with current open-source models, as shown in Table \ref{tab:exp-overall}.   Thereby, recent open-source efforts~\cite{meng2025mm_eureka,hong2025glm_v_thinking,team2025keye_vl,wang2025skywork_r1v2} aim to explore advanced reinforcement learning (RL) methods to mitigate the gap and pursue higher multimodal intelligence.  However, despite much effort in RL algorithms~\cite{shao2024deepseekmath,zheng2025gspo,meng2025mm_eureka,cui2025prime,cui2025entropy_mechanism,yue2025does,vl-rethinker,xi2025bmmr} and verifiers~\cite{wang2025visualprm,wang2023mathshepherd,luo2024omegaprm}, a stable, effective, and scalable reinforcement learning framework for MLLMs still remains an open problem in the community. Furthermore, the growth of multimodal capabilities, \emph{e.g.,} long visual context and high-resolution understanding~\cite{zhu2025internvl3, llava_uhd_v2, luo2024llava_hr, wang2025parameter, bai2025qwen2_5, wang2025lvbenchextremelongvideo}, often comes with ever increasing computational costs,  which have become a crucial  bottleneck of real-world applications.

In this work, we introduce InternVL3.5, an advanced family of InternVL series~\cite{chen2023internvl,chen2024internvl_1_5,chen2024internvl_2_5,zhu2025internvl3, luo2024mono_internvl, mono_internvl_v1.5, gao2024mini_internvl} with stronger capabilities in versatility, reasoning, and efficiency.  
Compared to InternVL3~\cite{zhu2025internvl3}, InternVL3.5 achieves superior performance through our proposed \textit{Cascade RL} framework, which enhances reasoning capabilities in an efficient, scalable, and stable manner. Cascade RL consists of two complementary substages: an offline RL stage~\cite{wang2024mpo}, which efficiently achieves satisfactory performance, and an online RL stage~\cite{zheng2025gspo}, which carefully refines the output distribution and further push the performance upper bound of the model. The offline stage serves as an effective warm-up, ensuring high-quality rollouts for the subsequent online stage, thereby enabling the progressive improvement of MLLM reasoning abilities.
In practice, Cascade RL demonstrates promising salability and stability, with a clear gain seen from InternVL3.5-1B to InternVL3.5-241B (Figure~\ref{fig:ablation_training}).

In addition, we equip InternVL3.5 with a much faster inference speed than its predecessor through two novel methods, namely \textit{Visual Resolution Router} (ViR) and  \textit{Decoupled Vision-Language Deployment} (DvD). In particular, ViR aims to dynamically select the best trade-off resolution of visual tokens, thus reducing the inference costs with a negligible performance sacrifice.   In practice, ViR can be efficiently integrated into InternVL3.5 with  a light training stage namely Visual Consistency Learning (ViCO).   Furthermore, DvD aims to deploy ViTs and LLMs on separate GPUs to maximize computational parallelism and hardware utilization.  These two methods can be seamlessly combined to realize the hardware-friendly implementation for InternVL3.5.

We conduct extensive experiments on public benchmarks to compare InternVL3.5 with existing MLLMs.   As shown in Figure \ref{fig:teaser}, the InternVL3.5 series consistently maintains a leading position among open source MLLMs in terms of overall score. Compared to the latest commercial model, \emph{i.e.,} GPT-5~\cite{gpt5}, InternVL3.5-241B-A28B narrows the gap to 3.9\%. In addition, our detailed ablation study demonstrates that InternVL3.5 achieves up to +16.0\% improvement in overall reasoning performance and  4.05$\times$ speed-up in inference efficiency compared to its predecessor (\ie, InternVL3~\cite{zhu2025internvl3}).  For example,  InternVL3.5-8B and InternVL3.5-241B-A28B achieve scores of 73.4 and 77.7, respectively, on the MMMU benchmark~\cite{yue2023mmmu}, showing their strong reasoning capabilities among existing open source MLLMs. 
In terms of versatility, InternVL3.5 remains competitive  against both open-source and closed-source MLLMs in text tasks,  GUI tasks, embodied tasks, SVG-based understanding and generation, \textit{etc}. For example,  InternVL3.5-30B-A3B and InternVL3.5-241B-A28B surpass the latest open-source MLLM (Step-3~\cite{wang2025step}) by +2.0 and +8.4 in text tasks, respectively.

In summary, our contributions  include three folds:

(1) We release InternVL3.5, the latest family of the InternVL series with advanced reasoning abilities, powerful  versatility, and promising efficiency. InternVL3.5 comprises various model scales (from 1B to 241B) with both dense and mixture-of-experts (MoE) models. All of our models and codes are publicly  released.

(2) We propose three innovative designs in InternVL3.5, including cascade reinforcement learning (Cascade RL), visual resolution router (ViR) and decoupled vision-language deployment (DvD). These technologies significantly improve the capabilities and efficiency of InternVL3.5, providing practical tips to the community.

(3)  We conduct extensive experiments and demonstrate that InternVL3.5 exhibits leading performance among open-source MLLMs~\cite{hong2025glm_v_thinking,team2025keye_vl,coreteam2025mimovltechnicalreport,yao2024minicpm,wang2024qwen2vl}. Compared to the latest commercial model, \emph{i.e.,} GPT-5~\cite{gpt5}, InternVL3.5 even achieves slightly better results on general multimodal capabilities. We believe our approach and open source will further advance the community.

\section{InternVL3.5}

\begin{figure}[t] 
  \centering 
  \includegraphics[width=1.0\linewidth]{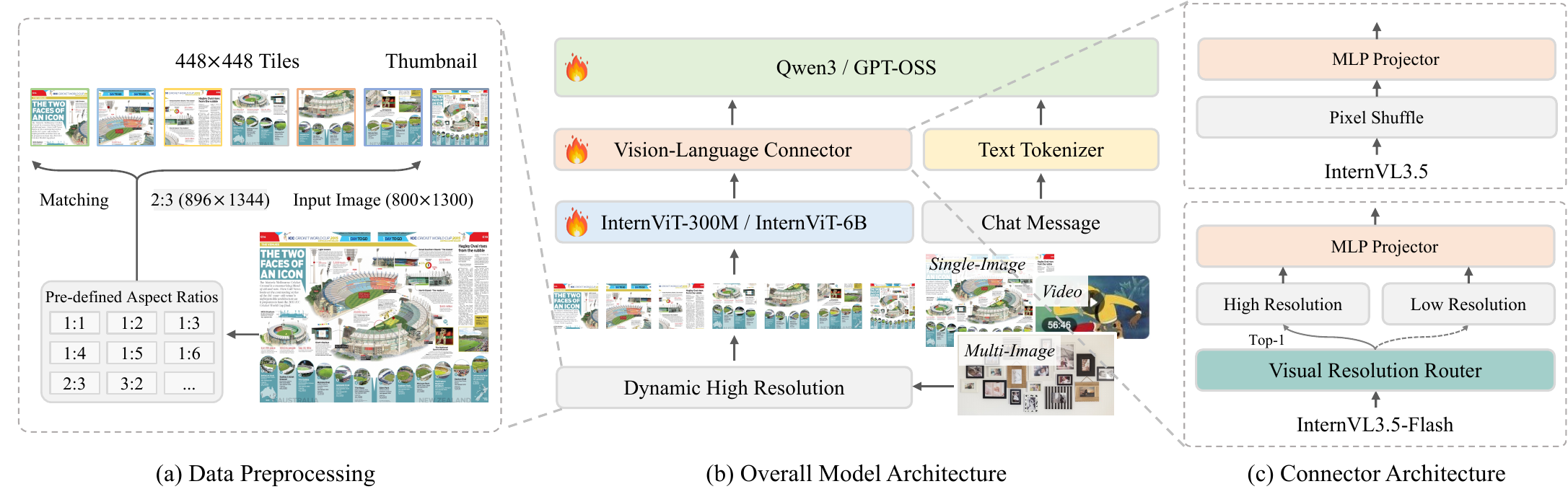}
  \caption{
    \textbf{Overall architecture.}
    InternVL3.5 adopts the ``ViT–MLP–LLM'' paradigm as in previous versions. Building upon InternVL3.5, we further introduce InternVL3.5-Flash, which is extended with an additional visual resolution router (ViR) to dynamically select the appropriate compression rate (\eg, $\frac{1}{4}$ or $\frac{1}{16}$) for each image patch. Unlike Dynamic High Resolution which only splits image patches from the perspective of image width and height, our proposed ViR further introduces adaptivity from the perspective of semantic content.
  }
  \label{fig:architecture}
\end{figure}

\begin{figure}[!t] 
  \centering 
  \includegraphics[width=1.0\linewidth]{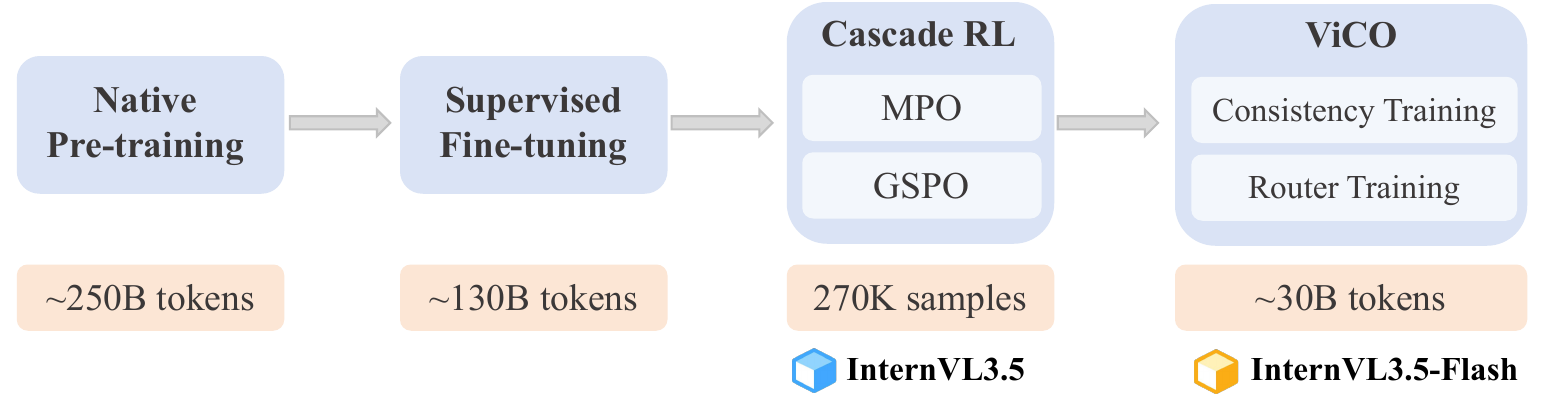}
  \caption{\textbf{Training recipes of InternVL3.5.} InternVL3.5  consists of three training stages: (1) native pre-training for vision-language alignment, (2) supervised fine-tuning for adaptation to downstream tasks, (3) Cascade RL for improvement on reasoning capabilities.  InternVL3.5-Flash is an efficient version of InternVL3.5, which further integrates a visual resolution router (ViR) through consistency training and router training.}
  \label{fig:training_recipes}
\end{figure}

Compared to its predecessors, the InternVL3.5 series achieves superior performance and faster inference.
In Section~\ref{sec:method-model-arch}, we introduce the model architectures of InternVL3.5 and InternVL3.5-Flash.
For InternVL3.5-Flash, we further incorporate a Visual Resolution Router (ViR) module that dynamically selects the minimal resolution of visual tokens, achieving better inference efficiency.
Section~\ref{sec:method-pre-training} and Section~\ref{sec:method-post-training} describe the pre-training and post-training procedures of InternVL3.5, respectively.
The details of our proposed Cascade Reinforcement Learning (Cascade RL) and Visual Consistency Learning (ViCO) methods are elaborated in Section~\ref{sec:method-post-training}.
In Section~\ref{sec:method-tts}, we present the test-time scaling approach used to further improve model performance.
Finally, in Section~\ref{sec:method-infra}, we describe the training and inference infrastructure supporting InternVL3.5, including implementation details of the Decoupled Vision-Language Deployment (DvD) framework. The overall architecture is shown in Figure~\ref{fig:architecture}, and the training recipes are illustrated in Figure~\ref{fig:training_recipes}.

\begin{table}[t]
\centering
\small

\begin{tabular}{lllrrr}
\toprule
\multirow{2}{*}{\textbf{Model}} & \multirow{2}{*}{\textbf{Vision Encoder}} & \multirow{2}{*}{\textbf{Language Model}} & \multicolumn{3}{c}{\textbf{\#Param}} \\
    \cmidrule(lr){4-6} %
    & & & \textbf{Vision} & \textbf{Language} & \textbf{Total} \\
\midrule
\multicolumn{6}{c}{\textit{Dense Models}}                                                                                                                             \\
\midrule
InternVL3.5-1B        & InternViT-300M & Qwen3-0.6B              & 0.3B                & 0.8B                  & 1.1B             \\
InternVL3.5-2B        & InternViT-300M & Qwen3-1.7B              & 0.3B                & 2.0B                 & 2.3B             \\
InternVL3.5-4B        & InternViT-300M & Qwen3-4B                & 0.3B                & 4.4B                 & 4.7B             \\
InternVL3.5-8B        & InternViT-300M & Qwen3-8B                & 0.3B                & 8.2B                 & 8.5B             \\
InternVL3.5-14B       & InternViT-300M & Qwen3-14B               & 0.3B                & 14.8B                & 15.1B            \\
InternVL3.5-38B       & InternViT-6B   & Qwen3-32B               & 5.5B               & 32.8B                & 38.4B            \\
\midrule
\multicolumn{6}{c}{\textit{MoE Models}}                                                                                                                 \\
\midrule
InternVL3.5-20B-A4B   & InternViT-300M & GPT-OSS-20B             & 0.3B                & 20.9B                & 21.2B\ \ \ (A4B)            \\
InternVL3.5-30B-A3B   & InternViT-300M & Qwen3-30B-A3B           & 0.3B                & 30.5B                & 30.8B\ \ \ (A3B)            \\
InternVL3.5-241B-A28B & InternViT-6B   & Qwen3-235B-A22B         & 5.5B               &  235.1B               & 240.7B (A28B)          \\ \bottomrule
\end{tabular}
\vspace{2mm}
\caption{
    \textbf{Pre-trained models used in the InternVL3.5 series}. ``A'' denotes the number of activated parameters. 
}
\label{tab:model-config}

\vspace{-3mm}

\end{table}

\subsection{Model Architecture}
\label{sec:method-model-arch}

\textbf{InternVL3.5}. We follow the ``ViT–MLP–LLM'' paradigm adopted in previous versions of InternVL~\cite{chen2024internvl_1_5,wang2024mpo,chen2024internvl_2_5,zhu2025internvl3,gao2024mini_internvl}.
As shown in Table~\ref{tab:model-config}, we initialize the language model using the Qwen3 series~\cite{yang2025qwen3} and GPT-OSS~\cite{gpt-oss}, and the vision encoder using InternViT-300M and InternViT-6B~\cite{chen2023internvl}.
The Dynamic High Resolution strategy introduced in InternVL1.5~\cite{chen2024internvl_1_5} is also retained in our design.

\textbf{InternVL3.5-Flash}.
Compared to InternVL3.5, InternVL3.5-Flash further integrates the \textit{Visual Resolution Router} (ViR), thus yielding a series of  efficient variants  suitable for  resource-constrained scenarios. 
Specifically, in InternVL3.5, each image patch is initially represented as 1024 visual tokens for the vision encoder, which are then compressed into 256 tokens via a pixel shuffle module before being passed to the Large Language Model (LLM).
In InternVL3.5-Flash, as shown in Figure~\ref{fig:architecture}, an additional pixel shuffle module with a higher compression rate is included, enabling compression of visual tokens down to 64 tokens.
For each patch, the patch router determines the appropriate compression rate by assessing its semantic richness, and routes it to the corresponding pixel shuffle module accordingly.
Benefiting from this patch-aware compression mechanism, InternVL3.5-Flash is able to reduce the number of visual tokens by 50\% while maintaining nearly 100\% of the performance of InternVL3.5, as shown in Section~\ref{sec:exp-ablation}.

\subsection{Pre-Training}
\label{sec:method-pre-training}

\textbf{Training Objective}. 
During the pre-training stage, we update all model parameters jointly using the combination of large-scale text and multimodal corpora. Specifically, given an arbitrary training sample consisting of a multimodal token sequence $\mathbf{x}=\left(x_1, x_2, \ldots, x_L\right)$, the next token prediction (NTP) loss~\cite{radford2019language} is calculated on each text token as follows:
\begin{equation}
    \mathcal{L}_{i}=-\log p_\theta\left(x_i \mid x_1, \ldots, x_{i-1}\right),
\end{equation}
where $x_i$ is the predicted token and  prefix tokens in $\{x_1, x_2, \ldots, x_{i-1}\}$ can be either  text tokens or  image tokens. In particular, for conversation samples, only response tokens  are included for the loss calculation.
Additionally, to mitigate bias toward either longer or shorter responses during training, we adopt the square averaging~\cite{chen2024internvl_2_5} to reweight the NTP loss  as follows:
\begin{equation}
\mathcal{L}_{i}^{'} = \frac{w_i}{\sum_j w_j} \cdot \mathcal{L}_i, \quad w_i =
\frac{1}{N^{0.5}},
\end{equation}
where $N$ denotes the number of tokens in the training sample on which the loss needs to be calculated. The random JPEG compression~\cite{chen2024internvl_2_5} is also included to enhance the model's real-world performance.

\textbf{Data}. The pre-training corpora can be classified into two categories:
(1) Multimodal data: this subset of data is mainly sourced from the training corpora of InternVL3~\cite{zhu2025internvl3}, covering a diverse range of domains such as image captioning, general question answering, mathematics, scientific disciplines, charts, optical character recognition (OCR), knowledge grounding, document understanding, multi-turn dialogue, and medical data.
(2) Text-only data: this part of data is constructed based on the training corpora of InternLM series~\cite{2023internlm,cai2024internlm2} and is further augmented with open-source datasets~\cite{benallal2024smollmcorpus,acemath2024,scp116k,muennighoff2025s1}.
The pre-training corpora contains approximately 116M samples, corresponding to about 250B tokens. 
The ratio between text-only and multimodal data is approximately $1:2.5$.
The maximum sequence length is set to 32K tokens to adapt long-context understanding and  reasoning.

\subsection{Post-Training}
\label{sec:method-post-training}

After the pre-training stage, we adopt a three-stage post-training strategy comprising:
(1) \textit{Supervised Fine-Tuning (SFT)}, which maintains the same training objective as pre-training but leverages higher-quality conversation data to further enhance the model's capabilities.
(2) \textit{Cascade Reinforcement Learning (Cascade RL)}, which combines the benefits of offline and online RL methods to facilitate the reasoning capabilities. 
 (3) \textit{Visual Consistency Learning (ViCO)}, which aims to integrate visual resolution router (ViR) into InternVL3.5 to construct InternVL3.5-Flash, by   minimizing the output divergence  of different visual compression rates.

\textbf{Supervised Fine-Tuning}.
During the SFT phase, we adopt the same objective as in the pre-training stage and use the  square averaging strategy~\cite{chen2024internvl_2_5} to calculate the final loss.  In this stage, the context window is set to 32K tokens to adapt long-context information. Compared to InternVL3, the SFT stage of InternVL3.5 contains  more high-quality and  diverse training data derived from three sources: 
(1) Instruction-following data from InternVL3, which are reused to preserve broad coverage of vision–language tasks. 
(2) Multimodal reasoning data in the ``Thinking'' mode, which are included to instill long-thinking capabilities in the model. To construct such data, we leverage a large-scale reasoning model to sample rollouts with detailed reasoning processes. In addition to validating whether answers are factually correct, we implement strict filtering measures for the reasoning processes themselves: this includes evaluating how clear the thinking is, weeding out redundancy, and ensuring that formatting remains consistent. The questions in these datasets cover various expert domains, such as mathematics and scientific disciplines, thereby strengthening performance on different reasoning tasks. 
(3) Capability-expansion datasets, which endow InternVL3.5 with new skills, including GUI-based interaction, embodied interaction, and scalable vector graphics (SVG) understanding and generation.

\textbf{Cascade Reinforcement Learning}.
Compared to Pre-training and Supervised Fine-tuning (SFT), the core advantage of RL lies in its ability to introduce negative samples, which prune low-quality regions in the model’s output space and thereby enhance the overall response quality.
As a derivative of the PPO algorithm~\cite{schulman2017ppo}, DPO~\cite{rafailov2024dpo} enables training based on existing rollouts, which we also regard as a form of offline RL.
Offline RL algorithms~\cite{rafailov2024dpo,wang2024mpo} often offer a higher training efficiency, but their performance ceiling is generally lower compared to online RL methods. In contrast, despite the effectiveness of online RL algorithms~\cite{shao2024deepseekmath,schulman2017ppo,zheng2023ppo_max,yu2025dapo}, they are often computationally expensive and time-consuming. In this work, we propose Cascade RL, which aims to combine the benefits of offline RL and online RL to progressively facilitate the post-training of MLLMs in an efficient manner.
Specifically, we first fine-tune the model using an offline RL algorithm~\cite{wang2024mpo} as an efficient warm-up stage to reach satisfied results, which can guarantee high-quality rollouts for the latter stage. 
Subsequently, we employ an online RL algorithm~\cite{zheng2025gspo} to further refine the output distribution based on rollouts generated by the model itself. 
Compared to the single offline or online RL stage, our cascaded RL achieves significant performance improvements at a fraction of the GPU time cost.

During the offline RL stage, we employ mixed preference optimization (MPO)~\cite{wang2024mpo} to fine-tune the model. Specifically, the training objective of MPO is a combination of preference loss $\mathcal{L}_{p}$, quality loss $\mathcal{L}_{q}$, and generation loss $\mathcal{L}_{g}$, which can be formulated as follows:
\begin{equation}
    \mathcal{L}_{\text{MPO}}=
    w_{p} \mathcal{L}_{p}
    +
    w_{q} \mathcal{L}_{q}
    +
    w_{g} \mathcal{L}_{g}
    ,
    \label{eqn:final_loss}
\end{equation}
where $w_{*}$ represents the weight assigned to each loss component.
The DPO loss~\cite{rafailov2024dpo}, BCO loss~\cite{jung2024bco}, and LM loss~\cite{brown2020gpt} serve as the preference loss, quality loss, and generation loss, respectively.

During the online RL stage, we employ GSPO~\cite{zheng2025gspo}, without reference model constraints, as our online RL algorithm, which we find more effective in training both dense and mixture-of-experts (MoE) models. Similar to GRPO~\cite{shao2024deepseekmath}, the advantage is defined as the normalized reward across responses sampled from the same query:
\begin{equation}
    \widehat{A}_i=\frac{r\left(x, y_i\right)-\operatorname{mean}\left(\left\{r\left(x, y_i\right)\right\}_{i=1}^G\right)}{\operatorname{std}\left(\left\{r\left(x, y_i\right)\right\}_{i=1}^G\right)},
\end{equation}
where $y_i$ is the $i$-th response generated for the query $x$, $G$ is the total number of generated responses to the query, and $r\left(x, y_i\right)$ denotes the reward for this response.
The training objective of GSPO is given by:
\begin{equation}
    \mathcal{L}_{\mathrm{GSPO}}(\theta)=\mathbb{E}_{x \sim \mathcal{D},\left\{y_i\right\}_{i=1}^G \sim \pi_{\theta \text { old }}(\cdot \mid x)}\left[\frac{1}{G} \sum_{i=1}^G \min \left(s_i(\theta) \widehat{A}_i, \operatorname{clip}\left(s_i(\theta), 1-\varepsilon, 1+\varepsilon\right) \widehat{A}_i\right)\right],
\end{equation}
where the importance sampling ratio is defined as the geometric mean of the per-token ratios:
\begin{equation}
    s_i(\theta)=\left(\frac{\pi_\theta\left(y_i \mid x\right)}{\pi_{\theta_{\text {old }}}\left(y_i \mid x\right)}\right)^{\frac{1}{\left|y_i\right|}}=\exp \left(\frac{1}{\left|y_i\right|} \sum_{t=1}^{\left|y_i\right|} \log \frac{\pi_\theta\left(y_{i, t} \mid x, y_{i,<t}\right)}{\pi_{\theta_{\text {old }}}\left(y_{i, t} \mid x, y_{i,<t}\right)}\right),
\end{equation}
where $\pi_\theta\left(y_{i} \mid x, y_{i,<t}\right)$ and $\pi_\theta\left(y_{i, t} \mid x, y_{i,<t}\right)$ denote the generation probability of response $y_i$ and token $y_{i,t}$ under the policy model with parameters $\theta$, respectively.

Compared to directly training the model with a single RL paradigm, Cascade RL offers the following advantages:
(1)~\textit{Better training stability}:
In the offline RL stage, the rollout collection and parameter updates are decoupled, effectively mitigating issues such as reward hacking.
During the online RL stage, we empirically observe that stronger models exhibit more stable and robust training dynamics.
As a result, the performance gains achieved in the MPO stage further enhance the stability of the GSPO stage and reduce sensitivity to the algorithm.
(2)~\textit{Improved training efficiency}:
In the MPO stage, rollouts can be shared across different models, amortizing the sampling cost typically incurred during online RL.
(3)~\textit{Higher performance ceiling}:
Moreover, as shown in Section~\ref{sec:exp-ablation}, models fine-tuned with MPO take fewer training steps to achieve higher performance in the subsequent online RL phase, further reducing training overhead.

\textbf{Visual Consistency Learning}.
We further include ViCO as an additional training stage to integrate the visual resolution router (ViR) into InternVL3.5, thereby reducing the inference cost of InternVL3.5. The obtained efficient version of InternVL3.5 are termed as \textit{InternVL3.5-Flash}.
In particular, ViCO comprises two stages:

(1) \textit{Consistency training}:
In this stage, the entire model is trained to minimize the divergence between response distributions conditioned on visual tokens with different compression rates.
In practice, we introduce an extra reference model, which is frozen and initialized with InternVL3.5.
Given a sample, each image patch is represented as either 256 or 64 tokens, and the training objective is defined as follows:
\begin{equation}
\mathcal{L}_\text{ViCO} =
\mathbb{E}_{\xi \sim \mathcal{R}} \Bigg[
\frac{1}{N} \sum_{i=1}^{N} \mathrm{KL} \Big(
\pi_{\theta_{ref}}\left(y_i \mid y_{<i}, I\right) \;\Big\|\;
\pi_{\theta_{policy}}\left(y_i \mid y_{<i}, I_\xi\right)
\Big)
\Bigg],
\end{equation}
where $\mathrm{KL}$ denotes the KL divergence and $\xi$ denotes the compression rate, which is uniformly sampled from $\{\frac{1}{4},\frac{1}{16}\}$. The image $I_\xi$ is represented as 256 tokens when $\xi=\frac{1}{4}$ and 64 tokens when $\xi=\frac{1}{16}$. We note that the reference model always performs inference with $\xi=\frac{1}{4}$.

(2) \textit{ Router training}:
This stage aims to train the ViR to select an appropriate trade-off resolution for different inputs.
ViR is formulated as a binary classifier and trained using standard cross-entropy loss.
To construct the route targets, we first compute the KL divergence between the model outputs conditioned on uncompressed visual tokens (\ie, 256 tokens per patch) and those conditioned on compressed visual tokens (\ie, 64 tokens per patch).
During this stage, the main MLLM (ViT, MLP and LLM) is kept frozen, and only the ViR is trained.
Specifically, we first compute the loss ratio for each patch:
\begin{equation}
r_i = \frac{\mathcal{L}_\text{ViCO}\big(y_i \mid I_{\frac{1}{16}}\big)}{\mathcal{L}_\text{ViCO}\big(y_i \mid I_{\frac{1}{4}}\big)},
\end{equation}
which quantifies the relative increase in loss caused by compressing the visual tokens. Based on this ratio, the binary ground-truth label for the patch router is defined as:
\begin{equation}
y_i^\text{router} =
\begin{cases}
0, & r_i < \tau \; \text{(compression has negligible impact)} \\
1, & r_i \ge \tau \; \text{(compression has significant impact)},
\end{cases}
\end{equation}
where $y_i^{\text{router}}=0$ and $y_i^{\text{router}}=1$  indicate that the compression rate $\xi$ is set to $\tfrac{1}{16}$ and $\tfrac{1}{4}$, respectively.
During training, we store the historical $r_i$ values of a sliding window,  and  $\tau$ is a dynamical threshold computed from the k-\textit{th} percentile of historical $r_i$ values.
In practice, the target distribution is balanced.
During the consistency training stage, all patches of the same image are represented with a random compression rate, in order to ensure that the model retains its capability when no compression is applied. 
As shown in Section~\ref{sec:exp-ablation}, InternVL3.5-Flash reduces 50\% of the visual tokens while maintaining nearly 100\% of the original performance.

\textbf{Data}.
For the supervised fine-tuning (SFT) stage, the datasets comprise approximately 56 million samples, which corresponds to around 130 billion tokens. The proportion of text-only data to multimodal data is roughly 1:3.5.
For the cascade reinforcement learning stage, we use MMPR-v1.2~\cite{wang2024mpo} as the training data for offline RL, which contains about 200K sample pairs. Based on MMPR-v1.2, we compute the accuracy of each query using the provided rollouts and select those whose model accuracy falls between 0.2 and 0.8 for online RL. We further extend the dataset with recent multimodal datasets~\cite{meng2025mm_eureka,yang2025r1_onevision,wang2025mv_math,liu2024cmmmath,deepscaler2025} to enhance diversity. The resulting dataset, termed MMPR-Tiny, consists of approximately 70K queries. We directly reuse the rollouts from MMPR-v1.2 for both offline RL and data filtering in online RL, thereby reducing the cost of sampling additional rollouts.

For the ViCO stage, we primarily leverage datasets identical to the SFT stage during consistency training, ensuring that the model retains its original performance. During router training, we use a subset of the SFT data, primarily composed of OCR and VQA examples, which are rich in visual information and sometimes require high-resolution understanding. This enables the resolution router to learn how to dynamically decide whether each image patch can be compressed based on the visual information.

\subsection{Test-Time Scaling}
\label{sec:method-tts}

Test-time scaling (TTS) has been empirically demonstrated as an effective approach to enhance the reasoning capabilities of LLMs and MLLMs, particularly for complex tasks that require multi-step inference~\cite{wang2025visualprm,snell2024test_time_scaling_efficient,zhang2025stair,luo2024omegaprm,lightman2023prm800k}.
In this work, we implement a comprehensive test-time scaling approach that simultaneously improves reasoning depth (\ie, deep thinking) and breadth (\ie, parallel thinking).
We note that unless otherwise specified, the experimental results reported in Section~\ref{sec:exp} are obtained without applying TTS. Thus far, we have only applied TTS to reasoning benchmarks, since we found that the model already exhibits strong perception and understanding capabilities, and initiating TTS yields no significant improvement.

\textbf{Deep Thinking}. By activating the Thinking mode, we guide the model to deliberately engage in step-by-step reasoning (\textit{i.e.,} decomposing complex problems into logical steps and validating intermediate conclusions) prior to generating the final answer. This approach systematically improves the logical structure of solutions for complex problems, particularly those requiring multi-step inference, and enhances reasoning depth.

\textbf{Parallel Thinking}. Following InternVL3, for reasoning tasks, we adopt the Best-of-N (BoN) strategy by employing VisualPRM-v1.1~\cite{wang2025visualprm} as the critic model to select the optimal response from multiple reasoning candidates.
This approach improves the reasoning breadth.

\subsection{Infrastructure}
\label{sec:method-infra}

\textbf{Training Framework}.
The model training is conducted mainly based on the XTuner framework~\cite{2023xtuner}, which incorporates a series of optimization strategies tailored for LLM and MoE training. These include fully shared data parallelism (FSDP)~\cite{zhao2023fsdp} to partition model parameters across GPUs, data packing~\cite{chen2024internvl_2_5} to reduce padding tokens while balancing the token computation load across ranks for improved training efficiency, FP8 training based on DeepGEMM~\cite{liu2024deepseekv3} and liger-kernel's fused cross-entropy operator~\cite{hsu2025ligerkernel} to accelerate the training process, FlashAttention-3~\cite{dao2022flashattention,dao2023flashattention2} to support packed inputs and speed up attention computation, and the TMA-Adaptive FP8 Grouped GEMM kernel~\cite{suzhongling2025tmaadaptive} to optimize the training of MoE models.
For the online stage, we use verl~\cite{sheng2024verl} as our codebase.
For InternVL3.5-20B-A4B, we implement an accelerated version of  window attention with sink in GPT-OSS-20B through Triton~\cite{tillet2019triton}.

\begin{figure}[t] 
  \centering 
  \includegraphics[width=1.0\linewidth]{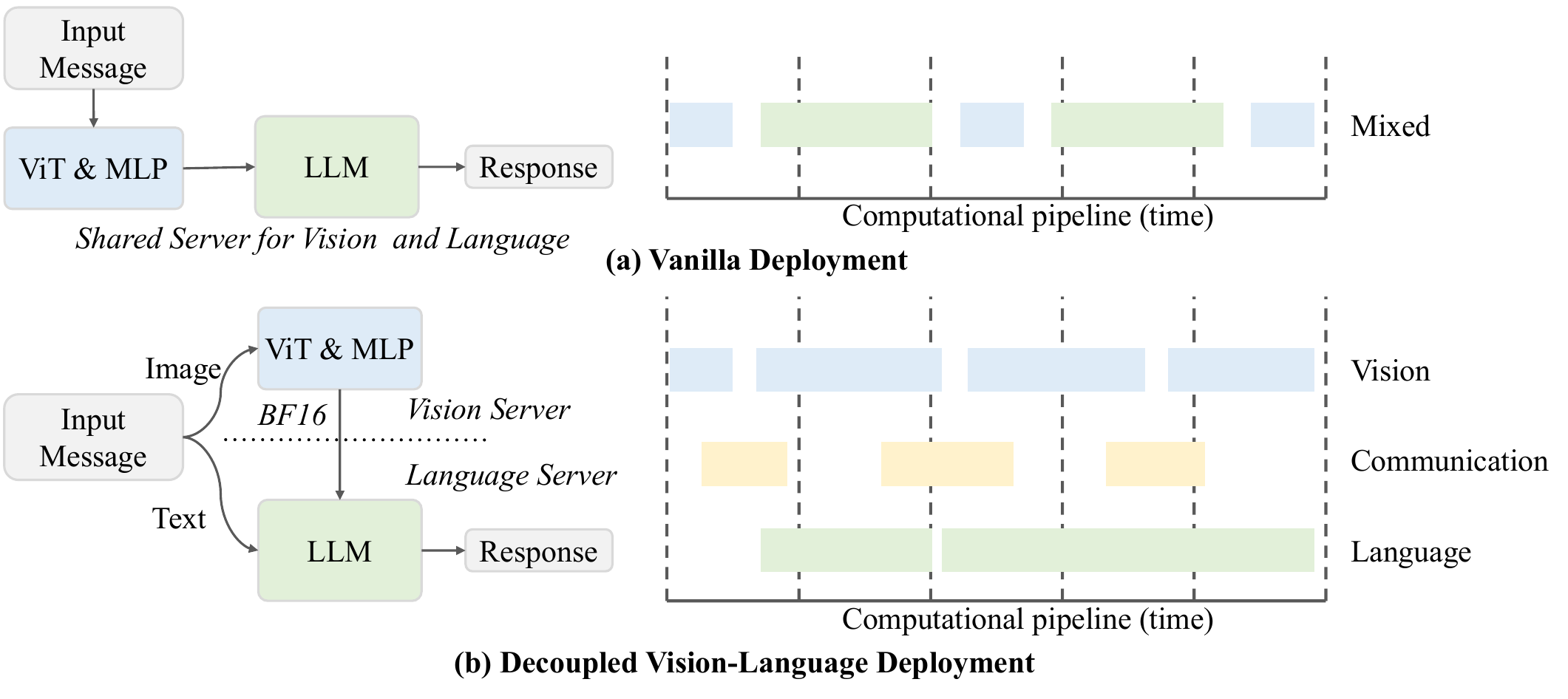}
  \caption{\textbf{Overview of Decoupled Vision-Language Deployment.} DvD decouples the vision and language models and deploys them on separate servers. The right side shows a time-consumption trace of the pipeline. \textbf{(a)} In the original deployment, the ViT, MLP, and LLM are executed sequentially. Given their substantial differences in size and computation patterns, this serial design significantly slows down inference. \textbf{(b)} With DvD, the inference of the ViT and the LLM is performed in parallel and asynchronously. Thus, ViT's computations can overlap with the LLM’s prefilling and decoding, reducing resource conflicts and improving inference speed.}
  \label{fig:DvD}
\end{figure}

\textbf{Decoupled Vision-Language Deployment}.
In multimodal inference, the vision encoder and language model have distinct computational characteristics. The vision encoder that transforms images into semantic features is highly parallelizable and does not rely on long-term history state.  In contrast,  the language model adopts the inference in an autoregressive manner, which requires previous states to compute the next one. This sequential property makes the language part more sensitive to memory bandwidth and latency. 
When MLLMs are deployed online at scale, the vision and language models often block each other, thus incurring additional inference cost. This effect becomes more pronounced with larger vision models or higher-resolution images.

As shown in Figure~\ref{fig:DvD}, we propose Decoupled Vision-Language Deployment (DvD) to address this issue by separating vision and language processing, with a particular focus on optimizing the prefilling stage. The vision subsystem batches and processes images to produce compact feature embeddings, which are then transmitted to the language subsystem for fusion with the text context prior to decoding. This separation alleviates blocking and brings multimodal prefilling performance closer to that of pure language models.
In our system implementation, the ViT and MLP (and ViR for InternVL3.5-Flash) are deployed on the vision server, while the language server executes only the LLM. The communication is unidirectional, transmitting BF16 visual features over TCP, with RDMA optionally employed to achieve a higher transmission speed. Vision processing, feature transmission, and language processing are organized into an asynchronous three-stage pipeline, enabling overlapped execution and minimizing pipeline stalls.

DvD increases GPU utilization and processing efficiency on the vision side, while enabling the language server to focus exclusively on the LLM’s prefilling and decoding without being blocked by vision computation. This design leads to improved throughput and responsiveness. Moreover, the architecture supports independent hardware cost optimization for the vision and language modules, and facilitates the seamless integration of new modules without requiring modifications to the language server deployment.

\section{Experiments}
\label{sec:exp}

Here, we first compare the overall performance of InternVL3.5 with recent leading multimodal large language models (MLLMs) in Section~\ref{sec:exp-overall}.
Subsequently, we evaluate our models in various domains, including multimodal reasoning and mathematics (Section~\ref{sec:exp-reasoning}), optical character recognition (OCR), chart and document understanding (Section~\ref{sec:exp-ocr}), multi-image understanding (Section~\ref{sec:exp-multi-image}), real-world comprehension (Section~\ref{sec:exp-real-world-comprehension}), comprehensive multimodal evaluation (Section~\ref{sec:exp-comprehensive-eval}), multimodal hallucination evaluation (Section~\ref{sec:exp-hallucination}), visual grounding (Section~\ref{sec:exp-grounding}), multimodal multilingual understanding (Section~\ref{sec:exp-multilingual}), video understanding (Section~\ref{sec:exp-video}), GUI (Section~\ref{sec:exp-gui}), embodied (Section~\ref{sec:exp-embody}), and SVG (Section~\ref{sec:exp-svg}) tasks, most of which were tested using VLMEvalKit~\cite{duan2024vlmevalkit}.
Additionally, the evaluation of the language capabilities of InternVL3.5 is presented in Section~\ref{sec:exp-language}.
Finally, we ablate newly proposed designs in InternVL3.5, including the Cascade Reinforcement Learning, Visual Resolution Router, and Decoupled Vision-Language Deployment (Section~\ref{sec:exp-ablation}).

\begin{table}[htbp]
\newcommand{\tablehead}[1]{\textbf{#1}}

\newcommand{\mycmidrulecolumntwo}{
    \noalign{\vspace{-0.8mm}}
    \cmidrule{2-10}
    \noalign{\vspace{-1mm}}
}

\newcommand{\reproduce}{*}
\newcommand{\fromglm}{$^\dag$}
\newcommand{\fromoc}{$^\ddag$}
\newcommand{\tocheck}{\todo{?}}
\newcommand{\original}{=}

\renewcommand{\original}{ }

\centering
\tiny
\begin{threeparttable}
\renewcommand{\arraystretch}{2.8}
\resizebox{\textwidth}{!}{%
\begin{tabular}{>{\fontsize{8pt}{10pt}\selectfont}l  >{\fontsize{8pt}{10pt}\selectfont}l  >{\cellcolor{myblue}\fontsize{8pt}{10pt}\selectfont}c  >{\cellcolor{myblue}\fontsize{8pt}{10pt}\selectfont}c >{\fontsize{8pt}{10pt}\selectfont}c >{\fontsize{8pt}{10pt}\selectfont}c >{\fontsize{8pt}{10pt}\selectfont}c >{\fontsize{8pt}{10pt}\selectfont}c >{\fontsize{8pt}{10pt}\selectfont}c >{\fontsize{8pt}{10pt}\selectfont}c >{\fontsize{8pt}{10pt}\selectfont}c} %
\mytoprule

\scalebox{1.5}{\textbf{Task}}  & \scalebox{1.5}{\textbf{Benchmark}}  & 
\tablehead{\scalebox{1.3}{InternVL3.5}} & 
\tablehead{\scalebox{1.3}{InternVL3.5}} & \tablehead{\scalebox{1.3}{GLM-4.1V}} & \tablehead{\scalebox{1.3}{Kimi-VL-2506}} &  
\tablehead{\scalebox{1.3}{Qwen2.5-VL}} & \tablehead{\scalebox{1.3}{GLM-4.5V}} & 
\tablehead{\scalebox{1.3}{Step-3}} & \tablehead{\scalebox{1.3}{GPT-5}}  \\ %

\mymidrule
\scalebox{1.5}{Size} & \scalebox{1.5}{--} & \scalebox{1.5}{30B-A3B} & \scalebox{1.5}{241B-A28B} & \scalebox{1.5}{9B} & \scalebox{1.5}{16B-A3B} & \scalebox{1.5}{72B} & \scalebox{1.5}{106B-A12B} & \scalebox{1.5}{321B-A38B} & \scalebox{1.5}{--} \\

\mymidrule
\multirow{14}{*}{\scalebox{1.5}{General}} & \scalebox{1.5}{MMStar} & \scalebox{1.5}{72.0\original} & \scalebox{1.5}{77.9\original} & \scalebox{1.5}{72.9\original} & \scalebox{1.5}{70.4\original} & \scalebox{1.5}{70.8\original} & \scalebox{1.5}{75.3\original} & \scalebox{1.5}{69.0\reproduce} & \scalebox{1.5}{75.7\reproduce} \\
& \scalebox{1.5}{MMVet} & \scalebox{1.5}{85.5\original} & \scalebox{1.5}{81.2\original} & \scalebox{1.5}{66.4\reproduce} & \scalebox{1.5}{78.1\original} & \scalebox{1.5}{76.2\original} & \scalebox{1.5}{75.2\reproduce} & \scalebox{1.5}{79.4\reproduce} & \scalebox{1.5}{77.6\reproduce}  \\
& \scalebox{1.5}{MMBench V1.1 (en)} & \scalebox{1.5}{84.8\original} & \scalebox{1.5}{87.4\original} & \scalebox{1.5}{85.8\original} & \scalebox{1.5}{84.4\original} & \scalebox{1.5}{88.4\original} & \scalebox{1.5}{88.2\original} & \scalebox{1.5}{81.1\reproduce} & \scalebox{1.5}{88.6\reproduce}  \\
& \scalebox{1.5}{MTVQA} & \scalebox{1.5}{33.7\original} & \scalebox{1.5}{39.3\original} & \scalebox{1.5}{25.5\reproduce} & \scalebox{1.5}{27.2\reproduce} & \scalebox{1.5}{31.7\original} & \scalebox{1.5}{30.5\reproduce} & \scalebox{1.5}{30.6\reproduce} & \scalebox{1.5}{33.1\reproduce}  \\
& \scalebox{1.5}{AI2D (w/ mask)} & \scalebox{1.5}{86.8\original} & \scalebox{1.5}{87.3\original} & \scalebox{1.5}{87.9\original} & \scalebox{1.5}{81.9\fromglm} & \scalebox{1.5}{88.7\original} & \scalebox{1.5}{88.1\original} & \scalebox{1.5}{83.7\reproduce} & \scalebox{1.5}{89.5\reproduce}  \\
& \scalebox{1.5}{OCRBench} & \scalebox{1.5}{88.0\original} & \scalebox{1.5}{90.7\original} & \scalebox{1.5}{84.2\original} & \scalebox{1.5}{86.9\original} & \scalebox{1.5}{88.5\original} & \scalebox{1.5}{87.2\original} & \scalebox{1.5}{83.7\reproduce} & \scalebox{1.5}{80.7\reproduce}  \\
& \scalebox{1.5}{WildVision} & \scalebox{1.5}{75.8\original} & \scalebox{1.5}{82.8\original} & \scalebox{1.5}{74.0\reproduce} & \scalebox{1.5}{64.8\reproduce} & \scalebox{1.5}{78.6\reproduce} & \scalebox{1.5}{79.0\reproduce} & \scalebox{1.5}{89.4\reproduce} & \scalebox{1.5}{77.4\reproduce}   \\
& \scalebox{1.5}{MME-RealWorld (en)} & \scalebox{1.5}{64.8\original} & \scalebox{1.5}{65.1\original} & \scalebox{1.5}{61.7\reproduce} & \scalebox{1.5}{54.5\reproduce} & \scalebox{1.5}{63.2\reproduce} & \scalebox{1.5}{61.7\reproduce} & \scalebox{1.5}{54.0\reproduce} & \scalebox{1.5}{68.0\reproduce}   \\
& \scalebox{1.5}{HallusionBench} & \scalebox{1.5}{53.8\original} & \scalebox{1.5}{57.3\original} & \scalebox{1.5}{63.2\original} & \scalebox{1.5}{59.8\fromglm} & \scalebox{1.5}{55.2\original} & \scalebox{1.5}{65.4\original} & \scalebox{1.5}{64.2\original} & \scalebox{1.5}{65.2\reproduce}  \\
& \scalebox{1.5}{MVBench} & \scalebox{1.5}{72.1\original} & \scalebox{1.5}{76.5\original} & \scalebox{1.5}{68.4\original} & \scalebox{1.5}{59.7\fromglm} & \scalebox{1.5}{70.4\original} & \scalebox{1.5}{73.0\original} & \scalebox{1.5}{64.2\reproduce} & \scalebox{1.5}{74.0\reproduce}   \\
& \scalebox{1.5}{VideoMME (w/o sub)} & \scalebox{1.5}{68.7\original} & \scalebox{1.5}{72.9\original} & \scalebox{1.5}{68.2\original} & \scalebox{1.5}{67.8\original} & \scalebox{1.5}{73.3\original} & \scalebox{1.5}{74.6\original} & \scalebox{1.5}{63.6\reproduce} & \scalebox{1.5}{81.8\reproduce}   \\
& \scalebox{1.5}{MLVU} & \scalebox{1.5}{73.0\original} & \scalebox{1.5}{78.2\original} & \scalebox{1.5}{71.5\reproduce} & \scalebox{1.5}{74.2\original} & \scalebox{1.5}{74.6\original} & \scalebox{1.5}{75.3\reproduce} & \scalebox{1.5}{62.2\reproduce} & \scalebox{1.5}{77.3\reproduce}   \\
& \scalebox{1.5}{LongVideoBench} & \scalebox{1.5}{63.8\original} & \scalebox{1.5}{67.1\original} & \scalebox{1.5}{65.7\reproduce} & \scalebox{1.5}{64.5\original} & \scalebox{1.5}{60.7\original} & \scalebox{1.5}{68.8\reproduce} & \scalebox{1.5}{57.7\reproduce} & \scalebox{1.5}{72.6\reproduce}   \\

\mycmidrulecolumntwo
& \scalebox{1.5}{Overall} & \scalebox{1.5}{71.0\original} & \scalebox{1.5}{74.1\original} & \scalebox{1.5}{68.9\original} & \scalebox{1.5}{67.2\original} & \scalebox{1.5}{70.8\original} & \scalebox{1.5}{72.5\original} & \scalebox{1.5}{67.9\original} & \scalebox{1.5}{74.0\original} \\

\mymidrule
\multirow{9}{*}{\scalebox{1.5}{Reasoning}} & \scalebox{1.5}{MMMU (val)} & \scalebox{1.5}{75.6\original} & \scalebox{1.5}{77.7\original} & \scalebox{1.5}{68.0\original} & \scalebox{1.5}{64.0\original} & \scalebox{1.5}{68.2\fromoc} & \scalebox{1.5}{75.4\original} & \scalebox{1.5}{74.2\original} & \scalebox{1.5}{84.2\original} \\
& \scalebox{1.5}{MathVista} & \scalebox{1.5}{80.9\original} & \scalebox{1.5}{82.7\original} & \scalebox{1.5}{80.7\original} & \scalebox{1.5}{80.1\original} & \scalebox{1.5}{74.2\fromoc} & \scalebox{1.5}{84.6\original} & \scalebox{1.5}{79.2\reproduce} & \scalebox{1.5}{81.9\reproduce} \\
& \scalebox{1.5}{MathVision} & \scalebox{1.5}{55.7\original} & \scalebox{1.5}{63.9\original} & \scalebox{1.5}{54.4\original} & \scalebox{1.5}{54.4\fromglm} & \scalebox{1.5}{39.3\fromoc} & \scalebox{1.5}{65.6\original} & \scalebox{1.5}{64.8\original} & \scalebox{1.5}{72.0\reproduce} \\
& \scalebox{1.5}{MathVerse (vision-only)} & \scalebox{1.5}{60.4\original} & \scalebox{1.5}{68.5\original} & \scalebox{1.5}{68.4\original} & \scalebox{1.5}{54.6\fromglm} & \scalebox{1.5}{47.3\fromoc} & \scalebox{1.5}{72.1\original} & \scalebox{1.5}{62.7\reproduce} & \scalebox{1.5}{81.2\reproduce} \\
& \scalebox{1.5}{DynaMath} & \scalebox{1.5}{36.5\original} & \scalebox{1.5}{46.5\original} & \scalebox{1.5}{42.5\original} & \scalebox{1.5}{28.1\fromglm} & \scalebox{1.5}{35.9\fromoc} & \scalebox{1.5}{53.9\original} & \scalebox{1.5}{50.1\original} & \scalebox{1.5}{60.9\reproduce} \\
& \scalebox{1.5}{WeMath} & \scalebox{1.5}{48.4\original} & \scalebox{1.5}{62.3\original} & \scalebox{1.5}{63.8\original} & \scalebox{1.5}{42.0\fromglm} & \scalebox{1.5}{49.1\fromoc} & \scalebox{1.5}{68.8\original} & \scalebox{1.5}{59.8\reproduce} & \scalebox{1.5}{71.1\reproduce} \\
& \scalebox{1.5}{OlympiadBench} & \scalebox{1.5}{62.9\original} & \scalebox{1.5}{68.7\original} & \scalebox{1.5}{56.3\reproduce} & \scalebox{1.5}{47.4\fromglm} & \scalebox{1.5}{37.8\reproduce} & \scalebox{1.5}{64.0\reproduce} & \scalebox{1.5}{66.8\reproduce} & \scalebox{1.5}{73.2\reproduce} \\
& \scalebox{1.5}{LogicVista} & \scalebox{1.5}{55.7\original} & \scalebox{1.5}{66.7\original} & \scalebox{1.5}{60.4\original} & \scalebox{1.5}{51.4\fromglm} & \scalebox{1.5}{55.7\fromoc} & \scalebox{1.5}{62.4\original} & \scalebox{1.5}{60.2\reproduce} & \scalebox{1.5}{70.0\reproduce} \\
\mycmidrulecolumntwo
& \scalebox{1.5}{Overall} & \scalebox{1.5}{59.5\original} & \scalebox{1.5}{67.1\original} & \scalebox{1.5}{61.8\original} & \scalebox{1.5}{52.8\original} & \scalebox{1.5}{50.9\original} & \scalebox{1.5}{68.4\original} & \scalebox{1.5}{64.7\original} & \scalebox{1.5}{74.3\original} \\

\mymidrule
\multirow{8}{*}{\scalebox{1.5}{Text}} & \scalebox{1.5}{MATH500} & \scalebox{1.5}{96.6\original} & \scalebox{1.5}{98.0\original} & \scalebox{1.5}{81.8\reproduce} & \scalebox{1.5}{91.8\reproduce} & \scalebox{1.5}{82.8\reproduce} & \scalebox{1.5}{94.2\reproduce} & \scalebox{1.5}{85.6\reproduce} & \scalebox{1.5}{97.8\reproduce} \\
& \scalebox{1.5}{AIME24} & \scalebox{1.5}{79.4\original} & \scalebox{1.5}{84.7\original} & \scalebox{1.5}{36.2\reproduce} & \scalebox{1.5}{54.0\reproduce} & \scalebox{1.5}{15.0\reproduce} & \scalebox{1.5}{80.1\reproduce} & \scalebox{1.5}{86.6\reproduce} & \scalebox{1.5}{90.0\reproduce} \\
& \scalebox{1.5}{AIME25} & \scalebox{1.5}{62.7\original} & \scalebox{1.5}{75.6\original} & \scalebox{1.5}{32.0\reproduce} & \scalebox{1.5}{39.1\reproduce} & \scalebox{1.5}{13.3\reproduce} & \scalebox{1.5}{72.8\reproduce} & \scalebox{1.5}{82.9\original} & \scalebox{1.5}{94.6\original} \\
& \scalebox{1.5}{GPQA} & \scalebox{1.5}{68.2\original} & \scalebox{1.5}{73.2\original} & \scalebox{1.5}{50.3\reproduce} & \scalebox{1.5}{42.3\reproduce} & \scalebox{1.5}{52.0\reproduce} & \scalebox{1.5}{56.6\reproduce} & \scalebox{1.5}{73.0\original} & \scalebox{1.5}{85.7\original} \\
& \scalebox{1.5}{MMLU-Pro} & \scalebox{1.5}{75.3\original} & \scalebox{1.5}{81.3\original} & \scalebox{1.5}{57.1\reproduce} & \scalebox{1.5}{68.5\reproduce} & \scalebox{1.5}{51.1\reproduce} & \scalebox{1.5}{69.7\reproduce} & \scalebox{1.5}{58.6\reproduce} & \scalebox{1.5}{85.6\reproduce} \\
& \scalebox{1.5}{C-Eval} & \scalebox{1.5}{83.2\original} & \scalebox{1.5}{90.9\original} & \scalebox{1.5}{72.3\reproduce} & \scalebox{1.5}{64.4\reproduce} & \scalebox{1.5}{88.2\reproduce} & \scalebox{1.5}{89.1\reproduce} & \scalebox{1.5}{84.7\reproduce} & \scalebox{1.5}{88.2\reproduce} \\
& \scalebox{1.5}{GAOKAO} & \scalebox{1.5}{91.9\original} & \scalebox{1.5}{94.5\original} & \scalebox{1.5}{78.4\reproduce} & \scalebox{1.5}{72.6\reproduce} & \scalebox{1.5}{92.9\reproduce} & \scalebox{1.5}{93.1\reproduce} & \scalebox{1.5}{70.2\reproduce} & \scalebox{1.5}{94.1\reproduce} \\
& \scalebox{1.5}{IFEval} & \scalebox{1.5}{74.3\original} & \scalebox{1.5}{83.7\original} & \scalebox{1.5}{71.5\reproduce} & \scalebox{1.5}{65.8\reproduce} & \scalebox{1.5}{83.9\reproduce} & \scalebox{1.5}{82.4\reproduce} & \scalebox{1.5}{73.4\reproduce} & \scalebox{1.5}{94.6\reproduce} \\
\mycmidrulecolumntwo
& \scalebox{1.5}{Overall} & \scalebox{1.5}{78.9\original} & \scalebox{1.5}{85.3\original} & \scalebox{1.5}{60.0\original} & \scalebox{1.5}{62.3\original} & \scalebox{1.5}{59.9\original} & \scalebox{1.5}{79.8\original} & \scalebox{1.5}{76.9\original} & \scalebox{1.5}{91.3\original} \\

\mymidrule
\multirow{9}{*}{\scalebox{1.5}{Agentic}} & \scalebox{1.5}{SGP-Bench} & \scalebox{1.5}{69.4\original} & \scalebox{1.5}{70.7\original} & \scalebox{1.5}{57.1\reproduce} & \scalebox{1.5}{44.9\reproduce} & \scalebox{1.5}{57.1\reproduce} & \scalebox{1.5}{66.1\reproduce} & \scalebox{1.5}{56.5\reproduce} & \scalebox{1.5}{77.5\reproduce} \\
& \scalebox{1.5}{ScreenSpot} & \scalebox{1.5}{86.6\original} & \scalebox{1.5}{89.8\original} & \scalebox{1.5}{--} & \scalebox{1.5}{--} & \scalebox{1.5}{87.1\original} & \scalebox{1.5}{--} & \scalebox{1.5}{--} & \scalebox{1.5}{--} \\
& \scalebox{1.5}{ScreenSpot-v2} & \scalebox{1.5}{87.3\original} & \scalebox{1.5}{92.9\original} & \scalebox{1.5}{--} & \scalebox{1.5}{91.4\original} & \scalebox{1.5}{--} & \scalebox{1.5}{--} & \scalebox{1.5}{--} & \scalebox{1.5}{--} \\
& \scalebox{1.5}{OSWorld-G} & \scalebox{1.5}{42.4\original} & \scalebox{1.5}{53.2\original} & \scalebox{1.5}{--} & \scalebox{1.5}{52.5\original} & \scalebox{1.5}{--} & \scalebox{1.5}{--} & \scalebox{1.5}{--} & \scalebox{1.5}{--} \\
& \scalebox{1.5}{VSI-Bench} & \scalebox{1.5}{63.7\original} & \scalebox{1.5}{69.5\original} & \scalebox{1.5}{39.2\reproduce} & \scalebox{1.5}{37.4\reproduce} & \scalebox{1.5}{36.1\reproduce} & \scalebox{1.5}{41.4\reproduce} & \scalebox{1.5}{34.2\reproduce} & \scalebox{1.5}{37.5\reproduce} \\
& \scalebox{1.5}{ERQA} & \scalebox{1.5}{41.5\original} & \scalebox{1.5}{46.8\original} & \scalebox{1.5}{45.8\fromglm} & \scalebox{1.5}{36.0\fromglm} & \scalebox{1.5}{44.8\fromglm} & \scalebox{1.5}{46.5\fromglm} & \scalebox{1.5}{44.5\fromglm} & \scalebox{1.5}{65.7\reproduce} \\
& \scalebox{1.5}{SpaCE-10} & \scalebox{1.5}{45.5\original} & \scalebox{1.5}{55.0\original} & \scalebox{1.5}{43.4\reproduce} & \scalebox{1.5}{39.2\reproduce} & \scalebox{1.5}{37.9\reproduce} & \scalebox{1.5}{51.6\reproduce} & \scalebox{1.5}{42.6\reproduce} & \scalebox{1.5}{43.8\reproduce} \\
& \scalebox{1.5}{OmniSpatial} & \scalebox{1.5}{48.1\original} & \scalebox{1.5}{51.9\original} & \scalebox{1.5}{47.7\fromglm} & \scalebox{1.5}{37.3\fromglm} & \scalebox{1.5}{47.9\fromglm} & \scalebox{1.5}{51.0\fromglm} & \scalebox{1.5}{47.0\fromglm} & \scalebox{1.5}{59.6\reproduce} \\
\mycmidrulecolumntwo
& \scalebox{1.5}{Overall} &  \scalebox{1.5}{60.6\original} & \scalebox{1.5}{66.2\original} & \scalebox{1.5}{--} & \scalebox{1.5}{--} & \scalebox{1.5}{--} & \scalebox{1.5}{--} & \scalebox{1.5}{--} & \scalebox{1.5}{--} \\

\mybottomrule
\end{tabular}
}
\vspace{2mm}
\caption{
    \textbf{The overall comparison of InternVL3.5 series and existing open-source and closed-source MLLMs.}
    \reproduce: reproduced through VLMEvalkit~\cite{duan2024vlmevalkit}. 
    \fromglm: reported by GLM-4.5V~\cite{hong2025glm_v_thinking}.
    \fromoc: reported by OpenCompass~\cite{opencompass2023}.
}
\label{tab:exp-overall} %
\end{threeparttable}
\end{table}

\subsection{Overall Comparison with Other Advanced MLLMs}
\label{sec:exp-overall}

Table~\ref{tab:exp-overall} presents a comprehensive evaluation of InternVL3.5’s performance across 35 benchmarks categorized into four key multimodal task types:
(1)
\textbf{General Tasks}: MMStar~\cite{chen2024mmstar}, MMVet~\cite{yu2023mmvet}, MMBench V1.1 (en)~\cite{liu2023mmbench}, MTVQA~\cite{tang2024mtvqa}, AI2D~\cite{kembhavi2016ai2d}, OCRBench~\cite{liu2023ocrbench},
WildVision~\cite{lu2024wildvision},
MME-RealWorld (en)~\cite{zhang2024mme}, HallusionBench~\cite{guan2023hallusionbench},
MVBench~\cite{li2024mvbench}, VideoMME~\cite{fu2024video}, MLVU~\cite{MLVU}, 
(2)
\textbf{Reasoning Tasks}: MMMU~\cite{yue2023mmmu}, MathVista~\cite{lu2023mathvista}, MathVision~\cite{lu2023mathvista}, MathVerse~\cite{zhang2024mathverse}, DynaMath~\cite{zou2024dynamath}, WeMath~\cite{qiao2024wemath},
OlympiadBench~\cite{he2024olympiadbench},
LogicVista~\cite{xiao2024logicvista};
(3)
\textbf{Text-Centric Tasks}: MATH500~\cite{lightman2023prm800k}, AIME24~\cite{aime2024}, AIME25~\cite{aime2025}, GPQA~\cite{rein2024gpqa}, MMLU-Pro~\cite{li2023cmmlu}, GAOKAO~\cite{Zhang2023gaokao}, IFEval~\cite{zhou2023instruction};
(4)
\textbf{Agentic Tasks}: SGP-Bench~\cite{qiu2024can}, ScreenSpot~\cite{cheng2024seeclick}, ScreenSpot-v2~\cite{wu2024atlas}, OSWorld-G~\cite{xie2024osworld}, VSI-Bench~\cite{yang2024think}, ERQA~\cite{erqa}, SpaCE-10~\cite{gong2025space10}.

We report results of our flagship models (InternVL3.5-30B-A3B and InternVL3.5-241B-A28B) and frontier open-source MLLMs (GLM-4.1V~\cite{hong2025glm_v_thinking}, Kimi-VL-A3B-2506~\cite{team2025kimi}, GLM-4.5V~\cite{hong2025glm_v_thinking}, Qwen2.5-VL-72B~\cite{bai2025qwen2_5} and Step-3~\cite{wang2025step}). We also include a state-of-the-art closed-source MLLM (GPT-5~\cite{gpt5}) for comparison.

These results highlight InternVL3.5’s strong capabilities across diverse tasks. For general multimodal tasks, InternVL3.5 demonstrates leading performance among open-source models on general multimodal understanding (\textit{e.g.,} MMVet), multilingual (\textit{e.g.,} MTVQA), OCR (\textit{e.g.,} OCRBench), real-world (\textit{e.g.,} MME-RealWorld), and video (\textit{e.g.,} LongVideoBench) benchmarks. It even achieves a similar overall score as GPT-5~\cite{gpt5} (74.1 \textit{vs.} 74.0), the state-of-the-art closed-source MLLM. Nevertheless, some tasks like HallusionBench still pose challenges, indicating the need for further refinements.

We also observe particularly significant gains in complex multimodal reasoning, as evidenced by the scores of 77.7 on MMMU and 82.7 on MathVista, which surpass most open-source models and approaching top-tier commercial systems.
These improvements are largely driven by enhanced training strategies, especially Cascade RL, and refined test-time scaling methodologies, enabling robust generalization in challenging domains such as mathematical reasoning (e.g., MathVerse) and multidisciplinary understanding (e.g., MTVQA).

For text-related tasks, InternVL3.5 outperforms most open-source models with significant margins. This is primarily due to our \textit{native pre-training strategy} introduced in InternVL3~\cite{zhu2025internvl3}, where we include a large amount of text-only data in the training process. This strategy allows the model to simultaneously acquire linguistic and multimodal abilities in a more efficient and integrated manner, and preserve the language capabilities of the pre-trained LLM to avoid the catastrophic forgetting issue~\cite{luo2024mono_internvl}. As a result, InternVL3.5 achieves high performance on both general (\textit{e.g.,} IFEval) and reasoning (\textit{e.g.,} AIME24 and MMLU-Pro) benchmarks and narrows the gap with GPT-5.

We also demonstrate the versatility of InternVL3.5 through agentic benchmarks. On SGP-Bench, an SVG understanding benchmark, InternVL3.5 achieves leading performance (69.4 and 70.7) that surpasses all open-source models. For GUI tasks, our models show strong abilities on GUI grounding (ScreenSpot) and online agentic (OSWorld-G) benchmarks. Evaluation on embodied tasks (VSI-Bench, ERQA, SpaCE-10, OmniSpatial) validates InternVL3.5's spatial reasoning skills and the competence to understand complex and dynamic environments, highlighting its potential in embodied agents, robotic navigation, and interactive scene perception.

\subsection{Multimodal Reasoning and Mathematics}
\label{sec:exp-reasoning}

To comprehensively evaluate the multimodal reasoning and mathematical capabilities of InternVL3.5, we conduct extensive experiments across a series of benchmarks, including MMMU~\cite{yue2023mmmu} for multidisciplinary reasoning, MathVista~\cite{lu2023mathvista}, MathVision~\cite{wang2024mathvision}, and MathVerse~\cite{zhang2024mathverse} for mathematical reasoning, as well as DynaMath~\cite{zou2024dynamath}, WeMath~\cite{qiao2024wemath}, and LogicVista~\cite{xiao2024logicvista} for complementary logical reasoning assessment.

As shown in Table~\ref{tab:exp-reasoning}, InternVL3.5 achieves state-of-the-art performance across all evaluated benchmarks among open-source models.
Compared with the previous generation InternVL3, InternVL3.5 improves reasoning performance by more than 10 points across all model sizes relative to their counterparts of the comparable scale.
Furthermore, InternVL3.5-241B-A28B consistently outperforms all open-source counterparts, obtaining an overall average score of 66.9. It is followed by InternVL3.5-38B, which secures the second highest score with an average of 66.0. At mid-scale, the performance of InternVL3.5 is particularly impressive, with InternVL3.5-30B-A3B and InternVL3.5-14B attaining scores of 75.6 and 73.3 on MMMU respectively, both outperforming the larger InternVL3-78B (72.2).
At lightweight scales, InternVL3.5 achieves substantial gains over open-source baselines. Compared to its predecessor InternVL3, it delivers marked improvements: the InternVL3.5-2B model has an average score of 50.7, significantly higher than that of InternVL3-2B with a score of 32.4; the InternVL3.5-4B model, with a score of 57.4, far exceeds the score of 33.5 of MiniCPM-V-4; and the InternVL3.5-8B model achieves a score of 60.3, substantially surpassing the score of 44.3 of InternVL3-8B.  These significant improvements  are mainly from our Cascade RL, showing its strong scalability for reasoning tasks.
The ablation study about how different training stages influence the reasoning abilities of our models is presented in Section~\ref{sec:exp-ablation}.

Furthermore, our experiments also demonstrate that Cascade RL can be seamlessly combined with parallel thinking and obtain further gains. For instance, with parallel thinking,  the overall reasoning  scores of InternVL3.5-4B, InternVL3.5-8B and InternVL3.5-241B-A28B are further improved by +2.6\%, +2.1\% and +1.8\%, respectively, highlighting the effectiveness of test-time scaling for reasoning-related tasks.

\begin{table}[t]
\centering
\scriptsize
\setlength{\tabcolsep}{4pt}
    \begin{tabular}{l|ccccccc|c}
    \mytoprule
    \textbf{Model} & \makecell{\textbf{MMMU}\\\textbf{(val)}} & \makecell{\textbf{MathVista}\\\textbf{(mini)}} & \makecell{\textbf{MathVision}} & \makecell{\textbf{MathVerse}\\\textbf{(vision-only)}} & \makecell{\textbf{DynaMath}\\\textbf{(worst case)}} & \textbf{WeMath} & \textbf{LogicVista} & \textbf{Overall} \\
    
    \mymidrule
    InternVL3-1B~\cite{zhu2025internvl3} & 43.4 & 45.8 & 18.8 & 18.7 & 5.8 & 13.4 & 29.8 & 25.1 \\
    \rowcolor{oursgray} InternVL3.5-1B & 44.2 & 59.3 & 27.3 & 37.8 & 17.2 & 21.5 & 29.3 & 33.8 \\
    \rowcolor{oursgray} \textit{\textcolor{blue}{w/ Parallel Thinking}~\cite{wang2025visualprm}} & 51.0 & 69.5 & 33.4 & 45.8 & 25.0 & 36.4 & 41.8 & 43.3 \\
    
    \mymidrule
    Ovis-2B~\cite{lu2024ovis} & 45.6 & 64.1 & 17.7 & 29.4 & 10.0 & 9.9 & 34.7 & 30.2 \\
    Qwen2.5-VL-3B~\cite{bai2025qwen2_5} & 51.2 & 61.2 & 21.9 & 31.2 & 13.2 & 22.9 & 40.3 & 34.6 \\
    InternVL3-2B~\cite{zhu2025internvl3} & 48.6 & 57.0 & 21.7 & 25.3 & 14.6 & 22.4 & 36.9 & 32.4 \\
    \rowcolor{oursgray} InternVL3.5-2B & 59.0 & 71.8 & 42.8 & 53.4 & 31.5 & 48.5 & 47.7 & 50.7  \\
    \rowcolor{oursgray} \textit{\textcolor{blue}{w/ Parallel Thinking}~\cite{wang2025visualprm}} & 64.8 & 74.3 & 49.0 & 54.2 & 33.3 & 52.6 & 49.4 & 53.9 \\

    \mymidrule
    Ovis-4B~\cite{lu2024ovis} & 49.0 & 69.6 & 21.5 & 38.5 & 18.0 & 16.9 & 35.3 & 35.5 \\
    MiniCPM-V-4-4B~\cite{yao2024minicpm} & 51.2  & 66.9  & 20.7  & 18.3  & 14.2  & 32.7  & 30.6  & 33.5 \\
    InternVL2.5-4B~\cite{chen2024internvl_2_5} & 51.8 & 64.1 & 18.4 & 27.7 & 15.2 & 21.2 & 34.2 & 33.2 \\
    \rowcolor{oursgray} InternVL3.5-4B & 66.6 & 77.1 & 54.4 & 61.7 & 35.7 & 50.1 & 56.4 & 57.4  \\
    \rowcolor{oursgray} \textit{\textcolor{blue}{w/ Parallel Thinking}~\cite{wang2025visualprm}} & 71.4 & 79.2 & 57.5 & 60.0 & 38.3 & 53.0 & 60.4 & 60.0 \\
    
    \mymidrule
    MiniCPM-o2.6~\cite{yao2024minicpm} & 50.9 & 73.3 & 21.7 & 35.0 & 10.4 & 25.2 & 36.0 & 36.1 \\
    Ovis-8B~\cite{lu2024ovis} & 57.4 & 71.8 & 25.9 & 42.3 & 20.4 & 27.2 & 39.4 & 40.6 \\
    Qwen2.5-VL-8B~\cite{bai2025qwen2_5} & 55.0 & 67.8 & 25.4 & 41.1 & 21.0 & 35.2 & 44.1 & 41.4 \\
    MiMo-VL-RL-8B~\cite{coreteam2025mimovltechnicalreport} & 66.7  & 81.5  & 60.4  & 71.5  & 45.9  & 66.3  & 61.4  & 64.8 \\
    Keye-VL-8B~\cite{team2025keye_vl} & 71.4  & 80.7  & 50.8  & 54.8  & 37.3  & 60.7  & 54.8  & 58.6 \\
    GLM-4.1V-9B~\cite{hong2025glm_v_thinking} & 68.0  & 80.7  & 54.4  & 68.4  & 42.5  & 63.8  & 60.4  & 62.6 \\
    InternVL3-8B~\cite{zhu2025internvl3} & 62.7 & 71.6 & 29.3 & 39.8 & 25.5 & 37.1 & 44.1 & 44.3 \\
    \rowcolor{oursgray} InternVL3.5-8B & 73.4 & 78.4 & 56.8 & 61.5 & 37.7 & 57.0 & 57.3 & 60.3  \\
    \rowcolor{oursgray} \textit{\textcolor{blue}{w/ Parallel Thinking}~\cite{wang2025visualprm}} & 73.4 & 80.8 & 59.9 & 62.6 & 39.9 & 57.6 & 62.6 & 62.4 \\

    \mymidrule
    Gemma-3-12B~\cite{team2025gemma} & 55.2  & 56.1  & 30.3  & 21.1  & 20.8  & 33.6  & 41.2  & 36.9 \\
    Ovis2-16B~\cite{lu2024ovis} & 60.7 & 73.7 & 30.1 & 45.8 & 26.3 & 45.0 & 47.4 & 47.0 \\
    InternVL3-14B~\cite{zhu2025internvl3} & 67.1 & 75.1 & 37.2 & 44.4 & 31.3 & 43.0 & 51.2 & 49.9 \\
    \rowcolor{oursgray} InternVL3.5-14B & 73.3 & 80.5 & 59.9 & 62.8 & 38.7 & 58.7 & 60.2 & 62.0  \\
    \rowcolor{oursgray} \textit{\textcolor{blue}{w/ Parallel Thinking}~\cite{wang2025visualprm}} & 74.3 & 81.5 &  61.4 &  64.1 &  41.3 &  60.8 &  61.3 &  63.5 \\

    \midrule
    Kimi-VL-A3B-2506~\cite{team2025kimi} & 64.0  & 80.1  & 54.4  & 54.6  & 28.1  & 42.0  & 51.4  & 53.5 \\
    \rowcolor{oursgray} InternVL3.5-20B-A4B & 72.6 & 78.0 & 53.0 & 57.0 & 33.1 & 41.4 & 56.8 & 56.0  \\
    \rowcolor{oursgray} \textit{\textcolor{blue}{w/ Parallel Thinking}~\cite{wang2025visualprm}} & 74.0 & 79.2 & 55.7 & 58.9 & 35.9 & 45.5 & 58.2 & 58.2 \\
    \rowcolor{oursgray} InternVL3.5-30B-A3B & 75.6 & 80.9 & 55.7 & 60.4 & 36.5 & 48.4 & 55.7 & 59.0  \\
    \rowcolor{oursgray} \textit{\textcolor{blue}{w/ Parallel Thinking}~\cite{wang2025visualprm}} & 75.0 & 82.1 & 58.5 & 61.4 & 38.3 & 57.5 & 59.7 & 61.8 \\

    \mymidrule
    Gemma-3-27B~\cite{team2025gemma} & 64.9  & 59.8  & 39.8  & 34.0  & 28.5  & 37.9  & 47.3  & 44.6  \\
    Ovis2-34B~\cite{lu2024ovis} & 66.7 & 76.1 & 31.9 & 50.1 & 27.5 & 51.9 & 49.9 & 50.6 \\
    Qwen2.5-VL-32B~\cite{bai2025qwen2_5} & 70.2  & 74.8  & 38.1  & 57.6  & 35.1  & 46.5  & 52.6  & 53.6 \\
    Skywork-R1V3-38B~\cite{shen2025skywork_r1v3} & 76.0  & 77.1  & 52.6  & 59.6  & 35.1  & 56.5  & 59.7  & 59.5 \\
    InternVL3-38B~\cite{zhu2025internvl3} & 70.1 & 75.1 & 34.2 & 48.2 & 35.3 & 48.6 & 58.4 & 52.8 \\
    \rowcolor{oursgray} InternVL3.5-38B & 76.9 & 81.9 & 63.7 & 67.6 & 41.7 & 64.8 & 65.3 & 66.0  \\
    \rowcolor{oursgray} \textit{\textcolor{blue}{w/ Parallel Thinking}~\cite{wang2025visualprm}} & 76.7 & 83.8 & 65.6 & 66.5 & 44.9 & 67.8 & 64.7 & 67.1 \\
    
    \mymidrule
    GPT-5-nano-20250807~\cite{gpt5} & 72.6  & 73.1  & 59.7  & 66.6  & 47.9  & 59.4  & 57.5  & 62.4 \\
    GPT-5-20250807~\cite{gpt5} & 81.8  & 81.9  & 72.0  & 81.2  & 60.9  & 71.1  & 70.0  & 74.1 \\
    Claude-3.7-Sonnet~\cite{claude3series2024} & 75.0 & 66.8 & 41.9 & 46.7 & 39.7 & 49.3 & 58.2 & 53.9 \\
    Gemini-2.0-Pro~\cite{gemini2_0} & 69.9 & 71.3 & 48.1 & 67.3 & 43.3 & 56.5 & 53.2 & 58.5 \\
    Gemini-2.5-Pro~\cite{gemini2_0} & 74.7  & 80.9  & 69.1  & 76.9  & 56.3  & 78.0  & 73.8  & 72.8  \\
    Doubao-1.5-Pro~\cite{guo2025seedvl1_5} & 73.8  & 78.6  & 51.5  & 64.7  & 44.9  & 65.7  & 64.2  & 63.3 \\
    GLM-4.5V~\cite{hong2025glm_v_thinking} & 75.4  & 84.6  & 65.6  & 72.1  & 53.9  & 68.8  & 62.4  & 69.0 \\
    QvQ-72B-Preview~\cite{qvq-72b-preview} & 70.3 & 70.3 & 34.9 & 48.2 & 30.7 & 39.0 & 58.2 & 50.2 \\
    Qwen2.5-VL-72B~\cite{bai2025qwen2_5} & 68.2 & 74.2 & 39.3 & 47.3 & 35.9 & 49.1 & 55.7 & 52.8 \\
    Step3-321B-A38B~\cite{wang2025step} & 74.2  & 79.2  & 64.8  & 62.7  & 50.1  & 59.8  & 60.2  & 64.4  \\
    InternVL3-78B~\cite{zhu2025internvl3} & 72.2 & 79.0 & 43.1 & 51.0 & 35.1 & 46.1 & 55.9 & 54.6 \\
    \rowcolor{oursgray} InternVL3.5-241B-A28B & 77.7 & 82.7 & 63.9 & 68.5 & 46.5 & 62.3 & 66.7 & 66.9  \\
    \rowcolor{oursgray} \textit{\textcolor{blue}{w/ Parallel Thinking}~\cite{wang2025visualprm}} & 78.7 & 84.8  &  65.9 &  71.6 &  47.8 &  64.4 &  67.6 &  68.7 \\
    \mybottomrule
    \end{tabular}
\vspace{2mm}
\caption{\textbf{Comparison of multimodal reasoning and mathematical performance.} MMMU~\cite{yue2023mmmu} is a multidisciplinary reasoning benchmark.
MathVista~\cite{lu2023mathvista}, MathVision~\cite{wang2024mathvision}, MathVerse~\cite{zhang2024mathverse}, DynaMath~\cite{zou2024dynamath}, and WeMath~\cite{qiao2024wemath} are mathematics benchmarks. 
LogicVista~\cite{xiao2024logicvista} is a logical reasoning benchmark.
Part of the results are collected from other papers~\cite{team2025keye_vl,hong2025glm_v_thinking,coreteam2025mimovltechnicalreport,wang2024qwen2vl,zhu2025internvl3} and the OpenCompass leaderboard~\cite{opencompass2023}.
The overall score is the average score of all benchmarks.
}
\label{tab:exp-reasoning}
\vspace{-3mm}
\end{table}

\subsection{OCR, Chart, and Document Understanding}
\label{sec:exp-ocr}

To evaluate the comprehensive capabilities of the model across tasks related to text, document, and chart comprehension, we conduct an extensive assessment on nine benchmarks: AI2D~\cite{kembhavi2016ai2d}, ChartQA~\cite{masry2022chartqa}, TextVQA~\cite{singh2019textvqa}, DocVQA~\cite{mathew2021docvqa}, InfoVQA~\cite{mathew2022infographicvqa}, OCRBench~\cite{liu2023ocrbench}, SEED-2-Plus~\cite{li2024seedbench2plus}, CharXiv~\cite{wang2024charxiv}, and VCR~\cite{zhang2024vcr}.

As shown in Table~\ref{tab:exp-ocr}, InternVL3.5 achieves competitive results on these benchmarks, outperforming other open-source and closed-source models.
At the lightweight scale, InternVL3.5 demonstrates significant potential.
For instance, InternVL3.5-2B attains an overall average score of 76.7 across nine benchmarks, surpassing InternVL3-2B of similar size, which scores 74.7.
Specifically, on DocVQA, InfoVQA, and SEED-2-Plus, InternVL3.5-2B achieves scores of 89.4, 70.8, and 68.0, respectively, compared to 88.3, 66.1, and 64.6 for InternVL3-2B, highlighting InternVL3.5’s strong vision-language understanding capabilities.

Moreover, as the model scale increases, InternVL3.5 continues to deliver enhanced performance in vision-language understanding.
Across the same nine benchmarks, InternVL3.5-4B, InternVL3.5-20B-A4B,  InternVL3.5-14B, InternVL3.5-30B-A3B, and InternVL3.5-38B achieve overall average scores of 80.0, 81.7, 82.0, 83.9, and 84.6, respectively, demonstrating consistent improvements with larger model sizes.
On AI2D, InternVL3.5-2B, InternVL3.5-4B, InternVL3.5-14B and InternVL3.5-38B obtain scores of 78.8/89.1, 82.6/92.3, 85.1/93.3, and 87.8/95.1, respectively, indicating a consistent upward trend across larger models.
On ChartQA, InternVL3.5-4B shows a notable improvement in chart understanding, increasing from 80.7 (InternVL3.5-2B) to 86.0, reflecting a significant gain in visual reasoning capability.
Similarly, on CharXiv and VCR-EN-Easy, larger model variants also achieve measurable improvements in text and document understanding.

\begin{table}[t]
\centering
\scriptsize
\setlength{\tabcolsep}{3pt}
\resizebox{\linewidth}{!}{
\begin{tabular}{l|ccccccccc|c}
\mytoprule
\textbf{Model} & \makecell{\textbf{AI2D}\\\textbf{(w / wo M)}} & \makecell{\textbf{ChartQA}\\\textbf{(test avg)}} & \makecell{\textbf{TextVQA}\\\textbf{(val)}} & \makecell{\textbf{DocVQA}\\\textbf{(test)}} & \makecell{\textbf{InfoVQA}\\\textbf{(test)}} & \makecell{\textbf{OCR}\\\textbf{Bench}} & \makecell{\textbf{SEED-2}\\\textbf{Plus}} & \makecell{\textbf{CharXiv}\\\textbf{(RQ / DQ)}} & \makecell{\textbf{VCR-EN-Easy}\\\textbf{(EM / Jaccard)}} & \textbf{Overall} \\
\mymidrule
InternVL3-1B~\cite{zhu2025internvl3} & 69.4 / 78.3 & 75.3 & 74.1 & 81.9 & 53.7 & 790 & 58.2 & 21.0 / 47.1 & 89.3 / 96.2 & 68.6 \\
\rowcolor{oursgray} InternVL3.5-1B & 71.1 / 81.8  & 77.7  & 71.5  & 85.6  & 60.5  & 795  & 62.3  & 26.9 / 60.6 & 83.5 / 94.0 & 71.3  \\
\mymidrule
Qwen2-VL-2B~\cite{wang2024qwen2vl} & 74.7 / 84.6 & 73.5 & 79.7 & 90.1 & 65.5 & 809 & 62.4 & -- & 81.5 / \rsp & -- \\
Aquila-VL-2B~\cite{gu2024aquilavl} & 75.0 / \rsp & 76.5 & 76.4 & 85.0 & 58.3 & 772 & 63.0 & -- & 70.0 / \rsp & -- \\
Qwen2.5-VL-3B~\cite{bai2025qwen2_5} & 81.6 / \rsp & 84.0 & 79.3 & 93.9 & 77.1 & 797 & 67.6 & 31.3 / 58.6 & -- & -- \\
InternVL3-2B~\cite{zhu2025internvl3} & 78.7 / 87.4 & 80.2 & 77.0 & 88.3 & 66.1 & 835 & 64.6 & 28.3 / 54.7 & 91.2 / 96.9 & 74.7 \\
\rowcolor{oursgray} InternVL3.5-2B & 78.8 / 89.1 & 80.7 & 76.5 & 89.4 & 70.8 & 836 & 68.0 & 31.6 / 65.0 & 90.1 / 96.4 & 76.7 \\

\mymidrule
MiniCPM-V-4-4B~\cite{yao2024minicpm} & 80.9 / 91.4 & 73.0 & 81.4 & 94.0 & 67.0 & 862 & 67.0 & 31.9 / 56.4 & 80.9 / 90.1 & 75.0 \\
\rowcolor{oursgray} InternVL3.5-4B & 82.6 / 92.3 & 86.0 & 77.9 & 92.4 & 78.0 & 822 & 69.4 & 39.6 / 71.1 & 91.6 / 97.0 & 80.0 \\

\mymidrule
Ovis1.6-Gemma2-9B~\cite{lu2024ovis} & 84.4 / \rsp & -- & -- & -- & -- & 830 & -- & -- & -- & -- \\
MiniCPM-V2.6-8B~\cite{yao2024minicpm}  & 82.1 / \rsp & 82.4 & 80.1 & 90.8 & -- & 852 & 65.7 & 31.0 / 57.1 & 73.9 / 85.7 & -- \\
Molmo-7B-D~\cite{deitke2024molmo}  & \rsp / 93.2 & 84.1 & 81.7 & 92.2 & 72.6 & 694 & -- & -- & -- & -- \\
Qwen2-VL-7B~\cite{wang2024qwen2vl} & 83.0 / 92.1 & 83.0 & 84.3 & 94.5 & 76.5 & 866 & 69.0 & -- & 89.7 / 93.8 & -- \\
Qwen2.5-VL-7B~\cite{bai2025qwen2_5} & 83.9 / \rsp & 87.3 & 84.9 & 95.7 & 82.6 & 864 & 70.4 & 42.5 / 73.9 & -- & -- \\
Keye-VL-8B~\cite{team2025keye_vl} & 85.8 / 88.5  & 72.5  & 75.7  & 87.0  & 63.0  & 853  & 67.8  & 36.8 / 75.2  & -- & --  \\
GLM-4.1V-9B~\cite{hong2025glm_v_thinking} & 82.2 / 87.0 & 70.0 & 79.6 & 93.3 & 80.3 & 823 & 71.8 & 53.4 / 82.4 & 32.7 / 55.2 & 72.5 \\
InternVL3-8B~\cite{zhu2025internvl3} & 85.2 / 92.6 & 86.6 & 80.2 & 92.7 & 76.8 & 880 & 69.7 & 37.6 / 73.6 & 94.5 / 98.1 & 81.3 \\
InternVL3-9B~\cite{zhu2025internvl3} & 84.6 / 92.9 & 86.2 & 79.4 & 93.6 & 79.6 & 877 & 68.8 & 38.0 / 72.5 & 94.2 / 97.9 & 81.3 \\
\rowcolor{oursgray} InternVL3.5-8B & 84.0 / 92.8  & 86.7  & 78.2  & 92.3  & 79.1  & 840  & 70.8  & 44.4 / 72.2 & 92.6 / 97.3 & 81.2  \\

\mymidrule
Gemma3-12B~\cite{team2025gemma} & 84.2 / 87.2  & 75.7  & 67.7  & 87.1  & 64.9  & 702  & 65.5  & 24.9 / 65.4 & -- & -- \\
InternVL3-14B~\cite{zhu2025internvl3} & 86.0 / 93.7 & 87.3 & 80.5 & 94.1 & 83.6 & 875 & 70.3 & 43.1 / 82.2 & 94.8 / 98.2 & 83.4 \\
\rowcolor{oursgray} InternVL3.5-14B & 85.1 / 93.3 & 86.5 & 77.8 & 93.4 & 78.3 & 836 & 70.7 & 47.9 / 76.6 & 93.4 / 97.7 & 82.0 \\

\mymidrule
Kimi-VL-A3B-2506~\cite{team2025kimi} & 81.9 / 91.2  & 73.7 & 77.7 & 93.5 & 74.5 & 869 & 70.8 & 46.8 / 71.5 & 81.1 / 90.6 & 78.3\\
\rowcolor{oursgray} InternVL3.5-20B-A4B & 85.9 / 93.5 & 86.6 & 78.5 & 92.9 & 78.1 & 870 & 69.3 & 38.2 / 78.5 & 93.7 / 97.8 & 81.7 \\
\rowcolor{oursgray} InternVL3.5-30B-A3B & 86.8 / 94.5 & 87.4 & 80.5 & 94.2 & 81.4 & 880 & 70.6 & 48.0 / 81.8 & 94.9 / 98.2 & 83.9 \\

\mymidrule
Gemma3-27B~\cite{team2025gemma} & 84.5 / 88.2  & 78.0  & 65.1  & 86.6  & 65.1  & 717  & 68.1  & 25.7 / 63.8 & -- & -- \\
Qwen2.5-VL-32B~\cite{bai2025qwen2_5} & 84.0 / 92.9 & 64.9 & 78.9 & 94.8 & 83.4 & 850 & 72.1 & 46.1 / 84.9 & 88.9 / 93.0 & 80.7 \\
Cambrian-34B~\cite{tong2024cambrian} & 79.5 / \rsp & 75.6 & 76.7 & 75.5 & 46.0 & 600 & -- & 27.3 / 59.7 & 79.7 / 89.3 & -- \\
VILA-1.5-40B~\cite{lin2024vila} & 69.9 / \rsp & 67.2 & 73.6 & -- & -- & 460 & -- & 24.0 / 38.7 & -- & -- \\
InternVL3-38B~\cite{zhu2025internvl3} & 88.9 / 95.5 & 89.2 & 83.9 & 95.4 & 85.0 & 886 & 71.6 & 46.4 / 87.2 & 96.1 / 98.7 & 85.5 \\
\rowcolor{oursgray} InternVL3.5-38B & 87.8 / 95.1  & 88.8  & 82.7  & 94.0  & 83.8  & 870  & 71.0  & 48.1 / 83.1 & 95.3 / 98.4 & 84.6  \\

\mymidrule
GPT-4V~\cite{gpt4v} & 78.2 / 89.4 & 78.5 & 78.0 & 88.4 & 75.1 & 645 & 53.8 & 37.1 / 79.9 & 52.0 / 65.4 & 70.0 \\
GPT-4o-20240513~\cite{chatgpt4o} & 84.6 / 94.2 & 85.7 & 77.4 & 92.8 & 79.2 & 736 & 72.0 & 47.1 / 84.5 & 91.6 / 96.4 & 81.6 \\
Claude-3-Opus~\cite{claude3series2024} & 70.6 / 88.1 & 80.8 & 67.5 & 89.3 & 55.6 & 694 & 44.2 & 30.2 / 71.6 & 62.0 / 77.7 & 67.3 \\
Claude-3.5-Sonnet~\cite{claude3series2024} & 81.2 / 94.7 & 90.8 & 74.1 & 95.2 & 74.3 & 788 & 71.7 & 60.2 / 84.3 & 63.9 / 74.7 & 78.7 \\
Gemini-1.5-Pro~\cite{reid2024gemini1_5} & 79.1 / 94.4 & 87.2 & 78.8 & 93.1 & 81.0 & 754 & -- & 43.3 / 72.0 & 62.7 / 77.7 & -- \\
GLM-4.5V~\cite{hong2025glm_v_thinking} & 88.1 / 93.7 & 86.6 & 72.0 & 94.5 & 84.1 & 872 & 74.0 & 59.5 / 88.2 & 36.5 / 44.9 & 75.8 \\
NVLM-D-72B~\cite{dai2024nvlm} & 85.2 / 94.2 & 86.0 & 82.1 & 92.6 & -- & 853 & -- & -- & -- & -- \\
Molmo-72B~\cite{deitke2024molmo} & \rsp / 96.3 & 87.3 & 83.1 & 93.5 & 81.9 & -- & -- & -- & -- & -- \\
Qwen2-VL-72B~\cite{wang2024qwen2vl} & 88.1 / \rsp & 88.3 & 85.5 & 96.5 & 84.5 & 877 & -- & -- & 91.3 / 94.6 & -- \\
Qwen2.5-VL-72B~\cite{bai2025qwen2_5} & 88.7 / \rsp & 89.5 & 83.5 & 96.4 & 87.3 & 885 & 73.0 & 49.7 / 87.4 & -- & -- \\
InternVL3-78B~\cite{zhu2025internvl3} & 89.7 / 96.0 & 89.7 & 84.3 & 95.4 & 86.5 & 906 & 71.9 & 46.0 / 85.1 & 96.0 / 98.6 & 85.8 \\
\rowcolor{oursgray} InternVL3.5-241B-A28B & 87.3 / 95.0 & 88.0 & 84.5 & 94.9 & 82.0 & 907 & 71.9 & 48.3 / 83.9 & 97.0 / 99.0 & 85.2 \\
\mybottomrule
\end{tabular}
}
\vspace{2mm}
\caption{\textbf{Comparison of OCR, chart, and document understanding performance.} 
We compare OCR-related performance across 9 benchmarks: AI2D~\cite{kembhavi2016ai2d}, ChartQA~\cite{masry2022chartqa}, TextVQA~\cite{singh2019textvqa}, DocVQA~\cite{mathew2021docvqa}, InfoVQA~\cite{mathew2022infographicvqa}, OCRBench~\cite{liu2023ocrbench}, SEED-2-Plus~\cite{li2024seedbench2plus}, CharXiv~\cite{wang2024charxiv}, and VCR~\cite{zhang2024vcr}.
Part of results are collected from the OpenCompass leaderboard~\cite{opencompass2023} and other papers~\cite{dubey2024llama3, deitke2024molmo, claude3series2024, wang2024charxiv, zhang2024vcr}.
When calculating Overall, the score of OCR Bench is normalized from 0-1000 to 0-100.
}
\label{tab:exp-ocr}
\vspace{-3mm}
\end{table}

\begin{table*}[t!]
\scriptsize
\centering

\newcommand{\MMIU}{\makecell{\textbf{MMIU}}}
\newcommand{\Muir}{\makecell{\textbf{Muir}\\\textbf{Bench}}} %
\newcommand{\BLINK}{\makecell{\textbf{BLINK}\\\textbf{(val)}}}
\newcommand{\Mantis}{\makecell{\textbf{Mantis}\\\textbf{Eval}}} %
\newcommand{\RWQA}{\makecell{\textbf{RealWorld}\\\textbf{QA}}} %
\newcommand{\MMERW}{\makecell{\textbf{MME-RW}\\\textbf{(EN)}}} %
\newcommand{\RBench}{\makecell{\textbf{R-Bench}\\\textbf{(test)}}}
\newcommand{\TaskMe}{\makecell{\textbf{TaskMe}-\\\textbf{Anything}}}
\newcommand{\MMT}{\makecell{\textbf{MMT}\\\textbf{(val)}}}
\newcommand{\WILDV}{\makecell{\textbf{WildVision}\\\textbf{(win rate)}}}
\newcommand{\MIRB}{\makecell{\textbf{MIRB}\\\textbf{(avg)}}}
\newcommand{\RB}{\makecell{\textbf{R-Bench}\\\textbf{(dis)}}}

\setlength{\tabcolsep}{3pt}
\resizebox{\linewidth}{!}{
\begin{tabular}{l|cccccc|c|cccc|c}
\mytoprule
\textbf{Model Name}                                 & \BLINK & \Mantis & \MMIU & \Muir & \MMT & \MIRB & \textbf{Overall} &\RWQA  & \MMERW & \WILDV  & \RB  & \textbf{Overall} \\
\mymidrule
InternVL3-1B~\cite{zhu2025internvl3}                               & 42.9   & 50.2    & 39.3  & 31.2  & 52.9 & 36.1  & 42.1    & 58.2  & 46.0   & 43.8    & 60.4 & 52.1    \\
\rowcolor{oursgray} InternVL3.5-1B & 44.0 & 54.8 & 45.2 & 41.7 & 54.5 & 44.2 & 47.4 & 57.6 & 46.8 & 49.2 & 57.4 & 50.6  \\

\mymidrule
Qwen2-VL-2B~\cite{wang2024qwen2vl}         & 44.4   & --      & --    & --    & 55.1 & --    & --      & 62.6  & --     & --      & --   & --      \\
Qwen2.5-VL-3B~\cite{bai2025qwen2_5}          & 47.6   & --      & --    & 47.7    & -- & --    & --      & 65.4  & 53.1   & --      & --   & --      \\
InternVL3-2B~\cite{zhu2025internvl3}                               & 50.3   & 65.9    & 43.0  & 38.8  & 59.5 & 42.9  & 50.1    & 64.3  & 53.8   & 48.8    & 67.5 & 58.6    \\
\rowcolor{oursgray} InternVL3.5-2B & 51.3 & 58.5 & 44.9 & 44.0 & 58.5 & 45.9 & 50.5 & 62.0 & 49.7 & 66.6 & 62.4 & 58.5 \\

\mymidrule
MiniCPM-V-4~\cite{yao2024minicpm} & 53.0 & 71.4& --& 46.1 & 59.7 & --& --& 66.0 & 58.4 & 46.0 & 64.2 & 58.7 \\
\rowcolor{oursgray} InternVL3.5-4B & 58.1 & 62.7 & 49.2 & 53.1 & 64.3 & 55.9 & 57.2 & 66.3 & 59.8 & 69.8 & 68.7 & 65.1 \\

\mymidrule
Qwen2-VL-7B~\cite{wang2024qwen2vl}         & 53.2   & --      & --    & --    & 64.0 & --    & --      & 70.1  & 56.5   & --      & 64.0 & --      \\
Qwen2.5-VL-7B~\cite{bai2025qwen2_5}          & 56.4   & --      & --    & 59.6  & --   & --    & --      & 68.5  & 57.4   & --      & --   & --      \\
MiniCPM-V2.6~\cite{yao2024minicpm}         & 53.0   & 69.0    & --    & --    & 60.8 & --    & --      & 65.0  & --     & --      & --   & --      \\
Keye-VL-8B~\cite{team2025keye_vl} & 53.4 & --& --& 50.4 & 65.4 & --& --& 65.9 & 48.8 & 66.6 & 65.7 & 61.7 \\
GLM-4.1V-9B~\cite{hong2025glm_v_thinking} & 65.9 & -- & -- & 74.7 & 68.4 & -- & -- & 70.6 & 61.7 & 74.0 & 69.1 & 68.9 \\
InternVL3-8B~\cite{zhu2025internvl3}                               & 55.5   & 70.1    & 46.8  & 55.0  & 65.0 & 56.8  & 58.2    & 70.8  & 62.0   & 69.8    & 74.1 & 69.2    \\
InternVL3-9B~\cite{zhu2025internvl3}                               & 58.6   & 70.1    & 50.4  & 51.4  & 65.4 & 58.6  & 59.1    & 70.5  & 61.3   & 63.8    & 70.3 & 66.5    \\
\rowcolor{oursgray} InternVL3.5-8B & 59.5 & 70.5 & 49.4 & 55.8 & 66.7 & 57.3 & 59.9 & 67.5 & 62.8 & 76.6 & 69.7 & 67.4 \\

\mymidrule
Gemma3-12B~\cite{team2025gemma} & 52.6 & --& --& 49.7 & 58.4 & --& --& 59.8 & 48.9 & 72.8 & 60.4 & 60.5 \\
InternVL3-14B~\cite{zhu2025internvl3}                              & 60.3   & 76.0    & 50.9  & 56.2  & 70.3 & 59.3  & 62.2    & 70.7  & 64.0   & 69.8    & 69.3 & 68.5    \\
\rowcolor{oursgray} InternVL3.5-14B & 57.6 & 73.7 & 51.3 & 58.0 & 68.0 & 59.0 & 61.3 & 70.5 & 63.2 & 73.0 & 70.9 & 69.4  \\

\mymidrule
Kimi-VL-A3B-2506~\cite{team2025kimi} & 56.8 & -- & -- & 64.6 & 63.9 & -- & -- & 72.4 & 54.5 & 64.8 & 69.3 & 65.3\\
\rowcolor{oursgray} InternVL3.5-20B-A4B & 59.0 & 74.2 & 50.5 & 58.5 & 66.6 & 54.4 & 60.5 & 71.2 & 60.0 & 69.6 & 71.5 & 68.1 \\
\rowcolor{oursgray} InternVL3.5-30B-A3B & 60.4 & 72.8 & 55.1 & 53.1 & 66.6 & 59.0 & 61.2 & 72.3 & 64.8 & 75.8 & 70.7 & 70.9  \\

\mymidrule
Gemma3-27B~\cite{team2025gemma} & 54.4 & --& --& 55.2 & 61.8 & --& --& 65.1 & 51.6 & 79.8 & 61.6 & 64.5 \\
Qwen2.5-VL-32B~\cite{bai2025qwen2_5} & 61.2 & --& --& 64.5 & 66.4 & --& --& 71.2 & 60.3 & 85.2 & 69.5 & 71.6 \\
Cambrian-34B~\cite{tong2024cambrian}       & --     & --      & --    & --    & --   & --    & --      & 67.8  & 44.1   & --      & --   & --      \\
InternVL3-38B~\cite{zhu2025internvl3}                              & 64.0   & 77.9    & 57.4  & 63.8  & 71.8 & 62.3  & 66.2    & 75.6  & 67.3   & 71.6    & 73.3 & 72.0    \\
\rowcolor{oursgray} InternVL3.5-38B & 60.9 & 77.4 & 58.9 & 63.7 & 71.8 & 71.9 & 67.4 & 75.9 & 66.0 & 80.0 & 73.1 & 73.8  \\
\mymidrule
GPT-4V~\cite{gpt4v}                        & 54.6   & 62.7    & --    & 62.3  & 64.3 & 53.1  & --      & 61.4  & --     & 71.8    & 65.6 & --      \\
GPT-4o-20240513~\cite{chatgpt4o}               & 68.0   & --      & 55.7  & 68.0  & 65.4 & --    & --      & 75.4  & 45.2   & 80.6    & 77.7 & 69.7    \\
Claude-3.5-Sonnet~\cite{claude3series2024} & --     & --      & 53.4  & --    & --   & --    & --      & 60.1  & 51.6   & --      & --   & --      \\
Gemini-1.5-Pro~\cite{reid2024gemini1_5}    & --     & --      & 53.4  & --    & 64.5 & --    & --      & 67.5  & 38.2   & --      & --   & --      \\
GLM-4.5V~\cite{hong2025glm_v_thinking} & 65.3 & -- & -- & 75.3 & 70.9 & -- & -- & 75.2 & 61.7 & 79.0 & 72.9 & 72.2 \\
Qwen2-VL-72B~\cite{wang2024qwen2vl}        & --     & --      & --    & --    & 71.8 & --    & --      & 77.8  & --     & --      & --   & --      \\
Qwen2.5-VL-72B~\cite{bai2025qwen2_5}         & 64.4   & --      & --    & 70.7  & --   & --    & --      & 75.7  & 63.2   & --      & --   & --      \\
InternVL3-78B~\cite{zhu2025internvl3}                              & 66.3   & 79.3    & 60.4  & 64.5  & 73.2 & 64.3  & 68.0    & 78.0  & 65.4   & 73.6    & 77.4 & 73.6    \\
\rowcolor{oursgray} InternVL3.5-241B-A28B & 61.4  & 77.0  & 61.3 & 47.5  & 72.7  & 73.0  & 65.5  & 75.2  & 65.1  &  82.8  & 75.4  & 74.6  \\
\mybottomrule
\end{tabular}
}
\caption{\textbf{Comparison of multi-image and real-world understanding performance. }
Multi-image benchmarks include BLINK~\cite{fu2024blink}, Mantis-Eval~\cite{jiang2024mantis}, MMIU~\cite{meng2024mmiu}, MuirBench~\cite{wang2024muirbench}, MMT-Bench~\cite{mmtbench}, and MIRB~\cite{zhao2024mirb}.
Real-world benchmarks include RealWorldQA~\cite{realworldqa}, MME-RealWorld~\cite{zhang2024mme}, WildVision~\cite{lu2024wildvision}, and R-Bench~\cite{li2024r}.
Part of the results are sourced from the benchmark papers and the OpenCompass leaderboard~\cite{opencompass2023}.
}
\label{tab:benchmark_multi_image_real_world}
\end{table*}

\begin{table}[t]
\centering
\scriptsize
\setlength{\tabcolsep}{3.5pt}
\resizebox{\linewidth}{!}{
\begin{tabular}{l|cccc|c|ccc|c}
\mytoprule
\textbf{Model} & \makecell{\textbf{MME}\\\textbf{(sum)}} & \makecell{\textbf{MMB v1.1}\\\textbf{(EN)}} & \makecell{\textbf{MMVet}\\\textbf{(turbo)}} & \textbf{MMStar} & \textbf{Overall} & \makecell{\textbf{HallBench}\\\textbf{(avg)}} & \makecell{\textbf{CRPE}\\\textbf{(relation)}} & \makecell{\textbf{POPE}\\\textbf{(avg)}} & \textbf{Overall} \\
\mymidrule
InternVL3-1B~\cite{zhu2025internvl3} & 1934.4  & 69.9 & 59.5 & 51.5 & 62.5 & 41.4 & 64.0 & 90.7 & 65.4 \\
\rowcolor{oursgray} InternVL3.5-1B & 1910.2  & 69.9 & 56.5 & 51.9 & 61.6 & 41.0 & 68.4& 86.8& 65.4  \\
\mymidrule
Qwen2-VL-2B~\cite{wang2024qwen2vl} & 1872.0  & 72.2 & 49.5 & 48.0 & 59.1 & 41.7 & -- & -- & -- \\
Qwen2.5-VL-3B~\cite{bai2025qwen2_5} & 2157.0  & -- & -- & -- & -- & 46.3 & 73.6 & -- & -- \\
InternVL3-2B~\cite{zhu2025internvl3} & 2221.2  & 78.6 & 62.2 & 60.7 & 70.2 & 42.5 & 71.5 & 89.6 & 67.9 \\
\rowcolor{oursgray} InternVL3.5-2B & 2123.3  & 76.6 & 71.7 & 62.7 & 71.7 & 48.6 & 75.6 & 87.2 & 70.5 \\

\mymidrule
MiniCPM-V-4-4B~\cite{yao2024minicpm} & 2167.7 & 79.1& 56.6& 59.0& 68.0 & 46.9 & 74.3& 82.4& 67.9 \\
\rowcolor{oursgray} InternVL3.5-4B & 2272.3 & 80.3 & 76.6 & 65.0 & 75.8 & 44.8 & 75.0 & 88.9 & 69.6 \\

\mymidrule
Qwen2-VL-7B~\cite{wang2024qwen2vl} & 2326.8  & 80.7 & 62.0 & 60.7 & 71.6 & 50.6 & 74.4 & 88.1 & 71.0 \\
Qwen2.5-VL-7B~\cite{bai2025qwen2_5} & 2347.0  & 82.6 & 67.1 & 63.9 & 74.4 & 52.9 & 76.4 & 86.4 & 71.9 \\
MiniCPM-V2.6~\cite{yao2024minicpm} & 2348.4  & 78.0 & 60.0 & 57.5 & 69.8 & 48.1 & 75.2 & 87.3 & 70.2 \\
Keye-VL-8B~\cite{team2025keye_vl} & 2214.7 & 76.3 & 70.0 & 72.8 & 74.5 & 57.7& 76.5& 87.0& 73.7 \\
GLM-4.1V-9B~\cite{hong2025glm_v_thinking}  & 2445.8  & 85.8 & 66.4 & 72.9 & 78.1 & 63.2 & 78.9 & 87.1 & 76.4 \\
InternVL3-8B~\cite{zhu2025internvl3} & 2415.4   & 81.7  & 81.3  & 68.2  & 79.4  & 49.9 & 76.3 & 91.1 & 72.4   \\
InternVL3-9B~\cite{zhu2025internvl3} & 2372.8  & 81.7 & 76.2 & 66.3 & 77.2 & 51.2 & 75.0 & 90.4 & 72.2 \\
\rowcolor{oursgray} InternVL3.5-8B & 2380.6  & 79.5 & 83.1 & 69.3 & 79.2 & 54.5& 75.1& 88.7& 72.8 \\

\mymidrule
Gemma3-12B~\cite{team2025gemma} & 2044.7  & 71.8  & 64.9  & 56.1  & 66.5  & 47.2  & 69.1 & 85.2 & 67.2 \\
InternVL3-14B~\cite{zhu2025internvl3} & 2478.3  & 83.5 & 80.2 & 68.8 & 80.3 & 55.1 & 77.3 & 90.2 & 74.2 \\
\rowcolor{oursgray} InternVL3.5-14B & 2398.4  & 81.5 & 81.7 & 70.4 & 79.8 & 54.0 & 76.6 & 87.7 & 72.8 \\

\mymidrule
Kimi-VL-A3B-2506~\cite{team2025kimi} &  2352.7  & 84.4 & 78.1 & 70.4 & 79.2 & 59.8  & 60.9 & 86.5 & 69.1\\
\rowcolor{oursgray} InternVL3.5-20B-A4B & 2318.1  & 85.2 & 80.3 & 70.4 & 79.7 & 52.0 & 76.8 & 89.4 & 72.7 \\
\rowcolor{oursgray} InternVL3.5-30B-A3B & 2461.9  & 84.8 & 85.5 & 72.0 & 82.6 & 53.8 & 77.6 & 89.6 & 73.7 \\

\mymidrule
Gemma3-27B~\cite{team2025gemma} & 1816.9    & 77.2  & 68.3  & 61.7  & 68.0  & 47.9  & 71.64 & 85.8 & 68.4  \\
Qwen2.5-VL-32B~\cite{bai2025qwen2_5} & 2402.9 & 85.2 & 69.6 & 67.8 & 77.1 & 53.7& 77.0 & 86.8 & 72.5 \\
Cambrian-34B~\cite{tong2024cambrian} & --  & 78.3 & 53.2 & 54.2 & -- & 41.6 & -- & -- & -- \\
InternVL3-38B~\cite{zhu2025internvl3} & 2523.6  & 86.9 & 83.9 & 71.5 & 83.1 & 57.1 & 77.1 & 90.6 & 74.9 \\
\rowcolor{oursgray} InternVL3.5-38B & 2492.4 & 87.3 & 82.2 & 75.3 & 83.5 & 59.7 & 77.7 & 90.4 & 75.9  \\

\mymidrule
GPT-4V~\cite{gpt4v} & 1926.6  & 80.0 & 67.5 & 56.0 & 68.1 & 46.5 & -- & -- & -- \\
GPT-4o-20240513~\cite{chatgpt4o} & --  & 83.1 & 69.1 & 64.7 & -- & 55.0 & 76.6 & 86.9 & 72.8 \\
Claude-3-Opus~\cite{claude3series2024} & 1586.8  & 60.1 & 51.7 & 45.7 & 53.5 & 37.8 & -- & -- & -- \\
Claude-3.5-Sonnet~\cite{claude3series2024} & -- & 80.9 & 70.1 & 65.1 & -- & 55.5 & -- & -- & -- \\
Gemini-1.5-Pro~\cite{reid2024gemini1_5} & --  & 74.6 & 64.0 & 59.1 & -- & 45.6 & -- & -- & -- \\
GLM-4.5V~\cite{hong2025glm_v_thinking} & 2423.8 & 88.2 & 75.2 & 75.3 & 81.3 & 64.5 & 55.4 & 85.9 & 68.6\\
Qwen2-VL-72B~\cite{wang2024qwen2vl} & 2482.7  & 85.9 & 74.0 & 68.3 & 79.2 & 58.1 & -- & -- & -- \\
Qwen2.5-VL-72B~\cite{bai2025qwen2_5} & 2448.0 & 88.4 & 76.2 & 70.8 & 80.7 & 55.2 & 79.2 & -- & -- \\
InternVL3-78B~\cite{zhu2025internvl3} & 2549.8  & 87.7 & 81.3 & 72.5 & 83.1 & 59.1 & 79.2 & 90.3 & 76.2 \\
\rowcolor{oursgray} InternVL3.5-241B-A28B & 2525.9    & 87.4  & 81.2  & 77.9  & 84.2  & 57.3  & 78.0  & 90.7  & 75.3  \\
\mybottomrule
\end{tabular}
}
\vspace{2mm}
\caption{\textbf{Comparison of comprehensive multimodal understanding and hallucination performance.}
Comprehensive multimodal benchmarks include MME~\cite{fu2023mme}, MMBench~\cite{liu2023mmbench}, MMVet~\cite{yu2023mmvet}, and MMStar~\cite{chen2024mmstar}.
Hallucination-related benchmarks encompass HallusionBench~\cite{guan2023hallusionbench}, CRPE~\cite{wang2024allseeingv2}, and POPE~\cite{li2023pope}.
Part of the results are sourced from the benchmark papers and the OpenCompass leaderboard~\cite{opencompass2023}.
When calculating Overall, the score of MME is normalized from 0-2800 to 0-100.
}
\label{tab:exp-comprehensive-hallucination}
\vspace{-2mm}
\end{table}

\begin{table*}[t] 
\centering 
\scriptsize
\renewcommand{\arraystretch}{1.2}
\begin{tabular}{l|ccc|ccc|cc|c}
\mytoprule
\multirow{2}{*}{{Model Name}}           & \multicolumn{3}{c|}{{\textbf{RefCOCO}}} & \multicolumn{3}{c|}{{\textbf{RefCOCO+}}} & \multicolumn{2}{c|}{{\textbf{RefCOCOg}}} & \multirow{2}{*}{{\textbf{Overall}}} \\
                                        & \textbf{val}    & \textbf{test-A}    & \textbf{test-B}    & \textbf{val}    & \textbf{test-A}    & \textbf{test-B}     & \textbf{val}     & \textbf{test}                 &         \\
\mymidrule
Grounding-DINO-L~\citep{grounding_dino} & 90.6   & 93.2      & 88.2      & 82.8   & 89.0      & 75.9       & 86.1    & 87.0                 & 86.6    \\
UNINEXT-H~\citep{uninext}               & 92.6   & 94.3      & 91.5      & 85.2   & 89.6      & 79.8       & 88.7    & 89.4                 & 88.9    \\
ONE-PEACE~\citep{one-peace}             & 92.6   & 94.2      & 89.3      & 88.8   & 92.2      & 83.2       & 89.2    & 89.3                 & 89.8    \\

\mymidrule
InternVL3-1B~\cite{zhu2025internvl3}    & 85.8   & 90.1      & 81.7      & 76.6   & 84.1 & 69.2 & 82.8    & 82.6   & 81.6    \\
\rowcolor{oursgray} InternVL3.5-1B & 85.4  & 89.7  & 80.2  & 77.7  & 85.5  & 69.5  & 81.9  & 81.6  & 81.4  \\

\mymidrule
InternVL3-2B~\cite{zhu2025internvl3}                             & 89.8   & 92.6 & 86.4  & 84.0 & 89.2      & 76.5       & 87.6    & 87.2                 & 86.7    \\
\rowcolor{oursgray} InternVL3.5-2B & 88.7 & 91.6 & 84.8 & 82.7 & 88.4 & 76.6 & 85.6 & 85.5 & 85.5  \\

\mymidrule
Qwen2.5-VL-3B~\cite{bai2025qwen2_5}       & 89.1   & 91.7      & 84.0      & 82.4   & 88.0      & 74.1       & 85.2    & 85.7                 & 85.0    \\
\rowcolor{oursgray} InternVL3.5-4B & 92.5 & 94.3 & 88.2 & 87.6 & 92.3 & 81.6 & 89.6 & 89.3 & 89.4  \\

\mymidrule
Shikra-7B~\citep{chen2023shikra}        & 87.0   & 90.6      & 80.2      & 81.6   & 87.4      & 72.1       & 82.3    & 82.2                 & 82.9    \\
CogVLM-Grounding~\citep{wang2023cogvlm} 
                                        & 92.8   & 94.8      & 89.0      & 88.7   & 92.9      & 83.4       & 89.8    & 90.8                 & 90.3    \\
Qwen2-VL-7B~\cite{wang2024qwen2vl}      & 91.7   & 93.6      & 87.3      & 85.8   & 90.5      & 79.5       & 87.3    & 87.8                 & 87.9    \\
Qwen2.5-VL-7B~\cite{bai2025qwen2_5}       & 90.0   & 92.5      & 85.4      & 84.2   & 89.1      & 76.9       & 87.2    & 87.2                 & 86.6    \\
TextHawk2~\citep{yu2024texthawk2}       & 91.9   & 93.0      & 87.6      & 86.2   & 90.0      & 80.4       & 88.2    & 88.1                 & 88.2    \\
InternVL3-8B~\cite{zhu2025internvl3}                            & 92.5   & 94.6      & 88.0      & 88.2   & 92.5      & 81.8       & 89.6    & 90.0                 & 89.6    \\
InternVL3-9B~\cite{zhu2025internvl3}                            & 91.8  & 93.2     & 86.6     & 86.4  & 91.0     & 79.9      & 88.0   & 88.5                & 88.2    \\  
\rowcolor{oursgray} InternVL3.5-8B & 92.4 & 94.7 & 88.7 & 87.9 & 92.4 & 82.4 & 89.6 & 89.4 & 89.7  \\

\mymidrule
Ferret-v2-13B~\citep{ferretv2}          & 92.6   & 95.0      & 88.9      & 87.4   & 92.1      & 81.4       & 89.4    & 90.0                 & 89.6    \\
InternVL3-14B~\cite{zhu2025internvl3}                           & 92.0  & 94.4     & 87.8     & 87.4  & 92.1     & 81.5      & 88.6   & 89.3                 & 89.1    \\  
\rowcolor{oursgray} InternVL3.5-14B & 92.6 & 94.7 & 89.4 & 88.3 & 92.7 & 82.5 & 90.1 & 90.5 & 90.1 \\
\rowcolor{oursgray} InternVL3.5-20B-A4B & 91.9 & 94.1 & 88.8 & 87.6 & 92.0 & 82.7 & 89.1 & 90.0 & 89.5  \\
\rowcolor{oursgray} InternVL3.5-30B-A3B & 93.1 & 95.4 & 90.1 & 89.6 & 93.2 & 84.4 & 90.6 & 91.0 & 90.9  \\
\mymidrule

InternVL3-38B~\cite{zhu2025internvl3}                           & 93.2   & 95.1      & 90.2      & 89.8   & 93.2      & 85.2       & 91.4    & 91.5                 & 91.2    \\
\rowcolor{oursgray} InternVL3.5-38B & 90.3 & 91.8 & 89.0 & 87.5 & 90.0 & 84.7 & 89.7 & 89.9 & 89.1  \\

\mymidrule
Qwen2-VL-72B~\cite{wang2024qwen2vl}     & 93.2   & 95.3      & 90.7      & 90.1   & 93.8      & 85.6       & 89.9    & 90.4                 & 91.1    \\
Qwen2.5-VL-72B~\cite{bai2025qwen2_5}      & 92.7   & 94.6      & 89.7      & 88.9   & 92.2      & 83.7       & 89.9    & 90.3                 & 90.3    \\
InternVL3-78B~\cite{zhu2025internvl3}                           & 93.4   & 95.4      & 90.3      & 90.1   & 93.8      & 85.3       & 91.5    & 91.5                 & 91.4    \\
\rowcolor{oursgray} InternVL3.5-241B-A28B & 94.1  & 96.3  & 91.5  & 91.6  & 94.6  & 86.9  & 92.0  & 92.1  & 92.4 \\

\mybottomrule
\end{tabular}
\caption{\textbf{Comparison of visual grounding performance.}
We evaluate InternVL3.5's visual grounding capability on RefCOCO, RefCOCO+, and RefCOCOg datasets~\cite{kazemzadeh2014referitgame, mao2016generation}. Part of the results are collected from \cite{wang2024qwen2vl}.
} 
\label{tab:benchmark-grounding}
\end{table*}

\begin{table}[t]
\centering
\small
\setlength{\tabcolsep}{3pt}
\renewcommand\arraystretch{1.2}
\resizebox{\linewidth}{!}{
\begin{tabular}{l|cccccc|cccccc|c|c}
\mytoprule
\multirow{2}{*}{\textbf{Model}} 
& \multicolumn{6}{c|}{\textbf{MMMB}} 
& \multicolumn{6}{c|}{\textbf{Multilingual MMBench}} 
& \textbf{MTVQA} 
& \multirow{2}{*}{\textbf{Overall}} \\
& \textbf{en} & \textbf{zh} & \textbf{pt} & \textbf{ar} & \textbf{tr} & \textbf{ru} 
& \textbf{en} & \textbf{zh} & \textbf{pt} & \textbf{ar} & \textbf{tr} & \textbf{ru} 
& \textbf{(avg)} & \\
\mymidrule
InternVL3-1B~\cite{zhu2025internvl3} & 79.4 & 70.1 & 62.3 & 58.0 & 47.6 & 61.9 & 72.6 & 66.2 & 62.3 & 48.0 & 39.5 & 60.3 & 22.2 & 47.9 \\
\rowcolor{oursgray} InternVL3.5-1B & 77.0  & 73.1  & 67.2  & 59.0  & 53.5  & 66.3  & 71.2  & 66.3  & 61.7  & 45.8 & 45.7 & 60.2 & 22.9 & 49.1  \\
\mymidrule
Qwen2-VL-2B~\cite{wang2024qwen2vl} & 78.3 & 74.2 & 72.6 & 68.3 & 61.8 & 72.8 & 72.1 & 71.1 & 69.9 & 61.1 & 54.4 & 69.3 & 20.0 & 52.6 \\
Qwen2.5-VL-3B~\cite{bai2025qwen2_5} & -- & -- & -- & -- & -- & -- & -- & -- & -- & -- & -- & -- & 24.8 & -- \\
InternVL3-2B~\cite{zhu2025internvl3} & 81.9 & 78.3 & 75.4 & 68.6 & 62.9 & 74.6 & 81.3 & 77.8 & 75.9 & 66.4 & 59.5 & 70.7 & 26.7 & 57.4 \\
\rowcolor{oursgray} InternVL3.5-2B & 80.2 & 77.7 & 75.9 & 68.5 & 69.1 & 76.3 & 78.4 & 75.9 & 73.7 & 63.7 & 62.0 & 71.4 & 28.5 & 58.0 \\

\mymidrule
MiniCPM-V-4-4B~\cite{yao2024minicpm} & 82.0  & 80.2  & 75.6  & 60.1  & 63.8  & 71.7  & 80.8  & 80.4  & 71.6  & 51.9  & 59.1  & 67.7  & 22.6  & 54.5  \\
\rowcolor{oursgray} InternVL3.5-4B & 84.3  & 82.6  & 81.0  & 76.4  & 75.2  & 81.4  & 81.5  & 81.1 & 76.7 & 71.0 & 72.4 & 75.7 & 29.6 & 62.1  \\

\mymidrule
mPLUG-Owl2~\cite{ye2023mplug2} & 67.3 & 61.0 & 59.7 & 45.8 & 45.4 & 62.6 & 66.2 & 59.4 & 58.2 & 37.9 & 47.7 & 60.4 & -- & -- \\
Qwen2-VL-7B~\cite{wang2024qwen2vl} & 83.9 & 82.4 & 81.2 & 79.0 & 74.7 & 82.4 & 81.8 & 81.6 & 79.1 & 75.6 & 74.5 & 79.3 & 25.6 & 61.6 \\
Qwen2.5-VL-7B~\cite{bai2025qwen2_5} & -- & -- & -- & -- & -- & -- & -- & -- & -- & -- & -- & -- & 29.2 & -- \\
Keye-VL-8B~\cite{team2025keye_vl} & 66.8  & 83.0  & 74.1  & 73.8  & 72.0  & 76.8  & 53.9  & 87.8 & 57.4 & 67.2 & 67.14 & 68.7 & 22.3 & 54.6 \\
GLM-4.1V-9B~\cite{hong2025glm_v_thinking} & 82.6 & 83.6 & 79.4 & 80.4 & 80.4 & 82.9 & 83.0 & 86.0 & 79.8 & 78.8 & 78.5 & 82.0 & 25.5 & 62.8 \\
InternVL3-8B~\cite{zhu2025internvl3} & 85.1 & 83.1 & 82.5 & 81.6 & 76.2 & 83.4 & 85.5 & 85.6 & 83.2 & 79.2 & 75.9 & 82.6 & 30.2 & 64.7 \\
InternVL3-9B~\cite{zhu2025internvl3} & 84.8 & 83.7 & 80.6 & 69.9 & 68.5 & 80.8 & 86.5 & 85.2 & 79.1 & 64.3 & 68.3 & 79.1 & 27.1 & 60.7 \\
\rowcolor{oursgray} InternVL3.5-8B & 84.9 & 83.0 & 81.4 & 79.6 & 77.4 & 82.1 & 82.5 & 80.7 & 79.0 & 75.9 & 74.8 & 77.6 & 35.2 & 65.0 \\

\mymidrule
Gemma3-12B~\cite{team2025gemma} & 77.6  & 77.3  & 77.3  & 73.4  & 73.9  & 75.9  & 72.1  & 72.5  & 68.3  & 53.8  & 60.5  & 60.4  & 24.4  & 55.0 \\
InternVL3-14B~\cite{zhu2025internvl3} & 85.7 & 84.7 & 83.1 & 83.7 & 79.3 & 83.6 & 86.7 & 85.8 & 83.2 & 81.1 & 80.7 & 83.8 & 31.6 & 66.2 \\
\rowcolor{oursgray} InternVL3.5-14B & 85.1 & 84.1 & 82.7 & 80.3 & 79.4 & 83.5 & 84.0 & 83.7 & 80.0 & 77.8& 77.0& 77.0& 34.2& 65.5  \\

\mymidrule
Kimi-VL-A3B-2506~\cite{team2025kimi} & 83.1 & 78.9 & 76.9 & 71.3 & 71.0 & 76.3 & 82.3 & 79.9 & 76.9 & 62.8 & 66.7 & 73.8 & 27.2 & 59.1 \\
\rowcolor{oursgray} InternVL3.5-20B-A4B & 85.1 & 83.1 & 83.2 & 82.2 & 80.4 & 83.8 & 85.3 & 84.2 & 82.1 & 78.9 & 79.5 & 82.5 & 28.2 &  64.4  \\
\rowcolor{oursgray} InternVL3.5-30B-A3B & 86.4 & 85.7 & 83.4 & 83.3 & 81.7 & 85.0 & 86.1 & 86.3 & 83.1& 82.3& 81.5& 83.5& 33.7& 67.3  \\

\mymidrule
Gemma3-27B~\cite{team2025gemma} & 78.9  & 77.8  & 77.8  & 75.4  & 76.1  & 76.9 & 79.0 & 77.2 & 74.5 & 71.8 & 74.0 & 74.1 & 27.5 & 59.9 \\
Qwen2.5-VL-32B~\cite{bai2025qwen2_5} & 85.2  & 83.0  & 81.9  & 81.8  & 79.9  & 83.5 & 88.1 & 85.9 & 80.2 & 81.6 & 79.7 & 84.5 & 31.4 & 65.7  \\
InternVL3-38B~\cite{zhu2025internvl3} & 86.7  & 85.6  & 84.5  & 84.8  & 82.6  & 85.1  & 89.0  & 89.3 & 87.1 & 84.6 & 84.3 & 87.4 & 32.4 & 68.1   \\
\rowcolor{oursgray} InternVL3.5-38B & 86.7 & 85.5 & 85.1 & 84.1 & 84.3 & 85.3 & 87.4 & 86.9 & 84.2 & 82.0& 83.4& 85.6& 36.1& 68.7 \\

\mymidrule
GPT-4V~\cite{gpt4v} & 75.0 & 74.2 & 71.5 & 73.5 & 69.0 & 73.1 & 77.6 & 74.4 & 72.5 & 72.3 & 70.5 & 74.8 & 22.0 & 56.1 \\
GPT-4o~\cite{chatgpt4o} & -- & -- & -- & -- & -- & -- & -- & -- & -- & -- & -- & -- & 27.8 & -- \\
Gemini-1.0-Pro~\cite{team2023gemini}  & 75.0 & 71.9 & 70.6 & 69.9 & 69.6 & 72.7 & 73.6 & 72.1 & 70.3 & 61.1 & 69.8 & 70.5 & -- & -- \\
GLM-4.5V~\cite{hong2025glm_v_thinking} & 87.1 & 86.9 & 84.8 & 84.5 & 84.6 & 84.3 & 89.1 & 89.3 & 86.9 & 83.7 & 84.0 & 87.2 & 30.5 & 67.5  \\
Qwen2-VL-72B~\cite{wang2024qwen2vl} & 86.8 & 85.3 & 85.2 & 84.8 & 84.2 & 85.3 & 86.9 & 87.2 & 85.8 & 83.5 & 84.4 & 85.3 & 30.9 & 67.2 \\
Qwen2.5-VL-72B~\cite{bai2025qwen2_5} & -- & -- & -- & -- & -- & -- & -- & -- & -- & -- & -- & -- & 31.7 & -- \\
InternVL3-78B~\cite{zhu2025internvl3} & 87.2 & 86.6 & 85.5 & 86.5 & 84.6 & 86.1 & 89.4 & 90.3 & 88.7 & 86.1 & 86.6 & 88.1 & 32.5 & 68.9 \\
\rowcolor{oursgray} InternVL3.5-241B-A28B & 87.6  & 86.4  & 85.3  & 84.2  & 85.1  & 86.0  & 88.9  & 87.7 & 87.0 & 86.5 & 86.7 & 87.6 & 39.3 & 70.8  \\
\mybottomrule
\end{tabular}
}
\vspace{2mm}
\caption{\textbf{Comparison of multimodal multilingual performance.} 
We evaluate multilingual capabilities across 3 benchmarks, including MMMB~\cite{sun2024parrot}, Multilingual MMBench~\cite{sun2024parrot} and MTVQA~\cite{tang2024mtvqa} with six languages: English (en), Chinese (zh), Portuguese (pt), Arabic (ar), Turkish (tr), and Russian (ru).
}
\label{tab:exp-multilingual}
\end{table}

\begin{table*}[t!]
\centering
\scriptsize
\setlength\tabcolsep{3pt}
\renewcommand{\arraystretch}{1.2}
\newcommand{\VMME}{\makecell{\textbf{Video-MME}\\\textbf{(wo / w sub)}}}
\newcommand{\MVB}{\makecell{\textbf{MVBench}}}
\newcommand{\MMBV}{\makecell{\textbf{MMBench-Video}\\\textbf{(val)}}}
\newcommand{\MLVU}{\makecell{\textbf{MLVU}\\\textbf{(M-Avg)}}}
\newcommand{\LVB}{\makecell{\textbf{LongVideoBench}\\\textbf{(val total)}}}
\newcommand{\CG}{\makecell{\textbf{CG-Bench}\\\textbf{(long / clue acc.)}}}
\begin{tabular}{l|ccccc|c}
\mytoprule
\textbf{Model Name}                                 &    \VMME    & \MVB & \MMBV & \MLVU & \LVB  & \textbf{Overall} \\
\mymidrule

InternVL3-1B~\cite{zhu2025internvl3}       & 51.0 / 53.0 & 63.1 & 1.30   & 53.0  & 48.1  & 51.9    \\
\rowcolor{oursgray} InternVL3.5-1B         & 52.4 / 55.0 & 61.0  & 1.39  & 56.6   & 53.0  & 54.1  \\
\mymidrule
Qwen2-VL-2B~\cite{wang2024qwen2vl}         & 55.6 / 60.4 & 63.2 & --    & --    & --    & --      \\
Qwen2.5-VL-3B~\cite{bai2025qwen2_5}        & 61.5 / 67.6 & 67.0 & 1.63  & 68.2  & 43.3  & 60.3      \\
InternVL3-2B~\cite{zhu2025internvl3}       & 58.9 / 61.4 & 70.4 & 1.42  & 64.2  & 55.4  & 59.6    \\
\rowcolor{oursgray} InternVL3.5-2B         & 58.4 / 61.9 & 65.9 & 1.56  & 64.4  & 57.4  & 60.0 \\

\mymidrule
MiniCPM-V-4-4B~\cite{yao2024minicpm}          & 61.2 / 65.8 & 58.7 &    -- & --   & -- & -- \\
\rowcolor{oursgray} InternVL3.5-4B         & 65.4 / 68.6 & 71.2 & 1.59  & 70.4  & 60.8 & 64.9 \\
\mymidrule

VideoChat2-HD~\cite{li2023videochat}       & 45.3 / 55.7 & 62.3 & 1.22  & 47.9  & --    & --      \\
LLaVA-OneVision-7B~\cite{li2024llavaov}    & 58.2 / \rsp & 56.7 & --    & --    & --    & --      \\
MiniCPM-V-2.6~\cite{yao2024minicpm}        & 60.9 / 63.6 & --   & 1.70  & --    & 54.9  & --      \\
Qwen2-VL-7B~\cite{wang2024qwen2vl}         & 63.3 / 69.0 & 67.0 & 1.44  & --    & 55.6  & --      \\
Qwen2.5-VL-7B~\cite{bai2025qwen2_5}        & 65.1 / 71.6 & 69.6 & 1.79  & 70.2  & 45.3  & 63.6      \\
Keye-VL-8B~\cite{team2025keye_vl}          & 67.7 / \rsp & --   & --    & --  & 64.8 & -- \\
GLM-4.1V-9B~\cite{team2025keye_vl}          & 68.2 / 73.6 &  68.4   &  1.63    &  71.5  &  65.7 &  67.0 \\
InternVL3-8B~\cite{zhu2025internvl3}       & 66.3 / 68.9 & 75.4 & 1.69  & 71.4  & 58.8  & 66.2    \\
InternVL3-9B~\cite{zhu2025internvl3}       & 66.7 / 68.9 & 74.3 & 1.69  & 70.8  & 62.5  & 66.6    \\
\rowcolor{oursgray} InternVL3.5-8B         & 66.0 / 68.6 & 72.1 & 1.67  & 70.2  & 62.1 &  65.8   \\
\mymidrule

InternVL3-14B~\cite{zhu2025internvl3}      & 70.4 / 73.0 & 76.6 & 1.73  & 73.3  & 63.9  & 69.1    \\
\rowcolor{oursgray} InternVL3.5-14B        & 67.9 / 71.0     &  72.8     &   1.73    &  72.1     &  62.7  & 67.4     \\
\mymidrule

Kimi-VL-A3B-2506~\cite{team2025kimi}      & 67.8 / 72.6 & 59.7 & --  & 74.2  & 64.5  & --    \\
\rowcolor{oursgray} InternVL3.5-20B-A4B    & 62.4 / 64.9 & 73.3 & 1.54  & 65.6  & 58.3 & 62.6    \\
\rowcolor{oursgray} InternVL3.5-30B-A3B    & 68.7 / 71.8 & 72.1 & 1.69  & 73.0  & 63.8 & 67.6    \\
\mymidrule

Oryx-1.5-32B~\cite{liu2024oryx}            & 67.3 / 74.9 & 70.1 & 1.52  & 72.3  & --    & --      \\  
Qwen2.5-VL-32B~\cite{bai2025qwen2_5}       & 70.5 / 77.9 & --   & 1.93  & --    & --    & --      \\
VILA-1.5-40B~\cite{lin2024vila}            & 60.1 / 61.1 & --   & 1.61  & 56.7  & --    & --      \\
InternVL3-38B~\cite{zhu2025internvl3}      & 72.7 / 75.0 & 76.9 & 1.81  & 77.8  & 67.3  & 71.7    \\
\rowcolor{oursgray} InternVL3.5-38B        & 70.9 / 74.2 & 75.0 &  1.90     &  77.0     & 65.7 & 71.0       \\

\mymidrule

GPT-4V/4T~\cite{openai2023gpt4}            & 59.9 / 63.3 & 43.7 & 1.53  & 49.2  & 59.1  & 54.4      \\
GPT-4o-20240513~\cite{gpt4v}               & 71.9 / 77.2 & --   & 1.63  & 64.6  & 66.7  & --      \\
GPT-4o-20240806~\cite{chatgpt4o}           & --          & --   & 1.87  & --    & --    & --      \\
Gemini-1.5-Pro~\cite{reid2024gemini1_5}    & 75.0 / 81.3 & --   & 1.30  & --    & 64.0  & --      \\
GLM-4.5V~\cite{li2024llavaov}   & 74.6 / 80.7 & 73.0 & 2.05    & 75.3  & 68.8  & 73.5 \\
VideoLLaMA2-72B~\cite{cheng2024videollama2}& 61.4 / 63.1 & 62.0 & --    & --    & --    & --      \\
LLaVA-OneVision-72B~\cite{li2024llavaov}   & 66.2 / 69.5 & 59.4 & --    & 66.4  & 61.3  & --      \\
Qwen2-VL-72B~\cite{wang2024qwen2vl}        & 71.2 / 77.8 & 73.6 & 1.70  & --    & --    & --      \\
Qwen2.5-VL-72B~\cite{bai2025qwen2_5}       & 73.3 / 79.1 & 70.4 & 2.02  & 74.6  & 60.7  & 70.9      \\
InternVL3-78B~\cite{zhu2025internvl3}      & 72.7 / 75.7 & 78.7 & 1.81  & 79.5  & 65.7  & 72.1    \\ 
\rowcolor{oursgray} InternVL3.5-241B-A28B  & 72.9 / 76.0 & 76.5 & 1.74 & 78.2 &  67.1  & 71.4 \\
\mybottomrule
\end{tabular}
\caption{\textbf{Comparison of video understanding performance.}
We evaluate InternVL3.5's video understanding capabilities across 5 benchmarks.
For Video-MME~\cite{fu2024video}, MMBench-Video~\cite{fang2024mmbench}, MLVU~\cite{MLVU}, and LongVideoBench~\cite{wu2024longvideobench}, we test with four different settings: 16, 32, 48, and 64 frames, and report the maximum results.
For MVBench~\cite{li2024mvbench}, we conduct testing using 16 frames.
When calculating Overall, the score of MMBench-Video is normalized from 0-3 to 0-100.
}
\label{tab:benchmark_video}
\end{table*}

\subsection{Multi-Image Understanding}
\label{sec:exp-multi-image}

To assess InternVL3’s ability to understand and reason over multiple images —-- a key aspect of multimodal interaction —-- we conduct comprehensive evaluations on a suite of widely recognized benchmarks, including BLINK~\cite{fu2024blink}, Mantis-Eval~\cite{jiang2024mantis}, MMIU~\cite{meng2024mmiu}, MuirBench~\cite{wang2024muirbench}, MMT-Bench~\cite{mmtbench}, and MIRB~\cite{zhao2024mirb}. These benchmarks evaluate critical skills such as cross-image reasoning and context integration, which are essential for effective multimodal systems.

As shown in Table~\ref{tab:benchmark_multi_image_real_world}, across various model scales, InternVL3.5 consistently outperforms other open-source and closed-source counterparts, including earlier versions such as InternVL3.
For example, InternVL3.5-38B achieves an overall score of 67.4, which is higher than the 66.2 achieved by InternVL3-38B.  %
On the lightweight scale, InternVL3.5-2B achieves an overall score of 50.5, and its performance on individual benchmarks is 51.3 on BLINK, 58.5 on Mantis, 44.9 on MMIU, 44.0 on Muir, 58.5 on MMT and 45.9 on MIRB. 
Furthermore, larger model sizes lead to significant improvements in multi-image understanding capabilities.
When the model is scaled up to InternVL3.5-4B, the overall score increases to 57.2, with scores of 58.1 on BLINK, 62.7 on Mantis, 49.2 on MMIU, 53.1 on Muir, 64.3 on MMT, and 55.9 on MIRB. 
As the model size continues to grow, performance across all benchmarks improves consistently, with InternVL3.5-8B achieving an overall score of 59.9, InternVL3.5-14B reaching 61.3, and InternVL3.5-241B-A28B improving to 65.5.

\subsection{Real-World Comprehension}
\label{sec:exp-real-world-comprehension}

To evaluate the performance of InternVL3.5 on realistic and complex tasks, we provide experimental results on four real-world comprehension benchmarks: RealWorldQA~\cite{realworldqa}, MME-RealWorld~\cite{zhang2024mme}, WildVision~\cite{lu2024wildvision}, and R-Bench~\cite{li2024r}. 

As shown in Table~\ref{tab:benchmark_multi_image_real_world}, InternVL3.5 achieves comparable or superior performance compared to existing methods, \textit{e.g.,} Qwen2.5-VL, MiniCPM-V-4 and Keye-VL. For example, the smallest variant InternVL3.5-1B demonstrates promising performance with a RealWorldQA score of 57.6, an MME-RealWorld score of 46.8, a WildVision win rate of 49.2, and an R-Bench score of 57.4. 
Scaling up the model results in further improvements, as larger models provide more robust representations and stronger comprehension capabilities in real-world scenarios.

At the higher end of the scale, the InternVL3.5-38B and InternVL3.5-241B-A28B models achieve top-tier results among the InternVL3.5 series. In particular, InternVL3.5-241B-A28B records an overall score of 74.6. Compared to competitive models, such as GPT-4o~\cite{gpt4v}—which scores 45.2 on MME-RealWorld and 80.6 on WildVision—the InternVL3.5 series exhibits competitive strengths. InternVL3.5-241B-A28B not only surpasses GPT-4o on RealWorldQA and closely matches its R-Bench performance but also considerably outperforms it on MME-RealWorld, indicating a robust overall performance on tasks demanding both perceptual precision and comprehensive understanding.

\subsection{Comprehensive Multimodal Understanding}
\label{sec:exp-comprehensive-eval}

In Table~\ref{tab:exp-comprehensive-hallucination}, we evaluate InternVL3.5 on a set of comprehensive multimodal understanding benchmarks, including MME~\cite{fu2023mme}, MMBench (English and Chinese)~\cite{liu2023mmbench}, MMBench v1.1 (English)~\cite{liu2023mmbench}, and MMVet~\cite{yu2023mmvet} and MMStar~\cite{chen2024mmstar}. 

We observe that InternVL3.5 outperforms existing methods like Keye-VL, Qwen2.5-VL and MiniCPM-V-4, especially on MMStar and MMVet.
For instance, InternVL3.5-4B achieves an MMVet score of 76.6 and MMStar of 65.0, compared to 56.6 and 59.0 of MiniCPM-V-4.
The improvements remain significant as model size grows, where InternVL3.5-241B-A28B finally achieves 87.4 on MMBench v1.1, 81.2 on MMVet, 77.9 on MMStar, and an overall score of 84.2.

We note that InternVL3.5 does not achieve a notable improvement compared to InternVL3. This is partly because the model's understanding performance has approached saturation, and also partly stems from our optimization of text and reasoning capabilities—which, while achieving improvements on relevant benchmarks, slightly impairs the performance of multimodal understanding.

\subsection{Multimodal Hallucination Evaluation}
\label{sec:exp-hallucination}

To evaluate the propensity for hallucination of InternVL3.5, we conduct experiments on three established benchmarks: HallusionBench~\cite{guan2023hallusionbench},  CRPE~\cite{wang2024allseeingv2}, and POPE~\cite{li2023pope}. The results are shown in Table~\ref{tab:exp-comprehensive-hallucination}. 
Compared with previous InternVL series, the new InternVL3.5 models provide consistent improvements in handling multimodal hallucination challenges across various model scales, \textit{e.g.,} +2.6 on 2B scale and +1.0 on 38B scale on the overall score. Despite these advancements, there are minor declines on some model scales such as 14B and 241B-A28B, indicating that further enhancement on data and training strategies is needed to achieve more consistent improvements on all model scales, which remains an important future direction to build a more trustworthy multimodal model.

\subsection{Visual Grounding}
\label{sec:exp-grounding}

For the visual grounding task, we evaluate InternVL3.5 on RefCOCO, RefCOCO+, and RefCOCOg datasets~\cite{kazemzadeh2014referitgame, mao2016generation}.  As shown in Table~\ref{tab:benchmark-grounding}, InternVL3.5 maintains the strong capabilities of InternVL3, which already achieves the upper bound of this tasks, \emph{i.e.,} approximately 90\% average accuracy. However, the InternVL3.5 training scheme still provides additional gains on several model sizes. For example, InternVL3.5-14B achieves an overall score of 90.1 on the RefCOCO series, outperforming InternVL3-14B by +0.8\%.  In addition, InternVL3.5-241B-A28B builds a new state-of-the-art performance on RefCOCO, with an overall score of 92.4, further highlighting the potential of InternVL3.5 for real-world applications requiring precise multimodal understanding.

\subsection{Multimodal Multilingual Understanding}
\label{sec:exp-multilingual}

InternVL3.5 exhibits strong multimodal multilingual understanding across a variety of benchmarks and languages. As summarized in Table~\ref{tab:exp-multilingual}, InternVL3.5 consistently achieves high scores on MMMB~\cite{sun2024parrot}, Multilingual MMBench~\cite{sun2024parrot} and MTVQA~\cite{tang2024mtvqa}, covering six languages including English, Chinese, Portuguese, Arabic, Turkish, and Russian. Compared to InternVL3, InternVL3.5 has  significant improvements in its language capabilities,  thus achieving better results on these multilingual benchmarks. For example, InternVL3.5-1B achieves up to 1.2\% gains  over InternVL3-1B. Compared to other leading multimodal models, such as Qwen2.5-VL and GPT-4V, InternVL3.5  also demonstrates notable improvements in both overall accuracy and language coverage, especially at larger scales.  For example, InternVL3.5-241B-A28B outperforms GPT-4V by +14.7\% on the overall score across all multilingual benchmarks.   The results highlight InternVL3.5’s robust capability to handle complex multilingual and multimodal tasks, making it highly effective for global applications that require comprehensive cross-language understanding.

\subsection{Video Understanding}
\label{sec:exp-video}

InternVL3.5 demonstrates remarkable video understanding capabilities across a comprehensive set of benchmarks. As presented in Table~\ref{tab:benchmark_video}, InternVL3.5 consistently achieves competitive or leading scores on Video-MME~\cite{fu2024video}, MVBench~\cite{li2024mvbench}, MMBench-Video~\cite{fang2024mmbench}, MLVU~\cite{MLVU}, LongVideoBench~\cite{wu2024longvideobench}. Performance improvements are observed across almost all metrics for small-size models.  In particular, InternVL3.5-1B outperforms InternVL3-1B by +2.2\% of overall performance.
For larger InternVL3.5 variants (such as 38B) also deliver comparable results with other state-of-the-art models of similar scale. Furthermore, InternVL3.5 exhibits robust generalization on challenging tasks involving long video sequences and complex reasoning, as reflected in its performance on LongVideoBench.  In particular, InternVL3.5-1B achieves significant improvements on LongVideoBench, \emph{i.e.,} +4.9\%.   The model’s ability to process multi-frame inputs and handle diverse video scenarios underscores its versatility. These results highlight InternVL3.5’s substantial progress in video understanding, positioning it as a highly capable solution for advanced multimodal video analysis tasks.

\begin{table*}[t]
\centering
\scriptsize
\setlength{\tabcolsep}{3.5pt}
\resizebox{\textwidth}{!}{
\begin{tabular}{l|ccc|cccc}
\mytoprule
\multirow{1}{*}{\textbf{Model}}
& \textbf{ScreenSpot} & \textbf{ScreenSpot-v2} & \textbf{OSWorld-G} 
&\textbf{WindowsAgentArena} & \textbf{WebArena-Lite-v2} \\
\mymidrule
ShowUI-2B~\cite{lin2024showui} & 75.1 & -- & -- &-- & -- \\
UI-TARS-2B~\cite{qin2025ui} & 82.3 & 84.7 & -- & -- & -- \\
JEDI-3B~\cite{xie2025scalingcomputerusegroundinguser} & -- & 80.9 & 27.3  & -- & -- \\
OS-Atlas-4B~\cite{wu2024atlas} & 70.1 & 71.9 & -- &  -- & -- \\
Qwen2.5-VL-3B~\cite{bai2025qwen2_5} & -- & 80.9 & 27.3 & -- & -- \\
\rowcolor{oursgray} InternVL3.5-4B & 83.6 & 85.1 & 33.9  & 9.7 & 7.8 \\
\mymidrule
OS-Atlas-7B~\cite{wu2024atlas} & 82.5 & 84.1 & 47.5 &  -- & -- \\
UGround-V1-7B~\cite{gou2025uground} & 86.3 & -- & 36.4 &  -- & -- \\
Aguvis-7B~\cite{xu2024aguvis} & 81.8 & -- & 38.7 & -- & -- \\
UI-TARS-7B~\cite{qin2025ui} & 89.5 & 91.6 & 47.5 &  -- & -- \\
UI-TARS-1.5-7B~\cite{qin2025ui} & -- & 89.7 & 64.2 &  15.9 & 17.5 \\
JEDI-7B~\cite{xie2025scalingcomputerusegroundinguser} & -- & 91.7 & 54.1  & -- & -- \\
Qwen2.5-VL-7B~\cite{bai2025qwen2_5} & -- & 88.8 & 31.4 & 3.4 & -- \\
GLM-4.1V-9B-Thinking\dag~\citep{hong2025glm_v_thinking}  & -- & -- & -- & -- & -- \\
MiMo-VL-7B-RL~\cite{coreteam2025mimovltechnicalreport} & 87.2 & 90.5 & 50.7  & -- & -- \\
InternVL3-8B~\cite{zhu2025internvl3} & 79.5 & 81.4 & --  & -- & -- \\
\rowcolor{oursgray} InternVL3.5-8B & 87.9 & 86.2 & 36.4 & 10.5  & 12.3 \\
\rowcolor{oursgray} InternVL3.5-14B & 87.5 & 88.6 & 44.7& 12.5 & 12.3 \\

\mymidrule
GTA1-32B~\cite{yang2025gta1} & -- & 93.2 & 61.9 &  -- & -- \\
Qwen2.5-VL-32B~\cite{bai2025qwen2_5} & -- & 91.3 & 46.5 &  -- & -- \\
Gemma-3-27B-IT\dag~\cite{team2025gemma}   & -- & -- & -- &  -- & -- \\
InternVL3-38B~\cite{zhu2025internvl3} & 85.6 & 88.3 & -- &  -- & -- \\
\rowcolor{oursgray} InternVL3.5-20B-A4B & 85.5 & 87.6 & 38.2 & 11.0 & 9.5 \\
\rowcolor{oursgray} InternVL3.5-30B-A3B & 86.6 & 87.3 & 42.4 & 11.0 & 10.4 \\
\rowcolor{oursgray} InternVL3.5-38B & 81.0 & 83.5 & 42.9 & 14.5 & 7.1 \\
\mymidrule
Operator\ddag~\cite{cua2025} & -- & 70.5 & 40.6 & -- & -- \\
Aguvis-72B~\cite{xu2024aguvis} & 89.2 & -- & --  & 3.5 & 9.0 \\
UI-TARS-72B~\cite{qin2025ui} & 88.4 & 90.3 & 57.1  & 17.9 & 10.3 \\
GPT-4o~\cite{chatgpt4o} & 18.1 & -- & --  & 3.5 & 1.9 \\
Claude-3.5-Sonnet~\cite{claude3series2024} & 83.0 & -- & -- & -- & -- \\
Claude-3.7-Sonnet~\cite{anthropic2025claude37} & -- & 87.6 & -- & 6.4 & 1.9 \\
Gemini-2.0-Flash~\cite{gemini2_0} & 84.0 & -- & -- &  -- & -- \\
Gemini-2.5-Pro~\cite{geminipro2.5} & -- & -- & 45.2 &  -- & -- \\
Seed1.5-VL~\cite{guo2025seedvl1_5} & -- & 95.2 & 62.9  & -- & -- \\
Qwen2.5-VL-72B~\cite{bai2025qwen2_5} & 87.1 & -- & --  & 9.7 & 14.4 \\
InternVL3-78B~\cite{zhu2025internvl3} & 88.7 & 90.9 & -- &  -- & -- \\
\rowcolor{oursgray} InternVL3.5-241B-A28B & 89.8 & 92.9 & 53.2 &  18.0 & 11.7 \\
\mybottomrule
\end{tabular}
}
\caption{\textbf{Comparison of GUI grounding and online agentic evaluation results.} 
To assess the GUI agent capabilities of InternVL3.5, we conducted evaluations on a diverse set of platforms. We evaluate GUI grounding capabilities across 3 benchmarks including ScreenSpot~\cite{cheng2024seeclick}, ScreenSpot-v2~\cite{wu2024atlas} and OSWorld-G~\cite{xie2025scalingcomputerusegroundinguser}. For online agentic evaluation, our assessment covers Ubuntu, Windows, and Web utilizing the OSWorld~\cite{xie2024osworld}, WindowsAgentArena~\cite{bonatti2024windows}, and WebArena-Lite-v2~\cite{wang2025mmbenchgui}. Models with symbols~\dag~and~\ddag~ are evaluated under 100 and 200 steps, respectively, while all other results were evaluated under 50 steps.}
\label{tab:exp_gui}
\end{table*}

\subsection{GUI Agent Tasks} 
\label{sec:exp-gui}
To validate the GUI agent  capabilities of InternVL3.5, we conduct detailed experiments in Table~\ref{tab:exp_gui}.  In particular, we evaluate  InternVL3.5 on six GUI grounding and agent tasks, namely ScreenSpot~\cite{cheng2024seeclick}, ScreenSpot-v2~\cite{wu2024atlas}, OSWorld-G~\cite{xie2024osworld},  OSWorld~\cite{xie2024osworld}, WindowsAgentArena~\cite{bonatti2024windows}, and  WebArena-Lite-v2~\cite{wang2025mmbenchgui}. In   GUI grounding tasks, InternVL3.5 outperforms most open-source models and is close to the performance of closed-source models. For example, InternVL3.5-241B-A28B outperforms the specialized model, \emph{e.g.,} +2.6\% against UI-TARS-72B~\cite{qin2025ui} on ScreenSpot-v2. Compared to the most advanced commercial model, \emph{i.e.,} Seed1.5-VL~\cite{guo2025seedvl1_5}, InternVL3.5-241B-A28B still maintains close performance on ScreenSpot-v2, \emph{i.e.,} 92.9 \textit{vs.} 95.2.  In GUI agent tasks, InternVL also demonstrates competitive results on different platforms.   In particular, InternVL3.5-241B-A28B achieves the best results against existing  generalist MLLMs, \emph{e.g.,} +8.3\% over Qwen2.5-VL-72B on WindowsAgentArena.  In particular, GPT-4o only achieves a score of 3.5 on this challenging benchmark.  On WebArena-Lite-v2, the performance score of GPT-4o further decreases to 1.9, but InternVL3.5-241B-A28B can still achieve a top-tier score of 11.7.  These results confirm the great potential of InternVL3.5 as a fundamental model for GUI tasks.

\begin{table}[t]
\centering
\small
\renewcommand{\arraystretch}{1.1}
\begin{tabular}{lccccc}
\mytoprule
\multicolumn{1}{l}{\textbf{Method}}      & \textbf{VSI-Bench} & \textbf{ERQA} & \textbf{SpaCE-10} & \textbf{OmniSpatial} & \textbf{Overall} \\ \mymidrule
InternVL3-1B~\cite{zhu2025internvl3}                           & 29.7         & 30.3          & 41.3                            & 35.5                 & 34.2             \\
\rowcolor{oursgray}InternVL3.5-1B                          & 49.3         & 35.3          & 33.6                            & 40.7                 & 39.7             \\
\mymidrule

Qwen2.5-VL-3B~\cite{bai2025qwen2_5}     & 27.9         & 38.0          & 34.8                            & 40.3                 & 35.3             \\
InternVL3-2B~\cite{zhu2025internvl3}                            & 31.5         & 31.5          & 44.2                            & 38.0                 & 36.3             \\
\rowcolor{oursgray}InternVL3.5-2B                          & 53.8         & 37.3          & 34.6                            & 42.3                 & 42.0             \\
\mymidrule

MiniCPM-V-4-4B~\cite{yao2024minicpm}                             & 30.3         & 36.3          & 39.0                            & 43.1                 & 37.2             \\
\rowcolor{oursgray}InternVL3.5-4B                          & 54.9         & 38.5          & 35.5                            & 45.8                 & 43.7             \\ \mymidrule

Qwen2.5-VL-7B~\cite{bai2025qwen2_5}     & 35.9         & 38.8          & 33.3                            & 39.2                 & 36.8             \\
MiMo-VL-RL-8B~\cite{coreteam2025mimovltechnicalreport}                              & 36.4         & 37.8          & 36.1                            & 46.5                 & 39.2             \\
Keye-VL-8B~\cite{team2025keye_vl}                              & 28.6         & 35.3          & 38.6                            & 46.5                 & 37.2             \\
GLM-4.1V-9B~\cite{hong2025glm_v_thinking}                             & 39.2      & 45.8          & 43.4                            & 47.7                 & 44.0             \\
InternVL3-8B~\cite{zhu2025internvl3}                            & 42.1         & 35.3          & 40.0                            & 41.6                 & 39.7             \\ 
\rowcolor{oursgray}InternVL3.5-8B                          & 56.3         & 41.0          & 39.5                            & 47.8                 & 46.1             \\
\mymidrule

Gemma-3-12B~\cite{team2025gemma}                              & 21.9         & 36.1          & 41.5                            & 43.7                 & 35.8             \\
InternVL3-14B~\cite{zhu2025internvl3}                           & 48.9         & 39.5          & 47.3                            & 45.9                 & 45.4             \\
\rowcolor{oursgray}InternVL3.5-14B                         & 60.8         & 41.8          & 48.8                            & 47.6                 & 49.7             \\ 
\mymidrule

Kimi-VL-A3B-2506~\cite{team2025kimi}                                 & 37.4         & 36.0          & 39.2                            & 37.3                 & 37.5             \\
\rowcolor{oursgray}InternVL3.5-20B-A4B                  &  60.1         & 41.6          & 51.6                            & 45.4                 & 46.7             \\
\rowcolor{oursgray}InternVL3.5-30B-A3B                     & 63.7         & 41.5          & 49.7                            & 48.1                 & 50.8             \\
\mymidrule

Gemma-3-27B~\cite{team2025gemma}                              & 22.0         & 37.5          & 41.5                            & 44.8                 & 36.4             \\
Qwen2.5-VL-32B~\cite{bai2025qwen2_5}    & 34.7         & 40.7          & 32.6                            & 47.4                 & 38.8             \\
InternVL3-38B~\cite{zhu2025internvl3}                           & 48.9         & 42.8          & 53.1                         & 48.5                 & 48.3             \\
\rowcolor{oursgray}InternVL3.5-38B                         & 66.3         & 43.3          & 43.8                            & 51.4                 & 51.2             \\ 

\mymidrule
GPT-4o-20241120~\cite{chatgpt4o}            & 34.0         & 47.0          & 49.0                            & 47.8                 & 44.5             \\
GPT-5-20250807~\cite{gpt5}                                  & 37.5         &     65.7          & 43.8                            & 59.6              & 51.7              \\
Gemini-2.5-Pro~\cite{gemini2_0} & 47.8         & 48.3          & 52.7                            & 55.2                 & 51.0               \\
Claude-3.7-Sonnet~\cite{claude3series2024}             & 47.0         & 35.5          & 46.2                            & 46.9                 & 43.9             \\
GLM-4.5V~\cite{hong2025glm_v_thinking}                                & 41.4         & 46.5          & 51.6                            & 51.0                 & 47.6             \\
Qwen2.5-VL-72B~\cite{bai2025qwen2_5}    & 36.1         & 44.8          & 37.9                            & 47.9                 & 41.7             \\
Step3-321B-A38B~\cite{wang2025step}                                   & 34.2      & 44.5          & 42.6                            & 47.0                 & 42.1              \\
InternVL3-78B~\cite{zhu2025internvl3}                           & 48.4         & 45.9          & 52.5                         & 49.3                 & 49.0              \\
\rowcolor{oursgray} InternVL3.5-241B-A28B                   & 69.5         & 46.8          &                       55.0          &  51.9                    & 55.8              \\ \mybottomrule
\end{tabular}
\vspace{3mm}
\caption{\textbf{Comparison of embodied task performance.} We compare InternVL3.5 and existing methods on VSI-Bench~\cite{yang2024think}, ERQA~\cite{erqa}, SpaCE-10~\cite{gong2025space10}, and OmniSpatial~\cite{jia2025omnispatial}. For SpaCE-10~\cite{gong2025space10}, we report the single-choice performance.}
\label{tab:exp-emboidied}
\end{table}

\begin{table*}[t]
\centering
\small
\renewcommand{\arraystretch}{1.2}
\begin{tabular}{l|ccccc|c}
\mytoprule
\textbf{Model} & \textbf{Semantics} $\uparrow$ & \textbf{Count} $\uparrow$ & \textbf{Color} $\uparrow$ & \textbf{Shape} $\uparrow$ & \textbf{Reasoning} $\uparrow$ & \textbf{Overall} $\uparrow$ \\
\mymidrule
Gemma-1.1-2B~\cite{team2024gemma} & 32.1 & 33.3 & 25.0 & 35.6 & 28.7 & 31.7 \\
\rowcolor{oursgray}
InternVL3.5-1B & 25.5 & 22.7 & 24.9 & 24.4 & 30.4 & 25.0 \\
\rowcolor{oursgray}
InternVL3.5-2B & 25.2 & 20.1 & 45.4 & 33.4 & 26.5 & 30.7 \\
\rowcolor{oursgray}
InternVL3.5-4B & 41.2 & 55.2 & 81.9 & 62.4 & 39.4 & 57.7 \\
\mymidrule

InternLM2.5-7B~\cite{cai2024internlm2} & 27.3 & 31.7 & 59.8 & 51.5 & 28.2 & 42.1 \\
Keye-VL-8B~\cite{team2025keye_vl} & 41.4 & 47.5 & 71.4 & 54.9 & 40.6 & 52.2 \\
GLM-4.1V-9B~\cite{hong2025glm_v_thinking} & 41.6 & 55.6 & 79.1 & 61.5 & 40.0 & 57.1 \\
InternVL3-8B~\cite{zhu2025internvl3} & 33.7 & 46.5 & 69.8 & 59.1 & 36.1 & 50.6 \\
\rowcolor{oursgray}
InternVL3.5-8B & 39.7 & 54.0 & 82.3 & 53.4 & 41.7 & 54.9 \\
\mymidrule

Gemma-3-12B~\cite{team2025gemma} & 24.8 & 30.8 & 47.2 & 25.7 & 22.8 & 30.5 \\
DeepSeek-Coder-V2-16B~\cite{zhu2024deepseek_coder_v2} & 30.9 & 37.9 & 63.7 & 54.8 & 26.8 & 45.1 \\
InternVL3-14B~\cite{zhu2025internvl3} & 38.2 & 52.9 & 74.4 & 54.1 & 41.7 & 52.9 \\
\rowcolor{oursgray}
InternVL3.5-14B & 44.3 & 55.3 & 77.8 & 63.7 & 45.1 & 58.5 \\
\mymidrule

Kimi-VL-A3B-2506~\cite{team2025kimi} & 31.1 & 41.5 & 67.0 & 47.4 & 32.4 & 44.9 \\
\rowcolor{oursgray}
InternVL3.5-20B-A4B & 51.2 & 60.6 & 89.8 & 74.7 & 55.5 & 67.6 \\
\rowcolor{oursgray}
InternVL3.5-30B-A3B & 51.8 & 66.8 & 91.8 & 75.7 & 53.8 & 69.4 \\
\mymidrule

Gemma-3-27B~\cite{team2025gemma} & 36.7 & 51.4 & 76.3 & 62.1 & 39.4 & 54.7 \\
Qwen2.5-VL-32B~\cite{bai2025qwen2_5} & 40.0 & 55.7 & 76.3 & 61.2 & 43.9 & 56.5 \\
InternVL3-38B~\cite{zhu2025internvl3} & 40.8 & 58.7 & 82.2 & 63.6 & 43.9 & 59.1 \\
\rowcolor{oursgray}
InternVL3.5-38B & 47.6 & 66.5 & 90.5 & 80.4 & 54.6 & 69.5 \\
\mymidrule

GPT-5~\cite{gpt5} & 67.8 & 72.6 & 91.7 & 81.9 & 68.7 & 77.5 \\
GLM-4.5V~\cite{hong2025glm_v_thinking} & 47.3 & 63.7 & 87.3 & 72.3 & 55.8 & 66.1 \\
Qwen2.5-VL-72B~\cite{bai2025qwen2_5} & 40.2 & 55.1 & 80.1 & 62.0 & 41.1 & 57.1 \\
Step3-321B-A38B~\cite{wang2025step} & 35.9 & 54.0 & 82.8 & 63.2 & 38.6 & 56.5 \\
InternVL3-78B~\cite{zhu2025internvl3} & 41.0 & 59.1 & 84.0 & 65.2 & 47.0 & 60.3 \\
\rowcolor{oursgray}
InternVL3.5-241B-A28B & 51.2 & 69.2 & 92.1 & 77.6 & 58.0 & 70.7 \\
\mybottomrule
\end{tabular}
\caption{
    \textbf{Comparison of SVG understanding performance on SGP-Bench~\cite{qiu2024can}.}
}
\label{tab:exp-sgp-bench}
\end{table*}

\begin{table*}[t]
\centering
\small
\renewcommand{\arraystretch}{1.2}
\begin{tabular}{l|rrr|cccc}
\mytoprule
\multirow{2}{*}{\textbf{Model}} & \multicolumn{3}{c|}{\textbf{Text2SVG}} & \multicolumn{4}{c}{\textbf{Img2SVG}} \\
& \textbf{FID} $\downarrow$ & \textbf{FID-C} $\downarrow$ & \textbf{CLIP} $\uparrow$
& \textbf{DINO} $\uparrow$ & \textbf{SSIM} $\uparrow$ & \textbf{LPIPS} $\downarrow$ & \textbf{PSNR} $\uparrow$ \\
\mymidrule

\rowcolor{oursgray}
InternVL3.5-1B & 22.50 & 12.16 & 72.43 & 0.79 & 0.57 & 0.35 & 7.35 \\
\rowcolor{oursgray}
InternVL3.5-2B & 20.98 & 11.26 & 72.71 & 0.81 & 0.56 & 0.34 & 7.44 \\
\rowcolor{oursgray}
InternVL3.5-4B & 17.06 & 7.54 & 74.35 & 0.84 & 0.61 & 0.30 & 8.37 \\
\mymidrule

Llama-3.1-8B~\cite{dubey2024llama3}           & 19.43 & 11.25 & 71.86 & -- & -- & -- & -- \\
Qwen2.5-VL-7B~\cite{bai2025qwen2_5}          & 24.78 & 15.45 & 71.38 & 0.78 & 0.51 & 0.38 & 6.53 \\
Keye-VL-8B~\cite{team2025keye_vl}     & 21.96 & 14.39 & 71.17 & 0.80 & 0.53 & 0.37 & 6.94 \\
GLM-4.1V-9B~\cite{hong2025glm_v_thinking}        & 22.68 & 10.45 & 73.20 & 0.82 & 0.54 & 0.35 & 7.33 \\
InternVL3-8B~\cite{zhu2025internvl3}      & 23.06 & 14.30 & 71.45 & 0.81 & 0.56 & 0.36 & 7.22 \\
\rowcolor{oursgray}
InternVL3.5-8B & 17.36 & 7.13 & 75.01 & 0.85 & 0.62 & 0.29 & 8.74 \\
\mymidrule

Llama-3.2-11B-Vision~\cite{dubey2024llama3}   & 28.16 & 14.35 & 71.49 & 0.76 & 0.47 & 0.39 & 5.91 \\
Gemma-3-12B~\cite{team2025gemma}              & 17.14 & 10.41 & 71.62 & 0.82 & 0.58 & 0.35 & 7.63 \\
InternVL3-14B~\cite{zhu2025internvl3}     & 19.00 & 13.22 & 71.49 & 0.83 & 0.56 & 0.36 & 7.34 \\
\rowcolor{oursgray}
InternVL3.5-14B & 15.90 & 5.99 & 75.91 & 0.86 & 0.61 & 0.31 & 8.46 \\
\mymidrule

Kimi-VL-A3B~\cite{team2025kimi} & 30.81 & 16.99 & 70.54 & 0.80 & 0.56 & 0.36 & 7.18 \\
\rowcolor{oursgray}
InternVL3.5-20B-A4B & 16.78 & 5.60 & 77.46 & 0.91 & 0.71 & 0.20 & 12.75 \\
\rowcolor{oursgray}
InternVL3.5-30B-A3B & 16.31 & 5.84 & 76.40 & 0.88 & 0.65 & 0.27 & 9.64 \\
\mymidrule

Gemma-3-27B~\cite{team2025gemma}              & 15.15 & 9.30 & 73.28 & 0.83 & 0.60 & 0.35 & 7.83 \\
Qwen2.5-VL-32B~\cite{bai2025qwen2_5}          & 20.04 & 10.39 & 73.23 & 0.84 & 0.56 & 0.36 & 7.50 \\
InternVL3-38B~\cite{zhu2025internvl3}     & 18.01 & 11.04 & 73.08 & 0.83 & 0.55 & 0.35 & 7.31 \\
\rowcolor{oursgray}
InternVL3.5-38B & 14.56 & 5.22 & 76.49 & 0.86 & 0.61 & 0.32 & 8.39 \\
\mymidrule

Grok-3~\cite{grok3}                            & 21.97 & 8.69 & 76.80 & -- & -- & -- & -- \\
Llama-3.1-70B~\cite{dubey2024llama3}          & 18.03 & 8.30 & 73.88 & -- & -- & -- & -- \\
Llama-3.1-405B~\cite{dubey2024llama3}         & 16.79 & 8.39 & 73.92 & -- & -- & -- & -- \\
DeepSeek-V3-671B-A37B~\cite{liu2024deepseekv3}          & 24.99 & 8.80 & 76.47 & -- & -- & -- & -- \\

GPT-4o~\cite{chatgpt4o}                       & 15.18 & 6.76 & 77.74 & 0.87 & 0.62 & 0.32 & 8.44 \\
GLM-4.5V~\cite{hong2025glm_v_thinking}                   & 16.64 & 5.09 & 78.35 & 0.87 & 0.63 & 0.32 & 8.67 \\
Claude-3.7-Sonnet~\cite{claude3series2024}    & 14.38 & 3.50 & 80.79 & 0.91 & 0.65 & 0.29 & 9.26 \\
Claude-4-Sonnet~\cite{anthropic2025claude4}   & 15.84 & 4.29 & 80.58 & 0.92 & 0.67 & 0.28 & 9.86 \\
Gemini-2.5-Flash~\cite{gemini2_5}             & 16.72 & 5.21 & 78.22 & 0.88 & 0.59 & 0.32 & 8.32 \\
Llama-3.2-90B-Vision~\cite{dubey2024llama3}   & 19.31 & 8.55 & 74.00 & 0.76 & 0.44 & 0.38 & 5.78 \\
Llama-4-Scout~\cite{meta2025llama4scout}     & 17.91 & 9.38 & 73.56 & 0.84 & 0.58 & 0.35 & 7.74 \\
Llama-4-Maverick~\cite{meta2025llama4maverick}& 14.93 & 6.53 & 75.82 & 0.86 & 0.60 & 0.33 & 8.03 \\

Step3-321B-A38B~\cite{wang2025step}           & 20.06 & 9.71 & 74.18 & 0.83 & 0.56 & 0.34 & 7.52 \\
Qwen2.5-VL-72B~\cite{bai2025qwen2_5}          & 15.95 & 9.88 & 73.68 & 0.84 & 0.58 & 0.35 & 7.83 \\
InternVL3-78B~\cite{zhu2025internvl3}     & 17.58 & 10.60 & 73.12 & 0.85 & 0.58 & 0.34 & 7.80 \\
\rowcolor{oursgray}
InternVL3.5-241B-A28B & 11.27 & 4.43 & 76.81 & 0.88 & 0.64 & 0.29 & 9.19 \\
\mybottomrule
\end{tabular}
\caption{\textbf{Comparison of SVG generation performance on SArena-Icon (Text2SVG and Img2SVG).} 
}
\label{tab:exp-sarena-grouped-40plus}
\end{table*}

\begin{table*}[t]
\centering
\scriptsize
\renewcommand{\arraystretch}{1.2}
\setlength\tabcolsep{4pt}

\newcommand{\QwenOne}{\rotatebox{90}{\makecell{Qwen3-0.6B}}}
\newcommand{\QwenTwo}{\rotatebox{90}{\makecell{Qwen3-1.7B}}}
\newcommand{\QwenThree}{\rotatebox{90}{\makecell{Qwen3-4B}}}
\newcommand{\QwenFour}{\rotatebox{90}{\makecell{Qwen3-8B}}}
\newcommand{\QwenFive}{\rotatebox{90}{\makecell{Qwen3-14B}}}
\newcommand{\QwenSix}{\rotatebox{90}{\makecell{Qwen3-30B-A3B}}}
\newcommand{\QwenSeven}{\rotatebox{90}{\makecell{Qwen3-32B}}}
\newcommand{\QwenEight}{\rotatebox{90}{\makecell{Qwen3-235B-A22B}}}

\newcommand{\InternVLOne}{\rotatebox{90}{\makecell{InternVL3.5-1B}}}
\newcommand{\InternVLTwo}{\rotatebox{90}{\makecell{InternVL3.5-2B}}}
\newcommand{\InternVLThree}{\rotatebox{90}{\makecell{InternVL3.5-4B}}}
\newcommand{\InternVLFour}{\rotatebox{90}{\makecell{InternVL3.5-8B}}}
\newcommand{\InternVLFive}{\rotatebox{90}{\makecell{InternVL3.5-14B}}}
\newcommand{\InternVLSix}{\rotatebox{90}{\makecell{InternVL3.5-30B-A3B}}}
\newcommand{\InternVLSeven}{\rotatebox{90}{\makecell{InternVL3.5-38B}}}
\newcommand{\InternVLEight}{\rotatebox{90}{\makecell{InternVL3.5-241B-A28B}}}

\begin{tabular}{lc|rr|rr|rr|rr|rr|rr|rr|rr}
Dataset                         &  Version                                       & \QwenOne & \InternVLOne & \QwenTwo & \InternVLTwo & \QwenThree & \InternVLThree & \QwenFour & \InternVLFour & \QwenFive & \InternVLFive & \QwenSix & \InternVLSix & \QwenSeven & \InternVLSeven & \QwenEight & \InternVLEight \\
\mymidrule
MMLU                 & 4d595a & 52.8 & 49.1 & 62.6 & 61.9 & 73.0 & 73.8 & 76.9 & 78.2 & 81.1 & 81.5 & 81.4 & 83.0 & 83.6 & 84.6 & 87.8 & 89.1 \\
CMMLU                & c13365 & 43.4 & 46.6 & 59.8 & 59.4 & 71.8 & 70.6 & 76.6 & 76.2 & 81.6 & 79.8 & 83.0 & 82.5 & 84.5 & 84.4 & 87.4 & 90.2 \\
C-Eval               & 2daf24 & 42.6 & 49.0 & 61.0 & 61.0 & 72.2 & 71.9 & 77.9 & 77.9 & 81.0 & 80.3 & 82.9 & 83.2 & 83.3 & 85.0 & 86.1 & 90.9 \\
GAOKAO               & 4c31db & 40.4 & 48.4 & 64.1 & 68.1 & 80.0 & 83.2 & 85.6 & 84.1 & 90.0 & 87.2 & 89.5 & 92.6 & 93.2 & 93.4 & 95.0 & 94.5 \\
\mymidrule
TriviaQA             & 2121ce & 18.7 & 20.4 & 30.6 & 31.7 & 39.9 & 40.2 & 52.0 & 49.7 & 60.9 & 55.3 & 58.4 & 57.6 & 63.4 & 59.9 & 73.7 & 74.8 \\
NaturalQuestions     & 3dcea1 & 11.2 & 15.0 & 21.4 & 23.5 & 29.3 & 29.0 & 36.5 & 32.6 & 42.2 & 36.7 & 42.5 & 36.7 & 45.8 & 40.1 & 54.8 & 50.4 \\
C3                   & 8c358f & 64.1 & 65.5 & 54.2 & 79.5 & 79.8 & 89.5 & 83.0 & 91.6 & 84.3 & 94.6 & 90.8 & 95.1 & 93.5 & 95.6 & 96.4 & 97.8 \\
RACE-High            & 69ee4f & 45.3 & 67.2 & 65.6 & 78.8 & 82.7 & 86.4 & 86.2 & 88.4 & 87.2 & 90.7 & 78.7 & 92.3 & 85.6 & 92.1 & 90.5 & 94.2 \\
\mymidrule
WinoGrande           & b36770 & 51.5 & 54.9 & 53.5 & 59.3 & 64.8 & 69.1 & 70.8 & 75.2 & 76.6 & 80.5 & 73.0 & 86.5 & 74.8 & 83.4 & 84.8 & 91.7 \\
HellaSwag            & e42710 & 42.7 & 49.5 & 59.0 & 74.4 & 78.1 & 88.5 & 84.6 & 91.5 & 88.2 & 94.1 & 89.0 & 96.3 & 90.5 & 95.9 & 91.1 & 97.0 \\
BBH                  & 5b92b0 & 41.5 & 46.8 & 54.5 & 62.3 & 72.6 & 78.6 & 78.4 & 79.8 & 81.1 & 82.3 & 81.5 & 83.8 & 87.4 & 87.5 & 88.9 & 86.5 \\
\mymidrule
GSM8K                & 1d7fe4 & 59.6 & 62.8 & 75.4 & 77.2 & 87.8 & 92.0 & 89.8 & 90.9 & 92.5 & 95.6 & 91.8 & 91.4 & 93.4 & 91.5 & 94.4 & 91.6 \\
MATH                 & 393424 & 32.4 & 68.2 & 43.5 & 85.5 & 54.1 & 94.4 & 60.8 & 93.3 & 62.0 & 93.6 & 59.0 & 93.6 & 61.6 & 94.3 & 71.8 & 94.5 \\
AIME2024             & --     & 10.0 & 13.7 & 40.0 & 44.2 & 66.7 & 72.8 & 76.0 & 77.7 & 80.0 & 77.4 & 83.3 & 79.4 & 73.3 & 81.5 & 86.7 & 84.1 \\
AIME2025             & --     & 13.3 & 14.7 & 23.3 & 42.9 & 50.0 & 57.6 & 67.3 & 64.0 & 66.7 & 63.9 & 70.0 & 62.7 & 60.0 & 72.1 &  83.3    & 75.6 \\
HumanEval        & 8e312c & 39.6 & 45.7 & 72.0 & 65.9 & 82.3 & 87.8 & 85.4 & 93.9 & 89.0 & 97.0 & 89.0 & 96.3 & 89.6 & 98.2 &  91.5    & 98.2 \\
\mymidrule
Overall          & -- & 38.1 & 44.8 & 52.5 & 61.0 & 67.8 & 74.1 & 74.6 & 77.8 & 77.8 & 80.7 & 77.8 & 82.0 & 79.0 & 83.7 &  85.3    & 87.6    \\        
\end{tabular}
\caption{\textbf{Comparison of text-related performance across multiple benchmarks.} Results were obtained using the OpenCompass toolkit~\cite{opencompass2023}. We compare InternVL3.5 with Qwen3 models, whose corresponding pre-trained base models are employed as the initialization of the language component in InternVL3.5.  
Please note that the evaluation scores of the Qwen3 series may differ from those officially reported, as we have adopted the prompt versions provided in the table across all datasets for OpenCompass evaluation.
}
\label{tab:language_model_comparison}
\end{table*}

\subsection{Embodied Agent Tasks} 
\label{sec:exp-embody}
In Table \ref{tab:exp-emboidied}, we evaluate the embodied capabilities of InternVL3.5 on four benchmarks: VSI-Bench~\cite{yang2024think}, ERQA~\cite{erqa}, Space10~\cite{gong2025space10}, and OmniSpatial~\cite{jia2025omnispatial}.   From this table, we observe the strong embodied capabilities of InternVL3.5 against previous works. On VSI-Bench, the most popular benchmark for spatial reasoning, InternVL3.5-1B achieves an overall score of 49.3, outperforming its predecessor  by +19.6\%.  We note that InternVL3.5-1B can already achieve the state-of-the-art performance on VSI-Bench against much larger models like Qwen2.5-VL-72B~\cite{bai2025qwen2_5}. When the model size of InternVL3.5 increases, the performance on VSI-Bench consistently improves from  49.3 to 69.5, greatly validating the salability of InternVL3.5 on embodied tasks. In addition to VSI-Bench, InternVL3.5 also demonstrates promising results on ERQA, Space10 and OminiSpatial.  Among them, InternVL3.5B-241B-A28B scores 46.8 on ERQA, which is close to the 48.3 score of the top closed-source model Gemini-2.5-Pro.  
In terms of overall performance, InternVL3.5B-241B-A28B also reaches the highest score against other models, confirming its strong capabilities in embodied tasks.

\subsection{SVG Tasks}
\label{sec:exp-svg}
Scalable vector graphics (SVG) is a common format for describing graphics on web pages, and its understanding is significant for web-based agent tasks. 
To evaluate this capability,  We evaluate InternVL3.5 on SGP-Bench~\cite{qiu2024can} (Table~\ref{tab:exp-sgp-bench}), where it delivers strong results across model scales and sets new open-source state-of-the-art at larger capacities. At the small scale, InternVL3.5-4B already surpasses models such as Kimi-VL-A3B~\cite{team2025kimi} and even the earlier InternVL3-14B~\cite{zhu2025internvl3}. In the mid-size range (8B, 14B), InternVL3.5 shows broad improvements, and the 14B variant surpasses larger counterparts such as Gemma-3-27B~\cite{team2025gemma}. At the high end, InternVL3.5-30B-A3B and InternVL3.5-38B achieve nearly 70\% accuracy on SGP-Bench, advancing the state-of-the-art among open models and outperforming competitors such as Step3-321B-A38B~\cite{wang2025step}, Qwen2.5-VL-72B~\cite{bai2025qwen2_5} and GLM-4.5V~\cite{hong2025glm_v_thinking}.
Finally, InternVL3.5-241B-A28B sets a new record for open-source models, achieving the best performance across all categories except when compared to GPT-5~\cite{gpt5}.

In the SArena-Icon generation tasks (Text2SVG and Img2SVG), InternVL3.5 establishes new state-of-the-art performance among open models (Table~\ref{tab:exp-sarena-grouped-40plus}).  
In Text2SVG, our models achieve markedly lower FID and FID-C scores than previous baselines, with the 38B variant reducing FID to 14.56. This performance even surpasses GPT-4o~\cite{chatgpt4o} (15.18), highlighting that InternVL3.5 is able to match or exceed the capabilities of much larger proprietary systems.
In Img2SVG, the 30B and 38B variants deliver leading results on structural similarity metrics, outperforming open-source peers of comparable scale and closely matching the best proprietary systems.
Furthermore, InternVL3.5-241B-A28B achieves an even stronger balance of fidelity and perceptual quality, with an FID of 11.27 and an FID-C of 4.43 in Text2SVG, together with competitive Img2SVG scores. These results highlight both the efficiency and scalability of InternVL3.5, showing that it consistently outperforms prior open models across all metrics, narrows the gap with proprietary leaders, and establishes itself as the most powerful open-source framework for SVG generation to date.

\subsection{Evaluation on Language Capability}
\label{sec:exp-language}

To evaluate the language capabilities of InternVL3.5,  we use benchmarks covering comprehensive assessments in general knowledge (MMLU~\cite{hendrycks2020measuring}, CMMLU~\cite{li2023cmmlu}, C-Eval~\cite{huang2023ceval}, GAOKAO-Bench~\cite{Zhang2023gaokao}), linguistic understanding (TriviaQA~\cite{joshi2017triviaqa}, NaturalQuestions~\cite{naturalquestion}, C3~\cite{sun2019investigating}, RACE~\cite{lai2017race}), reasoning (WinoGrande~\cite{sakaguchi2020winogrande}, HellaSwag~\cite{zellers2019hellaswag}, BigBench Hard~\cite{suzgun2023bigbench}), mathematics (GSM8K-Test~\cite{cobbe2021training}, MATH~\cite{DBLP:conf/nips/HendrycksBKABTS21}, AIME24~\cite{aime2024}, AIME25~\cite{aime2025}), and coding (HumanEval~\cite{chen2021evaluating}) tasks.   As shown in Table \ref{tab:language_model_comparison},  InternVL3.5 achieves even better performance than its corresponding language models  on most benchmarks.  Specifically, InternVL3.5-1B outperforms Qwen3-0.6B on 15 of 16 text-related benchmarks, with an overall performance gain of +6.7.  For larger models such as InternVL3.5-241B-A28B, the performance improvement is also obvious, \emph{i.e.,} +2.3 over Qwen3-235B-A22B.  These improvements come not only from the high-quality text corpora we use during pre-training and SFT, but also from our Cascade RL, which significantly benefits  text-based reasoning tasks. The improvement of InternVL3.5 in text capabilities has also greatly compensated for the shortcomings of open-source multimodal models in general capabilities.

\subsection{Ablation Study}
\label{sec:exp-ablation}

\begin{figure}[t] 
  \centering 
  \includegraphics[width=1.0\linewidth]{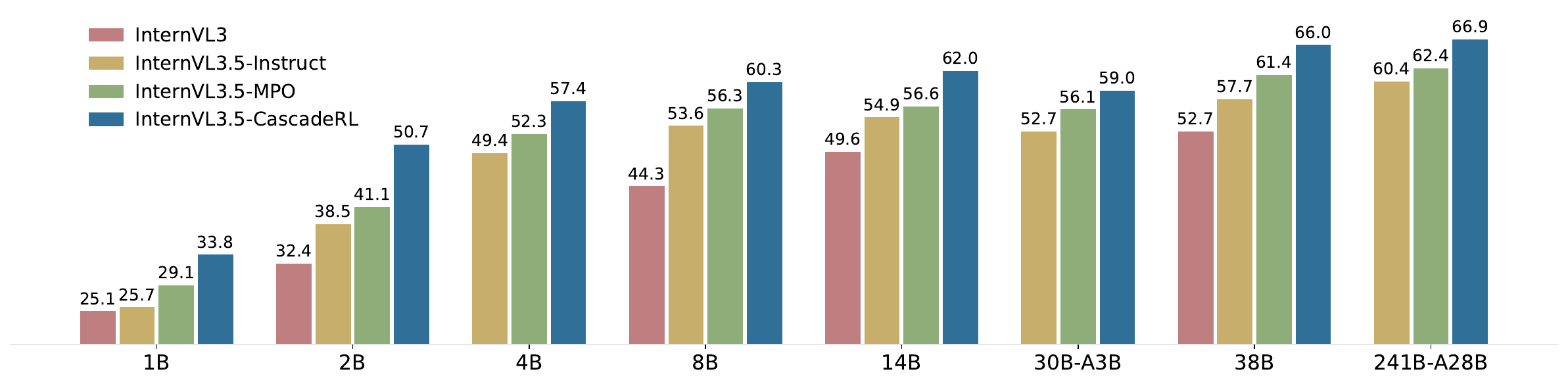}
  \caption{
    \textbf{Ablation on Cascade RL.} We report average scores on the same set of multimodal reasoning and mathematical benchmarks as in Table~\ref{tab:exp-reasoning}. Full results are provided in Table~\ref{tab:exp-ablation-cascade-rl}.
  }
  \label{fig:ablation_training}
\end{figure}

\begin{table}[t]
\centering
\scriptsize
\setlength{\tabcolsep}{3.5pt}
\renewcommand\arraystretch{1.2}
    \begin{tabular}{l|ccccccc|c}
    \mytoprule
    \textbf{Model} & \makecell{\textbf{MMMU}\\\textbf{(val)}} & \makecell{\textbf{MathVista}\\\textbf{(mini)}} & \makecell{\textbf{MathVision}} & \makecell{\textbf{MathVerse}\\\textbf{(vision-only)}} & \makecell{\textbf{DynaMath}\\\textbf{(worst case)}} & \textbf{WeMath} & \textbf{LogicVista} & \textbf{Overall} \\
    \mymidrule
    InternVL3-1B               & 43.4                                                        & 45.8                                                              & 18.8                                                               & 18.7                                                                     & 5.8                                                                 & 13.4                 & 29.8                 & 25.1                 \\
    InternVL3.5-1B-Instruct    & 37.2                                                        & 48.6                                                              & 15.8                                                               & 27.0                                                                     & 8.4                                                                 & 13.9                 & 29.1                 & 25.7                 \\
    InternVL3.5-1B-MPO         & 40.3                                                        & 50.5                                                              & 22.0                                                               & 32.1                                                                     & 9.0                                                                 & 16.8                 & 32.7                 & 29.1                 \\
    InternVL3.5-1B-CascadeRL   & {44.2}                                               & {59.3}                                                     & {27.3}                                                      & {37.8}                                                            & {17.2}                                                       & {21.5}        & {29.3}        & {33.8}        \\
    \mymidrule
    InternVL3-2B               & 48.6                                                        & 57.0                                                              & 21.7                                                               & 25.3                                                                     & 14.6                                                                & 22.4                 & 36.9                 & 32.4                 \\
    InternVL3.5-2B-Instruct    & 53.0                                                        & 60.8                                                              & 27.0                                                               & 39.6                                                                     & 19.8                                                                & 28.1                 & 41.2                 & 38.5                 \\
    InternVL3.5-2B-MPO         & 54.3                                                        & 62.6                                                              & 34.2                                                               & 46.4                                                                     & 21.0                                                                & 28.1                 & 40.9                 & 41.1                 \\
    InternVL3.5-2B-CascadeRL   & {59.0}                                               & {71.8}                                                     & {42.8}                                                      & {53.4}                                                            & {31.5}                                                       & {48.5}        & {47.7}        & {50.7}        \\
    \mymidrule
    InternVL3.5-4B-Instruct    & 64.3                                                        & 71.4                                                              & 40.5                                                               & 50.0                                                                     & 30.7                                                                & 35.6                 & 53.5                 & 49.4                 \\
    InternVL3.5-4B-MPO         & 65.4                                                        & 71.7                                                              & 48.0                                                               & 54.9                                                                     & 30.7                                                                & 39.8                 & 55.9                 & 52.3                 \\
    InternVL3.5-4B-CascadeRL   & {66.6}                                               & {77.1}                                                     & {54.4}                                                      & {61.7}                                                            & {35.7}                                                       & {50.1}        & {56.4}        & {57.4}        \\
    \mymidrule
    InternVL3-8B               & 62.7                                                        & 71.6                                                              & 29.3                                                               & 39.8                                                                     & 25.5                                                                & 37.1                 & 44.1                 & 44.3                 \\
    InternVL3.5-8B-Instruct    & 68.1                                                        & 74.2                                                              & 46.4                                                               & 55.8                                                                     & 30.7                                                                & 46.0                 & 53.9                 & 53.6                 \\
    InternVL3.5-8B-MPO         & 71.2                                                        & 75.9                                                              & 52.6                                                               & 54.8                                                                     & 33.1                                                                & 47.7                 & 58.6                 & 56.3                 \\
    InternVL3.5-8B-CascadeRL   & {73.4}                                               & {78.4}                                                     & {56.8}                                                      & {61.5}                                                            & {37.7}                                                       & {57.0}        & {57.3}        & {60.3}        \\
    \mymidrule
    InternVL3-14B              & 67.1                                                        & 75.1                                                              & 37.2                                                               & 44.4                                                                     & 31.3                                                                & 43.0                 & 51.2                 & 49.9                 \\
    InternVL3.5-14B-Instruct   & 71.8                                                        & 73.4                                                              & 48.7                                                               & 55.5                                                                     & 31.9                                                                & 45.7                 & 57.5                 & 54.9                 \\
    InternVL3.5-14B-MPO        & 73.3                                                        & 74.0                                                              & 53.0                                                               & 57.5                                                                     & 32.3                                                                & 45.2                 & 60.9                 & 56.6                 \\
    InternVL3.5-14B-CascadeRL  & {73.3}                                               & {80.5}                                                     & {59.9}                                                      & {62.8}                                                            & {38.7}                                                       & {58.7}        & {60.2}        & {62.0}        \\
    \mymidrule
    InternVL3.5-30B-A3B-Instruct   & 72.3                                                        & 73.3                                                              & 45.1                                                               & 50.4                                                                     & 31.9                                                                & 39.7                 & 56.4                 & 52.7                 \\
    InternVL3.5-30B-A3B-MPO        & 71.7                                                        & 75.3                                                              & 50.7                                                               & 58.5                                                                     & 32.9                                                                & 43.7                 & 59.7                 & 56.1                 \\
    InternVL3.5-30B-A3B-CascadeRL  & {75.6}                                               & {80.9}                                                     & {55.7}                                                      & {60.4}                                                            & {36.5}                                                       & {48.4}        & {55.7}        & {59.0}        \\
    \mymidrule
    InternVL3-38B              & 70.1                                                        & 75.1                                                              & 34.2                                                               & 48.2                                                                     & 35.3                                                                & 48.6                 & 58.4                 & 52.8                 \\
    InternVL3.5-38B-Instruct   & 73.9                                                        & 75.9                                                              & 58.2                                                               & 59.0                                                                     & 29.7                                                                & 47.5                 & 60.0                 & 57.7                 \\
    InternVL3.5-38B-MPO        & 76.9                                                        & 80.5                                                              & 56.3                                                               & 59.4                                                                     & 36.9                                                                & 55.6                 & 64.2                 & 61.4                 \\
    InternVL3.5-38B-CascadeRL  & {76.9}                                               & {81.9}                                                     & {63.7}                                                      & {67.6}                                                            & {41.7}                                                       & {64.8}        & {65.3}        & {66.0}        \\
    \mymidrule
    InternVL3-78B              & 72.2                                                        & 79.0                                                              & 43.1                                                               & 51.0                                                                     & 35.1                                                                & 46.1                 & 55.9                 & 54.6                 \\
    InternVL3.5-241B-A28B-Instruct  & 76.2                                                        & 80.1                                                              & 55.6                                                               & 61.7                                                                     & 36.5                                                                & 49.7                 & 63.3                 & 60.4                 \\
    InternVL3.5-241B-A28B-MPO       & 76.0                                                        & 82.2                                                              & 55.3                                                               & 64.1                                                                     & 38.3                                                                & 51.3                 & 69.4                 & 62.4                 \\
    InternVL3.5-241B-A28B-CascadeRL & {77.7}                                               & {82.7}                                                     & {63.9}                                                      & {68.5}                                                            & {46.5}                                                       & {62.3}        & {66.7}        & {66.9}        \\
    \mybottomrule
    \end{tabular}
\vspace{2mm}
\caption{\textbf{Comparison of multimodal reasoning  performance after different training stages.} 
}
\label{tab:exp-ablation-cascade-rl}
\end{table}

\begin{table}[t]
\centering
\scriptsize
\setlength{\tabcolsep}{3pt}
\renewcommand\arraystretch{1.1}
    \begin{tabular}{l|r|ccccccc|c}
\toprule
\textbf{Model}           & \textbf{GPU Hours}       & \textbf{\begin{tabular}[c]{@{}c@{}}MMMU\\ Val\end{tabular}} & \textbf{\begin{tabular}[c]{@{}c@{}}MathVista\\ MINI\end{tabular}} & \textbf{\begin{tabular}[c]{@{}c@{}}MathVision\\ MINI\end{tabular}} & \textbf{\begin{tabular}[c]{@{}c@{}}MathVerse\\ Vision-Only\end{tabular}} & \textbf{\begin{tabular}[c]{@{}c@{}}DynaMath\\ (Worst)\end{tabular}} & \textbf{WeMath} & \textbf{LogicVista} & \textbf{Overall} \\ \mymidrule
InternVL3.5-8B-Instruct  & --                       & 68.1                                                        & 74.2                                                              & 46.4                                                               & 55.8                                                                     & 30.7                                                                & 46.0            & 53.9                & 53.6             \\
+MPO       & $\sim$0.3K Hours           & 71.2                                                        & 75.9                                                              & 52.6                                                               & 54.8                                                                     & 33.1                                                                & 47.7            & 58.6                & 56.3             \\
+GSPO (1 episode)      & $\sim$5.5K Hours          & 73.8                                                        & 77.9                                                              & 51.6                                                               & 58.8                                                                     & 35.1                                                                & 48.9            & 54.8                & 57.3             \\
+GSPO (2 episodes)      & $\sim$11.0K Hours          & 72.0                                                        & 78.1                                                              & 51.6                                                               & 58.5                                                                     & 35.7                                                                & 54.1            & 57.0                & 58.2             \\
+CascadeRL (ours) & {$\sim$5.8K Hours} & {73.4}                                               & {78.4}                                                     & {56.8}                                                      & {61.5}                                                            & {37.7}                                                       & {57.0}   & {57.3}       & {60.3}    \\ \mybottomrule
\end{tabular}
\vspace{2mm}
\caption{
    \textbf{Comparison of training efficiency and effectiveness of MPO, GSPO, and Cascade RL.}
}
\label{tab:exp-ablation-gpu-hours}
\end{table}

\textbf{Cascade Reinforcement Learning (Cascade RL)}. To validate the effectiveness of Cascade RL, we conduct an
ablation study in Figure \ref{fig:ablation_training} and Table \ref{tab:exp-ablation-cascade-rl}. We evaluate a baseline model InternVL3 as well as InternVL3.5 after different training stages: InternVL3.5-Instruct (after SFT), InternVL3.5-MPO (after the first substage in Cascade RL), and InternVL3.5-CascadeRL (after both substages in Cascade RL).
From it we can see that InternVL3.5-Instruct already outperforms InternVL3 by margins, \emph{e.g.,} + 9.3\% on the 8B model.  Even based on these strong SFT baselines, the performance of InternVL3.5 can be further improved  after the MPO stage, providing up to +3.5\% average gains on reasoning tasks. Compared to  MPO-based models,  our Cascade RL still provides orthogonal gains for all dense and MoE models.  For example, InternVL3.5-2B obtains 12.2\% average performance gains on reasoning tasks compared to the SFT model. Similar merits can also be observed on larger models, \emph{e.g.,} +6.5\% on InternVL3.5-241B-A28B.  These ablations confirm the effectiveness, stability, and scalability of our Cascade RL.
Additionally, we present a comparison of the efficiency and effectiveness of our proposed Cascade RL against MPO and GSPO in Table~\ref{tab:exp-ablation-gpu-hours}. For MPO and Cascade RL, we report performance after training for one episode, whereas for GSPO, we report results after both one and two episodes. We observe that MPO achieves performance gains with only a small number of GPU hours, while GSPO yields more significant improvements but at the cost of substantially higher computational consumption. In contrast, Cascade RL attains even greater performance improvements while requiring only half the GPU hours of GSPO.

\begin{table}[t]
\centering
\scriptsize
\setlength{\tabcolsep}{3pt}
\renewcommand\arraystretch{1.2}
\begin{tabular}{lcccccccccc}
\toprule
\textbf{Model} & \textbf{DocVQA} & \textbf{ChartVQA} & \textbf{InfoVQA} & \textbf{TextVQA} & \textbf{OCRBench} & \textbf{AI2D} & \textbf{MMStar} & \textbf{MMMU} & \textbf{Mathvista} & \textbf{Overall} \\ 
\midrule
InternVL3.5-8B             & 92.3 & 86.7 & 76.2 & 78.2 & 83.2 & 84.0 & 69.3 & 73.4 & 78.4 & 80.2 \\
InternVL3.5-8B-Flash       & 91.9 & 86.6 & 76.0 & 77.2 & 83.0 & 83.6 & 68.6 & 72.9 & 78.0 & 79.8 \\ 
\midrule
InternVL3.5-38B            & 94.0 & 88.8 & 81.0 & 82.7 & 87.0 & 87.8 & 75.3 & 76.9 & 81.9 & 83.9 \\
InternVL3.5-38B-Flash      & 93.5 & 88.1 & 81.0 & 82.1 & 86.5 & 87.2 & 75.0 & 76.3 & 81.3 & 83.4 \\ 
\midrule
InternVL3.5-30B-MoE        & 94.2 & 87.4 & 77.8 & 80.5 & 88.0 & 86.8 & 72.0 & 75.6 & 80.9 & 82.6 \\
InternVL3.5-30B-MoE-Flash  & 93.2 & 87.3 & 77.6 & 80.2 & 87.8 & 86.3 & 71.7 & 75.2 & 80.5 & 82.2 \\ 
\midrule
InternVL3.5-235B-MoE       & 94.9 & 88.0 & 82.0 & 84.5 & 90.7 & 86.9 & 77.9 & 77.7 & 82.7 & 85.0 \\
InternVL3.5-235B-MoE-Flash & 94.0 & 87.9 & 81.0 & 84.1 & 90.3 & 86.1 & 77.4 & 77.0 & 83.0 & 84.5 \\
\bottomrule
\end{tabular}

\vspace{2mm}
\caption{\textbf{Performance comparison of InternVL3.5 and InternVL3.5-Flash.}}
\label{tab:exp-ablation-vir}
\end{table}

\begin{table}[t]
\centering
\small
\renewcommand{\arraystretch}{1.1}
\begin{tabular}{clcc}
\mytoprule
\multirow{2}{*}{\textbf{Resolution}} & \multirow{2}{*}{\textbf{Setting}} & 
\multicolumn{2}{c}{\textbf{Request Throughput (requests / s)}} \\
\cmidrule(lr){3-4}
& & InternVL3.5-38B & InternVL3.5-241B-A28B \\
\mymidrule
\multirow{3}{*}{448}
& Baseline          & 12.39 & 11.29 \\
& + DvD             & 14.69 (1.19 $\times$) & 14.05 (1.24 $\times$) \\
& + DvD + ViR       & 18.62 (1.50 $\times$) & 20.84 (1.85 $\times$) \\

\mymidrule
\multirow{3}{*}{896}
& Baseline          & 2.71  & 2.54 \\
& + DvD             & ~5.06 (1.87 $\times$)  & 4.73 (1.86 $\times$) \\
& + DvD + ViR       & 10.97 (4.05 $\times$) & 8.81 (3.47 $\times$) \\

\mymidrule
\multirow{3}{*}{1344}
& Baseline          & 1.48  & 1.37 \\
& + DvD             & 2.92 (1.97 $\times$)  & 2.75 (2.01 $\times$) \\
& + DvD + ViR       & 5.14 (3.47 $\times$) & 4.27 (3.12 $\times$) \\

\mybottomrule
\end{tabular}
\vspace{3mm}
\caption{\textbf{Ablation of Decoupled Vision-Language Deployment (DvD) and Visual Resolution Router (ViR) on inference efficiency.} We send 16 requests per second to the deployed server. In all settings, the language models are deployed on 8 A100 GPUs.}
\label{tab:exp-ablation-dvd}
\vspace{-3mm}
\end{table}

\textbf{Visual Resolution Router (ViR)}. In Tables~\ref{tab:exp-ablation-vir} and \ref{tab:exp-ablation-dvd}, we compare efficiency and performance of InternVL3.5 with and without ViR,  and models equipped with ViR are  called InternVL3.5-Flash.   In Table \ref{tab:exp-ablation-dvd} , we  validate the efficiency improvement  brought by ViR.  From it we can see that the proposed DvD can already accelerate inference by up to 2.01$\times$, based on which ViR still provides significant efficiency gains, \emph{e.g.,} 4.05$\times$ speedup. Note that the efficiency gains of ViR are also obvious for the large MoE model, \emph{i.e.,} InternVL3.5-241B-A28B, which is significant for real-world application.

In Table \ref{tab:exp-ablation-vir}, we  compare the performance of InternVL3.5 and InternVL3.5-Flash on several significant benchmarks.  For these results, we can see that InternVL3.5-Flash can maintain the multimodal understanding and reasoning performance. In high-resolution tasks like DocVQA and InfoVQA, InternVL3.5-Flash can reach almost the 100\% performance of InternVL3.5, \emph{e.g.,} 80.2 \textit{vs.} 79.8 on 8B model size. Even when the model size improves to 241B, similar observations  still hold.  These results further confirm that ViR can greatly benefit the model performance without sacrificing performance.

\textbf{Decoupled Vision-Language Deployment (DvD)}. In Table \ref{tab:exp-ablation-dvd}, we conduct detailed ablation on InternVL3.5 to show the benefit of DvD. From this table, the first observation is that DvD greatly accelerates the inference of both dense and MoE models, by up to 2.01 times and 1.97 times  for InternVL3.5-241B-A28B and InternVL3.5-38B, respectively. In addition, the efficiency gains of DvD can benefit both the pre-filling and next-token prediction stages of InternVL3.5. Another finding is that as the input resolution increases, DVD efficiency also improves. For example, on InternVL3.5-38B, the speed-up of DvD can be improved from 1.19 to 1.97 as the resolution increases from 448 to 1344. This phenomenon can be attributed to the fact that larger input resolution or visual backbone leads to higher visual computational costs and further blocks the computation of the LLM.
It is worth noting that the increased computational cost of high-resolution images arises because mainstream MLLMs typically adopt a dynamic high-resolution strategy, which increases the number of patches fed into the vision encoder and thereby raises the overall computation. In practical applications, beyond high-resolution images, many tasks also require multi-image and video understanding. In such scenarios, the number of patches processed by the vision encoder grows even further, leading to greater visual overhead. We believe that our proposed DvD can deliver even more significant performance gains in these scenarios.

\section{Conclusion}

In this work, we introduce InternVL3.5, the latest family of InternVL models that demonstrates stronger general performance and faster speed across a wide range of tasks. InternVL3.5 adopts a new reinforcement learning (RL) framework, namely Cascade RL, which combines the benefits of offline and online RL methods to boost reasoning capabilities.  In addition, two techniques are further introduced to reduce the inference cost of InternVL3.5, namely Visual Resolution Router (ViR) and Decoupled Vision-Language Deployment (DvD).  Benefiting from these innovations, InternVL3.5 achieves +16.0\% improvement in overall reasoning performance and  4.05$\times$ speed-up in inference efficiency compared to InternVL3. Besides, InternVL3.5 has significant improvements in its versatility against InternVL3.  Specifically, InternVL3.5-241B-A28B achieves the highest overall score across multimodal general, reasoning, text, and agency tasks among leading open-source MLLMs, significantly narrowing the performance gap with top-tier commercial models like GPT-5. We believe that our open source models and codes will push forward multimodal AI research and its real-world applications.

{
    \small
    \bibliographystyle{plain}
    \bibliography{main}
}

\end{document}